\makeatletter\def\graphicscache@inhibit{true}\makeatother
\documentclass{article}
\usepackage[utf8]{inputenc}
\usepackage{jfrExamplee}

\usepackage{subfiles}

\usepackage{graphicx}
\usepackage[export]{adjustbox}

\graphicspath{{../}%
{figures/figures_section1/}{figures/figures_section2/}{figures/figures_section3/}{figures/figures_section4/}{figures/figures_section5/}{figures/figures_section6/}{figures/figures_section7/}{figures/figures_section8/}{figures/figures_section9/}}

\usepackage{natbib}
\let\bibhang\relax
\usepackage{apalike}

\usepackage{setspace}

\usepackage[hidelinks]{hyperref}

\usepackage[bottom]{footmisc}

\usepackage{footnote}
\usepackage{amsmath}
\usepackage[capitalise]{cleveref}
\usepackage{amssymb}
\usepackage{bm}
\newcommand\w[1]{\mathbf{#1}}
\usepackage{textcomp}
\usepackage{adjustbox}
\usepackage{subfig}
\usepackage{makecell}
\IfFileExists{graphicscache.sty}{\usepackage[hashshortnames,qfactor=0.25]{graphicscache}}{}

\usepackage{tikz}
\usetikzlibrary{arrows}
\usetikzlibrary{positioning,calc}
\usetikzlibrary{decorations.pathreplacing}
\usetikzlibrary{decorations.markings}
\usetikzlibrary{fit}
\usetikzlibrary{shapes.callouts}
\usetikzlibrary{shapes.geometric}
\usetikzlibrary{matrix}

\usepackage{placeins} %
\usepackage{booktabs} %
\usepackage{threeparttable} %
\usepackage{todonotes}
\usepackage[per=frac,binary-units=true]{siunitx}
\usepackage{enumitem} %

\usepackage[tightpage]{preview}

\newenvironment{maybepreview}%
{\noindent\ignorespaces}%
{\par\noindent%
\ignorespacesafterend}
\PreviewEnvironment{maybepreview}

\newcommand{\ie}{i.e.,\ }
\newcommand{\eg}{e.g.,\ }

\newcommand{\norm}[1]{\left\lVert#1\right\rVert}

\title{Remote Mobile Manipulation with the\\Centauro Robot: Full-body Telepresence\\and Autonomous Operator Assistance}

\author{
Tobias Klamt\footnotemark[1] \\
\texttt{klamt@ais.uni-bonn.de} \\
\And
Max Schwarz\footnotemark[1]
\And
Christian Lenz\footnotemark[1]
\And
Lorenzo Baccelliere\footnotemark[4]
\And
Domenico Buongiorno\footnotemark[2]
\And
Torben Cichon\footnotemark[3]
\And
Antonio Di Guardo\footnotemark[2]
\And
David Droeschel\footnotemark[1]
\And
Massimiliano Gabardi\footnotemark[2]
\And
Malgorzata Kamedula\footnotemark[4]
\And
Navvab Kashiri\footnotemark[4]
\And
Arturo Laurenzi\footnotemark[4]
\And
Daniele Leonardis\footnotemark[2]
\And
Luca Muratore\footnotemark[4]\textsuperscript{\hspace{1ex},}\footnotemark[6]
\And
Dmytro Pavlichenko\footnotemark[1]
\And
Arul Selvam Periyasamy\footnotemark[1]
\And
Diego Rodriguez\footnotemark[1]
\And
Massimiliano Solazzi\footnotemark[2]
\And
Antonio Frisoli\footnotemark[2]
\And
Michael Gustmann\footnotemark[5]
\And
Jürgen Roßmann\footnotemark[3]
\And 
Uwe Süss\footnotemark[5]
\And
Nikos G. Tsagarakis\footnotemark[4]
\And
Sven Behnke\footnotemark[1]
}

\usepackage{soul}
\usepackage{multicol}
\usepackage{multirow}
\usepackage{xcolor}

\usepackage{tabularx}

\newcommand{\executeiffilenewer}[3]{%
	\ifnum\pdfstrcmp{\pdffilemoddate{#1}}%
	{\pdffilemoddate{#2}}>0%
	{\immediate\write18{#3}}\fi%
}

\begin{document}

\maketitle

\footnotetext[1]{Autonomous Intelligent Systems, University of Bonn, Bonn, Germany}
\footnotetext[2]{PERCRO Laboratory, TeCIP Institute, Scuola Superiore Sant'Anna, Pisa, Italy}
\footnotetext[3]{Man-Machine Interaction, RWTH Aachen University, Aachen, Germany}
\footnotetext[4]{Department of Advanced Robotics, Italian Institute of Technology, Genoa, Italy}
\footnotetext[5]{Kerntechnische Hilfsdienst GmbH, Karlsruhe, Germany}
\footnotetext[6]{School of Electrical and Electronic Engineering, The University of Manchester, Great Britain}

\vspace{-1em}
\begin{abstract}

\begin{tikzpicture}[remember picture,overlay]%
\node[anchor=north west,align=left,font=\sffamily,yshift=-0.2cm,inner sep=1cm] at (current page.north west) {%
  Published in: Journal of Field Robotics (JFR), Wiley, 2019
};%
\node[anchor=north east, align=right,font=\sffamily,yshift=-0.2cm,inner sep=1cm] at (current page.north east) {%
  DOI: \href{https://doi.org/10.1002/rob.21895}{10.1002/rob.21895}
};%
\end{tikzpicture}%
Solving mobile manipulation tasks in inaccessible and dangerous environments is an important application of robots to support humans. 
Example domains are construction and maintenance of manned and unmanned stations on the moon and other planets.
Suitable platforms require flexible and robust hardware, a locomotion approach that allows for navigating a wide variety of terrains, dexterous manipulation capabilities, and respective user interfaces. 
We present the CENTAURO system which has been designed for these requirements and consists of the Centauro robot and a set of advanced operator interfaces with complementary strength enabling the system to solve a wide range of realistic mobile manipulation tasks.
The robot possesses a centaur-like body plan and is driven by torque-controlled compliant actuators. 
Four articulated legs ending in steerable wheels allow for omnidirectional driving as well as for making steps. 
An anthropomorphic upper body with two arms ending in five-finger hands enables human-like manipulation. 
The robot perceives its environment through a suite of multimodal sensors. 
The resulting platform complexity goes beyond the complexity of most known systems which puts the focus on a suitable operator interface.
An operator can control the robot through a telepresence suit, which allows for flexibly solving a large variety of mobile manipulation tasks. 
Locomotion and manipulation functionalities on different levels of autonomy support the operation.
The proposed user interfaces enable solving a wide variety of tasks without previous task-specific training.
The integrated system is evaluated in numerous teleoperated experiments that are described along with lessons learned.

\end{abstract}

\pagebreak
\section{Introduction}
Capable mobile manipulation robots are desperately needed in environments which are inaccessible or dangerous for humans.
Missions include construction and maintenance of manned and unmanned stations, as well as exploration of unknown environments on the moon and other planets.
Furthermore, such systems can be employed in search and rescue missions on earth.
It applies to all these missions that human deployment is impossible or dangerous, and depends on extensive logistical and financial effort. 

To address the wide range of possible tasks, a suitable platform needs to provide a wide range of capabilities. 
Regarding locomotion, exemplary tasks are to overcome a variety of obstacles which can occur on planetary surfaces and in man-made environments, e.g., in space stations. 
Regarding manipulation, tasks may be to use power tools, to physically connect and disconnect objects such as electrical plugs, or to scan surfaces, \eg for radiation. 
Since maintenance is not possible during missions, a high hardware and software reliability is necessary. 
Furthermore, suitable operator interfaces are key to enable the control of a system that must solve such a large variety of tasks.
In the European H2020 project CENTAURO\footnote{\url{https://www.centauro-project.eu}}, we develop a system according to the above requirements.  
It consists of the Centauro robot, operator interfaces including a full-body telepresence suit, autonomous locomotion and manipulation functions, and modules for communication and simulation.

\begin{figure}
	\centering\begin{maybepreview}
\begin{tikzpicture}[
 	font=\sffamily\footnotesize,
    every node/.append style={text depth=.2ex},
	box/.style={rectangle, inner sep=0.5, anchor=west},
	line/.style={red, thick}
]

\node[anchor=south west,inner sep=0] (image) at (0,0) {\includegraphics[height=7cm]{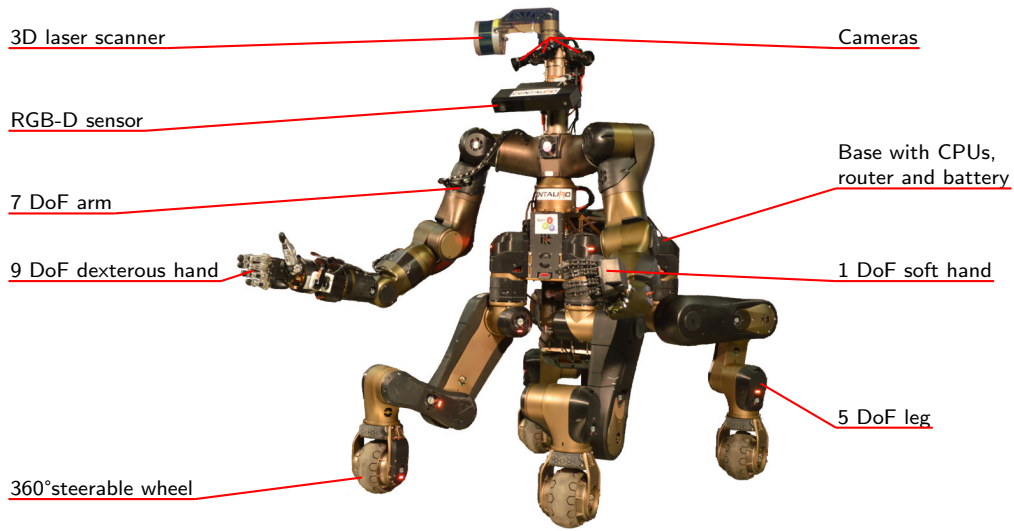}};

\node[box](laser_scanner) at(-3,6.6){3D laser scanner};
\draw[line](laser_scanner.south west)--(laser_scanner.south east);
\draw[line](laser_scanner.south east)--(3.3,6.6);

\node[box](cameras) at(8,6.6){Cameras};
\coordinate(camera_split) at (4.2,6.6);
\draw[line](cameras.south west)--(cameras.south east);
\draw[line](cameras.south west)--(4.2,6.6);
\draw[line](camera_split)--(4.1, 6.4);
\draw[line](camera_split)--(3.8,6.3);
\draw[line](camera_split)--(4.6,6.4);

\node[box](kinect) at(-3,5.5){RGB-D sensor};
\draw[line](kinect.south west)--(kinect.south east);
\draw[line](kinect.south east)--(3.5,5.7);

\node[box](arm) at(-3,4.4){7 DoF arm};
\draw[line](arm.south west)--(arm.south east);
\draw[line](arm.south east)--(3,4.6);

\node[box](schunk) at(-3,3.5){9 DoF dexterous hand};
\draw[line](schunk.south west)--(schunk.south east);
\draw[line](schunk.south east)--(0.25,3.5);

\node[box](soft_hand) at(8,3.5){1 DoF soft hand};
\draw[line](soft_hand.south west)--(soft_hand.south east);
\draw[line](soft_hand.south west)--(4.95,3.5);

\node[box](leg) at(8,1.55){5 DoF leg};
\draw[line](leg.south west)--(leg.south east);
\draw[line](leg.south west)--(7,2);

\node[box](wheel) at(-3,0.6){360\textdegree steerable wheel};
\draw[line](wheel.south west)--(wheel.south east);
\draw[line](wheel.south east)--(1.7,0.75);

\node[box, align=left](base) at(8,4.9){Base with CPUs,\\router and battery};
\draw[line](base.south west)--(base.south east);
\draw[line](base.south west)--(5.7,3.9);

\end{tikzpicture}
\end{maybepreview}
	\caption{The Centauro robot.}
	\label{fig:centauro_robot}
\end{figure}

Centauro, shown in~\cref{fig:centauro_robot}, has a centaur-like body plan with four articulated legs and an anthropomorphic upper body.
It is driven by torque-controlled series-elastic actuators. 
Each leg has five degrees of freedom (DoF) and ends in a directly driven, 360\textdegree\, steerable wheel. 
This leg design allows for both omnidirectional driving and stepping locomotion and combines their advantages. 
Sufficiently flat terrain can be traversed easily by driving, which is fast, energy efficient, and provides high stability. 
Omnidirectional motion enables the robot to navigate precisely in narrow spaces and to position itself accurately for manipulation tasks. 
More challenging terrain and height differences can be overcome by stepping, which only requires isolated footholds. 
The design further enables motions that are neither possible for pure driving nor for pure walking robots, such as changing the robot footprint under load.

Centauro's anthropomorphic upper body consists of a yaw joint in the spine and two 7\,DoF arms. 
In combination with the adjustable base height and orientation, this results in a workspace equal to an adult person.
The arms end in two different anthropomorphic five-finger hands to provide a wide range of manipulation capabilities.
A SoftHand, driven by a single actuator, is mounted on the left arm. 
Through its compliant design, it allows for robust manipulation. 
The right arm ends in a 9\,DoF Schunk hand with 20 joints, capable of dexterous, human-like manipulation.

Environment perception is realized through a variety of sensors. 
A continuously 3D rotating scanner with a spherical field-of-view provides range measurements. 
Three wide-angle RGB cameras capture images from the robot head perspective. 
An additional RGB-D sensor is mounted on a pan-tilt unit below these cameras. 
For some tasks, the robot can be equipped with additional RGB cameras at suitable positions such as under the robot base.

A system with this amount of kinematic capabilities and sensory input is novel and directs the focus to an effective operator interface. 
This needs to be intuitive to enable fast and safe operation and flexible to address unknown tasks. 
Situation awareness for the Centauro operators is realized through a simulation-based environment, generated from sensor data and enriched with semantic information. A digital twin of the robot is placed in this representation. 
This allows for flexible views on the current scene.  
A full-body telepresence suit allows an operator to intuitively control the whole robot. 
Key component is an upper-body exoskeleton, which transfers the operator's arm, wrist and hand movements to the robot. 
It provides force feedback and, thus, allows the operator to solve a large variety of manipulation tasks.
 
For operator support, we developed autonomous locomotion and manipulation functions.
This is especially helpful when operating the robot with a high latency and low bandwidth connection.
A locomotion planner provides hybrid driving-stepping paths and a respective controller executes them. 
The only required operator input is a desired robot goal pose. 
An autonomous manipulation interface detects and categorizes objects, plans and executes optimized arm trajectories towards these objects and finally provides grasping movements. 
The latter even applies to unknown objects, since suitable grasping poses are derived from known instances of the same object class.
Support operator can further control the robot through a set of locomotion and manipulation control interfaces which range from joint space control to a semi-autonomous stepping controller and possess complementary strengths.
Operators can choose from the provided interfaces depending on the individual task and resulting in an overall large flexibility and wide range of controllable capabilities.
This is especially helpful when tasks are previously unknown.
While many of these components have been individually presented in other works, we believe that the integration into a holistic remote mobile manipulation system which is evaluated in a wide range of realistic tasks is novel and goes beyond the state of the art. 

The system was evaluated during an intensive testing period at the facilities of Kerntechnische Hilfsdienst GmbH (KHG), Karlsruhe, Germany, which is part of the German nuclear disaster response organization.
KHG operates a variety of remote-controlled manipulator vehicles and has deep knowledge of the application domain.
Locomotion capabilities were evaluated in tasks like driving up a ramp, overcoming a gap, moving through an irregular step field, and autonomously climbing a flight of stairs. 
Semi-autonomous and autonomous manipulation was evaluated in tasks like drilling a hole with a power drill, screwing a screw with an electrical screw driver, cutting a cable with an electrical cutting tool, mounting a snap-hook, opening and closing lever-type and gate-type valves, connecting and disconnecting a 230V power plug and a fire hose, scanning a surface with a radiation measurement device, and grasping and operating an electrical screw driver. 
A combination of locomotion and manipulation capabilities was required for solving the task of opening and passing a regular door. 
Visual contact between the operators and the robot was not provided for any of the tasks, so the operators had to rely on the environment representation provided by the robot sensors. 
Most of the tasks were performed successfully and without previous training during this evaluation period which demonstrates the wide range of capabilities the system provides and the flexibility to adapt to unknown tasks.
The project has led to many important lessons learned regarding the design of the hardware, the operator interfaces, and the autonomous functionalities, which will be discussed after reporting the evaluation results.

\pagebreak
\section{Related Work} 
Mobile manipulation robots have been developed for a variety of fields such as search and rescue, planetary exploration, or personal assistance. 
Those robots vary in their locomotion strategy which can be realized by driving with wheels or tracks, by walking with multiple legs, or by a combination of both. 
Further differences between mobile manipulation robots can be found in their manipulation setup regarding the number of arms, their DoFs, and their effectors. 
Key to the applicability in a wide range of scenarios are teleoperation interfaces which enable teleoperators to utilize large parts of the robot capabilities while keeping the cognitive load low and the interface flexibility high.    

Driving robots have the capability to overcome long distances quickly with low energy consumption and high robot stability. 
Examples for mobile manipulation robots with a wheeled base are the NASA Centaur~\citep{mehling2007centaur} and the NimbRo Explorer~\citep{stuckler2016nimbro}. 
Tracked mobile manipulation robots, as the iRobot Packbot~\citep{yamauchi2004packbot}, are capable of overcoming more challenging terrain at the cost of less energy efficiency. 
However, all driving platforms are limited to suitable terrain. 

Legged robots have the ability to overcome more challenging terrain since they only require isolated footholds. 
Examples for biped manipulation robots are the Boston Dynamics Atlas provided to some teams for the DARPA Robotics Challenge (DRC), e.g. IHMC~\citep{johnson2015team}, or the Walk-Man robot~\citep{tsagarakis2017walk} which features a powerful actuation concept with actively controllable compliance. 
Recently, Toyota unveiled their humanoid T-HR3 robot\footnote{\url{http://corporatenews.pressroom.toyota.com/releases/toyota+unveils+third+generation+humanoid+robot+thr3.htm}}. 
It is controlled by an exoskeleton and mimics the operator's behavior. 
Examples for quadruped platforms are the Titan-IX~\citep{hirose2005quadruped}, the Boston Dynamics BigDog~\citep{raibert2008bigdog} with an attached powerful arm~\citep{abe2013dynamic}, the HyQ~\citep{semini2011design} with an attached HyArm~\citep{rehman2016design}, and the Boston Dynamics SpotMini with an attached arm\footnote{\url{https://www.youtube.com/watch?v=tf7IEVTDjng}}. 
A platform with six legs, of which some can also be used as arms, is the Lauron-V~\citep{roennau2014lauron}.
However, legged robots are often slow, compared to driving robots, less energy efficient, and require a more complex control strategy since the robots are less stable while walking.

Hybrid driving-stepping platforms combine the advantages of both locomotion strategies. 
Long distances of sufficiently even terrain can be traversed by driving, while stepping is performed on rough terrain where driving is infeasible. 
This combination allows for quick and energy-efficient locomotion in a wide variety of terrain types. 
Hybrid locomotion designs result in platforms with many DoFs, though, requiring the development of suitable teleoperation approaches to utilize all of their capabilities.  
An early example for a hybrid locomotion manipulation platform is WorkPartner~\citep{halme2003workpartner}. 
A recent bipedal platform is the Boston Dynamics Handle\footnote{\url{https://www.youtube.com/watch?v=-7xvqQeoA8c}}. 
In 2015, the DRC pushed research teams to develop robots that are capable of performing several manipulation tasks and overcoming terrain with variable complexity. 
Four of the five best teams chose a hybrid driving-stepping platform which might be an indication for the advantages of such a locomotion strategy.

RoboSimian~\citep{hebert2015mobile} is a statically stable quadrupedal robot with four generalized limbs, developed for the DRC. 
Each limb consists of seven joints with identical actuators and ends in an under-actuated hand. 
Both stepping locomotion and manipulation tasks can be performed with each limb. 
Furthermore, RoboSimian has two active wheels at its trunk and two caster wheels at two of its limbs which allow for driving on flat terrain. 
To switch between driving and stepping locomotion, the robot has to lower itself to a predefined sitting pose. 
The environment is perceived with multiple stereo cameras and the operator interface is a standard laptop from which the operator can design, parametrize and sequence predefined behaviors. 
Recently, Motiv Robotics unveiled its successor RoboMantis\footnote{\url{https://www.youtube.com/watch?v=x2s_ufwxRJE}}. 
At first sight, its design shows many similarities to RoboSimian. 
However, RoboMantis' kinematic concept moved away from the idea of generalized limbs. 
Its four legs end in wheels allowing for hybrid locomotion while additional arms can be mounted on the platform for manipulation. 
In addition, a laser scanner is utilized for environment perception. 

The DRC winner robot DRC-HUBO~\citep{zucker2015general} is a humanoid robot with 32 DoF. 
In addition to bipedal walking, it is able to lower itself to a kneeing pose and perform driving locomotion over flat terrain due to four small wheels at its knees and ankles. 
The robot uses two 7-DoF arms which end in hands with three fingers. 
The right hand is extended by an additional trigger finger. 
The environment is perceived by a laser scanner, three RGB cameras and a RGB-D camera. 
Operation is apportioned among three operators with different tasks which control the robot by selecting and adapting predefined poses.

Another hybrid locomotion platform from the DRC is CHIMP~\citep{stentz2015chimp}. 
It has a roughly anthropomorphic body with two arms and two legs providing 39 DoF. 
All four limbs are equipped with powered tracks which enable driving with two or four ground contact points. 
Each arm has seven DoF and ends in a three-finger gripper. 
Environment perception is done with two laser scanners, a panomorphic camera and four stereo cameras. 
The CHIMP operator interface is a combination of manual and autonomous control. 
Task-specific motions are configured through wizards by the operator before being executed.

Finally, the centaur-shaped mobile manipulation robot Momaro~\citep{schwarz2017nimbro} of team \mbox{NimbRo} was one of the top participants at the DRC. 
With its 4-DoF passively compliant legs ending in 360\textdegree\,steerable pairs of wheels, it provides hybrid driving-stepping locomotion as well. 
In contrast to the previously presented systems, Momaro is capable of omnidirectional driving which is helpful for precise navigation. 
Since no posture change is necessary to switch between driving and walking mode, it is possible to perform unique motions such as changing its configuration of ground contact points under load. 
Momaro's upper body consists of two arms with seven DoF each and two grippers with eight DoF and exchangeable fingers. 
Environment perception is realized through a continuously rotating laser scanner with spherical field-of-view, seven RGB cameras and an infrared distance sensor mounted at one hand. 
The operator interface offers a variety of options to control the 56\,DoF robot. 
Omnidirectional driving is controlled by a joystick. 
Legs and arms can be moved with a keyframe editor in Cartesian and joint space. 
In addition, the operator can access predefined motions. 
For stepping locomotion, a semi-autonomous stepping strategy was presented for stair climbing which relies on perceived terrain heights~\citep{schwarz2016hybrid}, and manipulation tasks can be performed via bimanual telemanipulation~\citep{Rodehutskors:Humanoids2015}. 
A teleoperator wears a head-mounted display which shows an egocentric view of the scene from the robot perspective and includes head movements in the displayed scene. 
Two hand-held controllers with magnetic trackers project the operators hand movements to the robot arms. 
Momaro was also used at the DLR SpaceBot Cup 2015~\citep{schwarz2016supervised}. 
In a planetary exploration scenario, it demonstrated navigation and omnidirectional driving in rough terrain, the identification and transportation of objects, taking a soil sample, and performing assembly tasks. 
All tasks were performed with high degree of autonomy and only few operator interventions. 

Centauro combines Momaro's kinematic design with the compliant actuation concept of Walk-Man.
While the centaur-like hybrid wheeled-legged body plan is still used, several aspects were improved. 
Momaro's off-the-shelf actuators tended to overheat in some situations and the passively compliant leg links made precise end-effector control difficult. 
For the Centauro design, we use custom built series-elastic actuators which provide the option of compliant behavior that can be actively adapted to the situation. 
In contrast to Momaro, which only provided leg movement in the sagittal plane, Centauro's legs contain an additional hip yaw joint which increases movement capabilities, especially while stepping. 
Since teleoperating such hybrid-driving stepping locomotion is very complex and puts a high cognitive load on the teleoperator, we introduce a respective planning approach which generates and executes hybrid-driving stepping locomotion paths autonomously. 
Regarding manipulation, the bimanual teleoperation approach showed the capabilities to transmit the operator's behavior to the robot but due to imprecision and the lack of force feedback, the operation was still challenging and limited in its capabilities. 
Centauro's manipulation can be teleoperated by a full-body telepresence suit which eliminates those weak points. 
Autonomous manipulation functionalities for arm movements and grasping further support the operator. 
To increase manipulation capabilities, we utilize an anthropomorphic Schunk hand~\citep{ruehl2014experimental} and a flexible SoftHand~\citep{catalano2014adaptive} with complementary characteristics. 
Finally, a simulation-based visualization system provides an operator interface with high flexibility.

\pagebreak
\section{System Overview} 

\begin{figure}
	\centering\begin{maybepreview}
\begin{tikzpicture}[
 	font=\sffamily\footnotesize,
    every node/.append style={text depth=.2ex},
	bigbox/.style={rectangle,rounded corners,draw=black,align=center},
	smallbox/.style={rectangle,rounded corners,draw=black,align=center, font=\sffamily\scriptsize}
]

\coordinate (oper) at (0,-0.5);
\draw[rounded corners, fill=green!25] (oper) rectangle ++(2.5, 7.5);
\node at ($(oper) + (1.25,3.75)$) {\textbf{Operators}};

\coordinate (opvis) at (3.5, 6);
\draw[rounded corners, fill=orange!35] (opvis) rectangle ++(4,1);
\node at ($(opvis) + (2,0.5)$) {\textbf{Operator Visualization}};

\coordinate (advenvper) at (9,4);
\draw[rounded corners, fill=blue!5] (advenvper) rectangle ++(4,3);
\node at ($(advenvper) + (2,2.5)$) [align=center]{\textbf{Advanced Environment}\\\textbf{Perception}};
\node[smallbox, fill=blue!20] at ($(advenvper) + (1.25,1.6)$) {Ground Contact\\Estimation};
\node[smallbox, fill=blue!20] at ($(advenvper) + (3.1,1.6)$) {3D Laser\\SLAM};
\node[smallbox, fill=blue!20] at ($(advenvper) + (1.1,0.7)$){Object\\Segmentation};
\node[smallbox, fill=blue!20] at ($(advenvper) + (2.94,0.7)$) {Object Pose\\Estimation};

\coordinate (teleop) at (3.5,1.8);
\draw[rounded corners, fill=blue!20!red!20] (teleop) rectangle ++(4,4);
\node at ($(teleop) + (2,3.7)$) [align=center]{\textbf{Teleoperation}};
\node[gray] at ($(teleop) + (2,3.3)$) [align=center]{Locomotion};
\node[smallbox, fill=blue!40!red!40] at ($(teleop) + (0.75,2.85)$) {Joystick};
\node[smallbox, fill=blue!40!red!40] at ($(teleop) + (0.6,2.15)$) {Pedal\\Box};
\node[smallbox, fill=blue!40!red!40] at ($(teleop) + (2.85,2.85)$) {Keyframe Editor};
\node[smallbox, fill=blue!40!red!40] at ($(teleop) + (2.72,2.15)$) {Semi-autonomous\\Stepping Controller};
\node[gray] at ($(teleop) + (2,1.4)$) [align=center]{Manipulation};
\node[smallbox, fill=blue!40!red!40] at ($(teleop) + (0.65,0.75)$) {6D\\Mouse};
\node[smallbox, fill=blue!40!red!40] at ($(teleop) + (1.95,0.75)$) {Keyframe\\Editor};
\node[smallbox, fill=blue!40!red!40] at ($(teleop) + (3.3,0.75)$) {Exo-\\skeleton};

\coordinate (auto) at (3.5, -0.5);
\draw[rounded corners, fill=yellow!15] (auto) rectangle ++(4,2);
\node at ($(auto) + (2,1.7)$) [align=center]{\textbf{Autonomy}};
\node[smallbox, fill=yellow!50] at ($(auto) + (1,0.8)$) {Hybrid\\Driving-\\stepping\\Locomotion};
\node[smallbox, fill=yellow!50] at ($(auto) + (2.9,0.8)$) {Autonomous\\Manipulation};

\coordinate (robo) at (\textwidth, -0.50);
\draw[rounded corners, fill=red!10] (robo) rectangle ++(-2.5, 7.5);
\node at ($(robo) + (-1.25,7)$) {\textbf{Robot}};
\node[smallbox, fill=red!30] at ($(robo) + (-1.25,6)$) {Robot State\\Sensors};
\node[smallbox, fill=red!30] at ($(robo) + (-1.25,5)$) {Environment\\Sensors};
\node[smallbox, fill=red!30] at ($(robo) + (-1.25,3)$) {Wrist\\Force/Torque\\Sensors};

\draw[-latex] ($(opvis) +(0,0.5)$) -- ++(-1,0);
\draw[-latex] ($(advenvper) +(0,2.8)$) -- ++(-1.5,0);
\draw[-latex] ($(teleop) +(4,3.6)$) -| ++(0.6,0.8) -- ++(-0.6,0);
\draw[-latex] ($(auto) +(4,1.6)$) -| ++(0.8,5.4) -- ++(-0.8,0);
\draw[-latex] ($(robo) +(-2.5,6.5)$) -- ++(-1,0);
\draw[-latex] ($(advenvper) +(0,0.7)$) -- ++(-1.5,0);
\draw[-latex] ($(advenvper) +(2,0)$) |- ++(-3.5,-3.5);
\draw[-latex] ($(oper) +(2.5,4.25)$) -- ++(1,0);
\draw[-latex] ($(oper) +(2.5,1)$) -- ++(1,0);
\draw[-latex] ($(teleop) +(4,1)$) -- ++(6.5,0);
\draw[-latex] ($(auto) +(4,0.5)$) -- ++(6.5,0);
\draw[-latex] ($(robo)+(-2.1,3)$) -- ++(-7.05,0);

\end{tikzpicture}
\end{maybepreview}
	\caption{System architecture overview.}
	\label{fig:system_overview}
\end{figure}
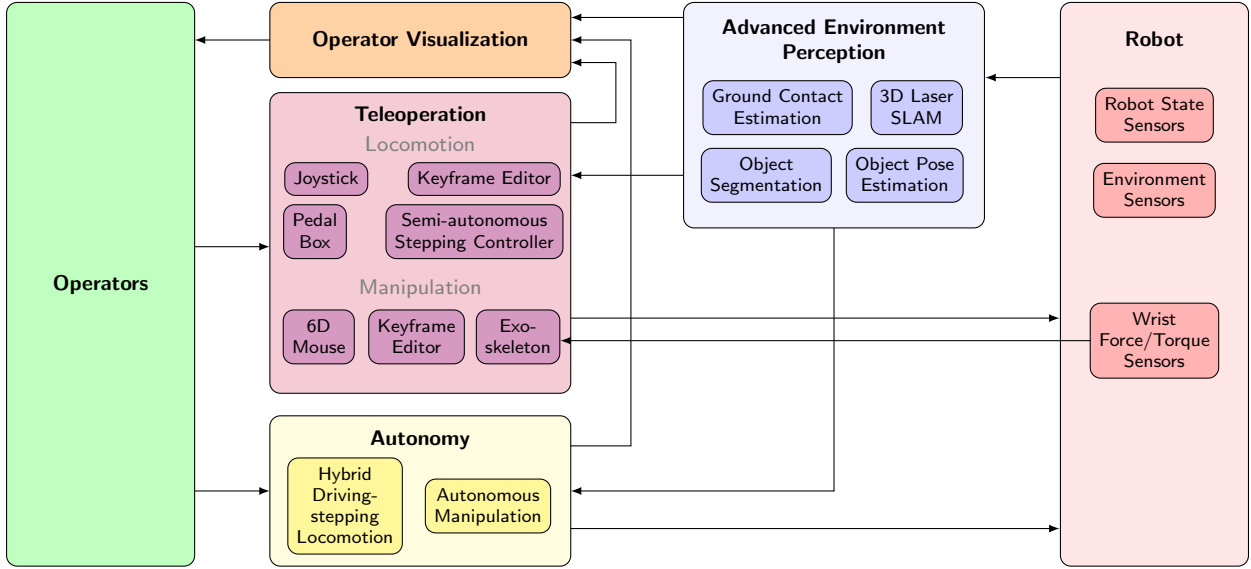

The proposed system architecture possesses multiple hardware and software components.
An overview over these components and the communication architecture between them is given in this section.

Starting point is the Centauro robot. 
It perceives information about its internal state and the environment through a 3D rotating laser scanner, a RGB-D sensor, multiple RGB cameras, and internal sensors.
The hardware and sensor setup is described in~\cref{sec:centauro_hardware}.

Sensor measurements are processed by the advanced environment perception components to obtain ground contact estimation for each foot, registered point clouds for 3D mapping and localization, as well as object segmentation and respective pose estimation.
Those components are described in~\cref{sec:advanced_environment_perception}.
In addition, raw RGB images are used for operator visualization and measurements from the force/torque sensors at the robot wrists are used for force feedback in the telepresence suit. 

To provide situation awareness, the operator visualization displays RGB images, task specific control interfaces as well as a simulation-based digital twin of the robot.
The latter uses data from RGB(-D) cameras and registered point clouds to generate an environment and robot model.
Details can be found in~\cref{sec:oi:situationawareness}.

Regarding the robot control, we propose a set of teleoperation interfaces with different levels of autonomy to provide locomotion and manipulation control for a wide range of tasks.
Driving locomotion can be controlled by a joystick while stepping locomotion is addressed by a keyframe editor and a semi-autonomous controller for stepping locomotion in rough terrain.
The latter uses the ground contact estimation to autonomously adapt to the ground structure.
Details are given in~\cref{sec:locomotion}.
Manipulation can be controlled by the same keyframe editor as well as a 6D input device.
Major effort has been spent in the development of a full-body telepresence station to provide advanced teleoperation.
It consists of a pedal box for driving control and an upper-body exoskeleton for intuitive bimanual teleoperation of the robot arms, wrists, and end-effectors while providing force-feedback.
The exoskeleton hardware is described in~\cref{sec:telepresence_suit} while~\cref{sec:manipulation_control} describes the manipulation control interfaces.
While, besides the telepresence suit, most operator interfaces are well known, we see a novelty of our work in the combination and the integration of these interfaces to enable the operator-robot team to solve a considerably wide range or realistic tasks.

Autonomous control functions require less input from the operator which decreases the dependency on the communication link while relying on additional data from the robot sensors.
A hybrid driving-stepping locomotion planner which uses registered point clouds for the environment representation and plans and executes paths to goal poses that are defined by the operator.
It is described in~\cref{sec:locomotion_planner}.
For autonomous manipulation, objects from the object segmentation component and their respective poses from the pose estimation component are used to plan grasps.
Arm trajectories towards the grasp poses are optimized using registered point clouds for environment representation.
Details are given in~\cref{sec:autonomous_grasping}.
All control interfaces require knowledge about the current robot state and output desired poses to the robot as well as respective information for the operator visualization.

\section{Mechatronic Design of Centauro Robot and Exoskeleton}
\label{sec:mechanic_design}
The CENTAURO system is designed for mobile manipulation missions in challenging environments. 
Since the task details in such missions are in general unknown in advance, 
the system has to cover a wide variety of navigation and manipulation capabilities. 
Regarding navigation capabilities, addressed environments are either planetary surfaces or man-made and thus, contain, \eg flat and rough surfaces, ramps, and stairs. 
Those environments may, however, be affected, e.g., by cluttered debris. 
Regarding manipulation, the usage of tools and objects is essential for many tasks such as drilling holes or cutting objects.
Moreover, it is desirable to use tools which are designed for human usage since those are easily available, and sufficiently tested to provide a high reliability.
The robot hardware design is described in~\cref{sec:centauro_hardware}. 
The design of a full-body telepresence suit for the operator is described in~\cref{sec:telepresence_suit}.

\subsection{The Centaur-like Robot Centauro}
\label{sec:centauro_hardware}
According to the requirements of considered scenarios, the Centauro robot (see \cref{fig:centauro_robot}) is designed as a centaur-like platform embodying an anthropomorphic upper-body and a quadrupedal lower-body---bringing together Momaro's kinematics and Walk-Man's compliant actuation.
The legs are equipped with active wheels at their lower end and the two arms possess two hands with complementary capabilities as end-effectors. 
\cref{sec:centauro_kinematics} describes the kinematics in detail,~\cref{sec:centauro_actuation} explains the developed actuators and~\cref{sec:centauro_software} focuses on the software architecture. 
Moreover, the robot platform includes a head module comprising a set of cameras and sensors. 
It encompasses a \emph{Microsoft Kinect V2} RGB-D sensor~\citep{fankhauser2015} whose tilt and pen angle can be actively controlled.
In addition, an array of three \emph{PointGrey BlackFly BFLY-U3-23S6C} wide angle color cameras and a rotating \emph{Velodyne Puck} 3D laser scanner with a spherical field-of-view provide further environment perception.
A \emph{VectorNav VN-100} inertial measurement unit (IMU) is mounted in the torso.
Two additional RGB cameras are mounted under the robot base to get a view on the feet.

Furthermore, the robot base incorporates three computing units.
One unit is responsible for the real-time control of the robot. 
It possesses an \emph{Intel Core i7-6820EQ}@2.8\,Ghz, 16\,GB RAM and a 120\,GB SSD and runs with the XENOMAI RT development kit\footnote{\url{https://xenomai.org/}}.
The other two computing units are used for perception and high-level robot control.
They both possess an \emph{Intel Core i5-7500T}@2.7\,GHz, 32\,GB RAM, a \emph{GeForce GTX 1070}, and a 500\,GB SSD. 
In addition, to allow for untethered operations, the robot pelvis accommodates a \emph{Netgear Nighthawk X10 R900} wireless communication router and a 7.5\,kg Lithium-Ion polymer battery of 34.6\,Ah capacity supplying 48\,V with 80\,A max current discharge (limited by PCM).
This provides power for about two hours of operation when performing standard manipulation and locomotion tasks.

\subsubsection{Kinematics}
\label{sec:centauro_kinematics}

\begin{figure}
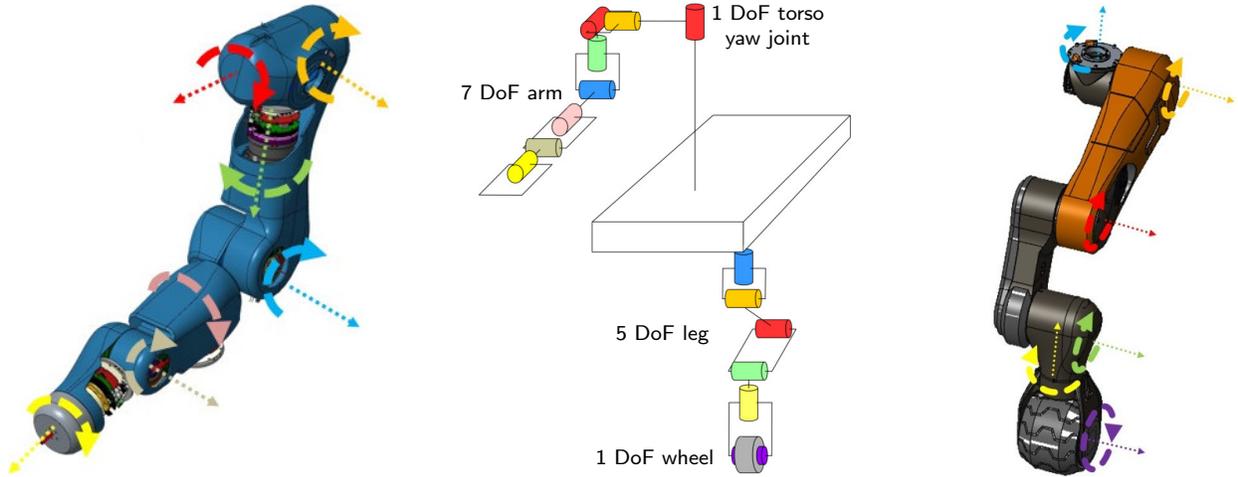

	\centering\begin{maybepreview}
\begin{tikzpicture}[
    font=\sffamily\footnotesize,
    every node/.append style={text depth=.2ex},
	sensor/.style={rectangle,rounded corners,draw=black,fill=red!20,align=center},
    module/.style={rectangle,rounded corners,draw=black,fill=yellow!20,align=center},
    group/.style={rectangle,rounded corners,draw=black},
    extinput/.style={sensor,fill=black!10},
    l/.style={font=\sffamily\scriptsize},
]
\definecolor {mywhite}{RGB}{255, 255, 255};

\node[anchor=north east,inner sep=0] (image) at (\textwidth,0) {\includegraphics[height=0.38\textwidth]{figures/figures_section3/Centauro_leg3.png}};
\node[anchor=north west,inner sep=0] (image) at (0,0) {\includegraphics[height=0.38\textwidth]{figures/figures_section3/Centauro_arm3.png}};
\node[anchor=north west,inner sep=0] (image) at (6.25,0) {\includegraphics[width=0.3\textwidth]{figures/figures_section3/kinematic_chain2.png}};

\node[] at (6.7,-1.2) {7 DoF arm};
\node[align=center] at (10.1,-0.3) {1 DoF torso\\yaw joint};
\node[] at (8.7,-4.4) {5 DoF leg};
\node[] at (8.6,-6.05) {1 DoF wheel};

\end{tikzpicture} %
\end{maybepreview}
	\caption{Kinematic layout. Left: The right arm. Joint axes are marked with colored lines. Center: Kinematic tree. For clarity, only one arm and one leg are pictured. The hands are neglected. Proportions are not to scale. Right: The front left leg.}
	\label{fig:centauro_kinematic_overview}
\end{figure}

The lower-body comprises four wheeled legs of 81\,cm length, whose first joints are accommodated in the four corners of the base structure. 
\cref{fig:centauro_kinematic_overview} illustrates the kinematic scheme and joint positioning.
To allow for versatile locomotion, each leg consists of five DoF in a spider-like configuration, which can be more beneficial in terms of stability required for the manipulation of powerful tools, as shown in~\citep{KashiriEvaluationHipKinematics2016}. 
Furthermore, in this configuration, the first leg joint has to deliver substantially lower effort and power compared to a mammal-like configuration. 
According to the chosen spider-like configuration, each hip module consists of a yaw and a pitch joint, followed by another pitch joint in the knee. 
Each ankle consists of a pitch and a yaw joint which allow for steering the wheel and adjusting its steering axis to the ground. 
Finally, each leg ends in an actively drivable wheel with 78\,mm radius. 
The described configuration allows for omnidirectional driving as well as for articulated stepping locomotion. 
Since no posture change is needed to switch between the two, it is even possible to perform motions which are unique for that design such as moving a foot relative to the base while under load. 
Thus, a wide range of locomotion capabilities is provided. 
To permit versatile leg articulation in difficult terrains, the ranges of the leg joints were maximized while taking into account the mechanical and electrical interfacing constraints.

The robot torso incorporates two arms with seven DoF each and an additional rotational joint in the waist, to endow the upper-body with yaw rotation. 
The kinematics of the two arms closely resembles an anthropomorphic arrangement to provide a large workspace, to enable dexterous manipulation, and to simplify teleoperation. 
Each arm is 73\,cm long and comprises of three DoFs at the shoulder, one DoF at the elbow and another three DoFs at the wrist. 
The degree of redundancy helps to overcome possible constraints that may be introduced in the task space by the surrounding environment. 
Even though this is a traditional design that aims at replicating the anthropomorphic structure of the human arm with seven DoF, it is only approximately equivalent for the human arm kinematic structure. 
To extend the range of motion of the elbow joint, an off-center elbow configuration was chosen. 
Similarly, for the wrist, a non-anthropomorphic configuration with non-intersecting axes was considered to maximize the range of motion of the wrist flexion and abduction motions. 
Finally, humans have the ability to elevate (upward/downward) and to incline (forward/backward) the shoulder joint, utilizing supplementary kinematic redundancy of the arm to achieve certain goals in task coordinates. 
This, however, would require the addition of two more DoF to each arm, increasing the complexity/weight and dimensions of the overall platform.
To avoid this, while at the same time obtain, to some extent, the benefits provided by the elevation (upward/downward) and inclination (forward/backward) of the shoulder, a fixed elevated and inclined shoulder arrangement was selected based on the optimization study in which important manipulation indices were considered and evaluated in a prioritized order~\citep{BaccelliereDevelopmenthumansize2017}.

\begin{figure}
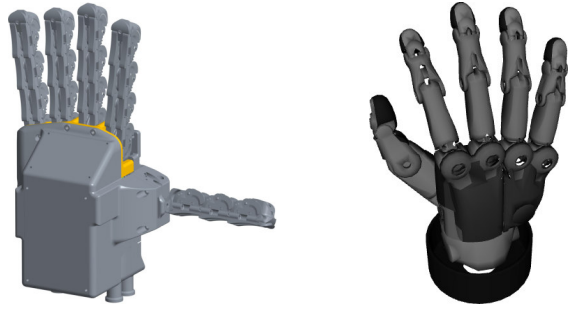

	\centering\begin{maybepreview}
	\includegraphics[height=4cm]{softhand.jpg}
	\hspace{1cm}
	\includegraphics[height=4cm]{schunk_hand.png}\end{maybepreview}
	\caption{Centauro manipulation end-effectors: 1\,DoF SoftHand (l.) and anthropomorphic 9\,DoF Schunk hand (r.).}
	\label{fig:centauro_hands}
\end{figure}

The two arms end in different end-effectors with complementary properties to provide an overall wide range of manipulation capabilities. 
Both hands are shown in~\cref{fig:centauro_hands}.
On the left side, a SoftHand with one DoF provides compliant and robust manipulation. 
The right arm utilizes an anthropomorphic Schunk hand with nine DoF for dexterous manipulation tasks. 
A customized force-torque sensor between the right arm wrist and the Schunk hand measures 6D forces/torques which are applied to the end-effector and can be used for force feedback by the exoskeleton.

Overall, the robot possesses 52 DoFs and weighs 92\,kg. The base width and length are both 61\,cm.
The whole robot is 171\,cm tall when legs are fully extended, while the robot height can be reduced to 112\,cm when folding the legs. 

\subsubsection{Compliant Actuation}
\label{sec:centauro_actuation}

\begin{figure}[b!]
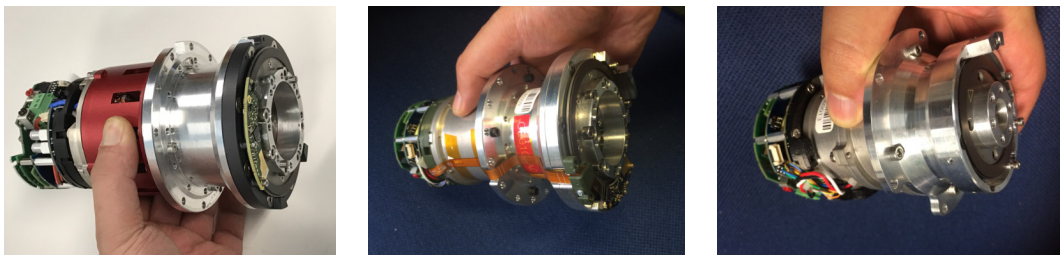

	\centering\begin{maybepreview}
	\includegraphics[height=0.2\linewidth, trim=200 200 200 100, clip]{figures/figures_section3/Actuator-Large.JPG} \hspace{0.2cm}
	\includegraphics[height=0.2\linewidth, trim=200 100 200 100, clip]{figures/figures_section3/Actuator-Medium.JPG} \hspace{0.2cm}
	\includegraphics[angle=90, height=0.2\linewidth, trim=300 0 0 300, clip]{figures/figures_section3/Actuator-Small.JPG}\end{maybepreview}
	\caption{Manufactured actuation units of the identified classes (f.l.t.r.): Large, Medium, Small.}
	\label{fig:actuators}
\end{figure} 

Having defined the kinematics of the limbs, the required actuation performance for each joint could be derived. 
To get an estimation of the actuation needs, a series of simulation studies was executed using an initial model of the robot based on estimated rigid body dynamic properties~\citep{KamedulaCompliantActuationDynamics2016}. 
A set of trajectories with different frequencies which explored the overall robot workspace was executed, while carrying a 10\,kg payload per arm. 
Respective joint torques were examined, resulting in the identification of a set of actuator classes with different sizes and torque levels ranging from about 30 to 270\,Nm. 
\cref{fig:actuators} shows the manufactured actuation units of the identified classes. 
\cref{tab:actuator_specs} shows the respective specifications. 
For the medium and small sized actuators there are two versions each with either a higher maximum velocity or higher continuous torque.

The series-elastic actuation (SEA) technology is utilized to protect the reduction gear against impacts---improving the system sturdiness, while at the same time being used for measuring the actuator torque through the monitoring of the elastic element deflection. 
Considering the influence of different joints' stiffness levels on the natural dynamics and control of the robot, as discussed by~\citep{Kashiristiffnessdesignintrinsic2013} and~\citep{roozing2017stiffness}, and taking into account the available space for the different actuators, two technologies were utilized for joint torque measurement based on strain-gauge and deflection-encoder principles~\citep{KashiriSensorDesignTorque2017}. 
The stiffness of the SEA deflection-encoder-based sensor is defined with respect to the required torque measurement resolution across the different joints. 
It was set ten times lower than the cogging torque of the motor drive when deflection of the sensor flexure is measured using a high resolution 19-bit absolute encoder.

\begin{table}
 \centering
 \caption{Actuator specifications.}\label{tab:actuator_specs}
 \begin{tabular}{l|c|c|c|c|c|c|r}
 	Type & \makecell{Gear\\ratio} & Joints & \makecell{Maximum\\velocity\\{[rad/s]}} & \makecell{Torque [Nm]\\peak -\\continuous} & \makecell{Power [W]\\peak -\\continuous} & \makecell{Torque\\sensing\\resolution\\{[Nm]}} & \makecell{Mass\\{[kg]}} \\ \hline \hline
 	Large    & 120 & Hip, Knee                                        &  8.8 & 268 - 92 & 2096 - 778 & 0.2  & 1.73 \\ \hline
 	Medium A & 160 & \makecell{Ankle Pitch,\\Elbow,\\Shoulder yaw}     &  6.1 & 147 - 46 & 820 - 259  & 0.07 & 1.28 \\ \hline
 	Medium B & 160 & \makecell{Torso,\\Shoulder pitch,\\Shoulder roll} &  3.9 & 147 - 81 & 1014 - 295 & 0.07 & 1.45 \\ \hline
 	Small A  & 100 & \makecell{Forearm yaw,\\Forearm pitch}           & 11.7 &  55 - 17 & 556 - 179  & 0.07 & 1.0 \\ \hline
 	Small B  & 100 & \makecell{Wheel,\\Ankle yaw,\\Wrist yaw}          & 20.4 &  28 - 9  & 518 - 167  & 0.07 & 0.87 \\ \hline
 \end{tabular}
\end{table}

\subsubsection{Low-Level Software Architecture}
\label{sec:centauro_software}
For the low-level control of the Centauro platform, we developed \emph{XBotCore} (\emph{Cross-Bot-Core}), a light-weight, real-time (RT) software platform for robotics~\citep{MuratoreXBotCoreRealTimeCrossRobot2017}. 
\emph{XBotCore} is open-source\footnote{\url{https://github.com/ADVRHumanoids/XBotCore}} and is designed to be both an RT robot control framework and a software middleware. 
It satisfies hard RT requirements, while ensuring a 1\,kHz control loop even in complex multi-DoF systems. 
The \emph{XBotCore} Application Programming Interface (API) enables an easy transfer of developed software components to multiple robot platforms (cross-robot feature), inside any robotic framework or with any kinematics/dynamics library as a back-end. 
Out-of-the-box implementations are available for the \emph{YARP} and \emph{ROS} software frameworks and for the \emph{RBDL} and \emph{iDynTree} dynamics libraries.

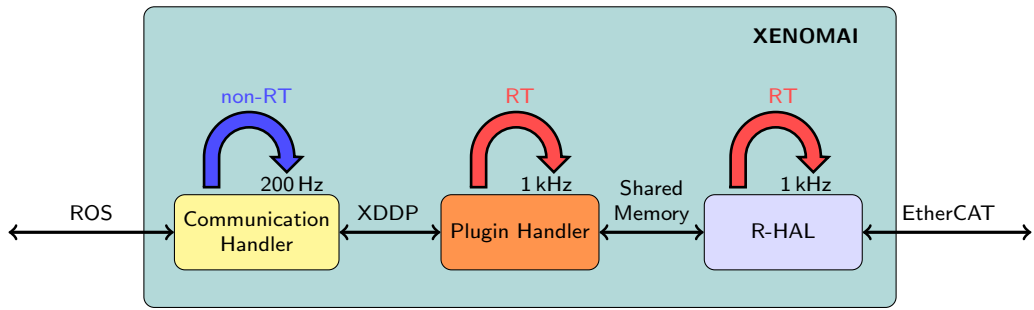
\begin{figure}[b!]
	\centering\begin{maybepreview}
\makeatletter
\newcommand\current{\the\tikz@lastxsaved,\the\tikz@lastysaved}
\makeatother

\begin{tikzpicture}[
    font=\sffamily\footnotesize,
    every node/.append style={text depth=.2ex},
	sensor/.style={rectangle,rounded corners,draw=black,fill=red!20,align=center,minimum width=2cm,minimum height=1cm},
    module/.style={rectangle,rounded corners,draw=black,fill=blue!50!green!30!white,align=center,minimum width=2cm,minimum height=1cm},
    group/.style={rectangle,rounded corners,draw=black},
    test/.style={rectangle, rounded corners, minimum height=1cm, minimum width=2cm, fill=yellow!20, align=center},
    extmodule/.style={module,dashed,fill=black!10},
    l/.style={font=\sffamily\scriptsize},
   simmodule/.style={module,fill=blue!20},
]

\node[module,anchor=center,minimum width=10cm, minimum height=4cm] (box) at (0,1) {};

\node[rectangle, rounded corners, minimum height=1cm, minimum width=2.1cm, fill=yellow!50, align=center,draw=black](ch) at(-3.5,0){Communication\\Handler};
\node[rectangle, rounded corners, minimum height=1cm, minimum width=2.1cm, fill=red!60!yellow!70, align=center,draw=black](ph) at(0,0){Plugin Handler};
\node[rectangle, rounded corners, minimum height=1cm, minimum width=2.1cm, fill=blue!70!white!20, align=center,draw=black](rhal) at(3.5,0){R-HAL};
\node[align=center] at(3.8,2.6){\textbf{XENOMAI}};

\draw[<->,very thick] (ch) -- (ph) node [anchor=south] at(-1.75,0) {XDDP};
\draw[<->,very thick] (ph) -- (rhal) node [anchor=south, align=center] at(1.75,0) {Shared\\Memory};
\draw[<->, very thick] (ch) -- (-6.8,0) node [anchor=south, align=center] at (-5.7,0) {ROS};
\draw[<->, very thick](rhal) -- (6.8,0) node [anchor=south] at (5.7,0){EtherCAT};

\draw[fill=blue!70,line width=1pt] (-4.2,0.6) -- (-4.2,1) arc (180:0.5:0.6cm) -- (-2.9,1) -- (-3.1,0.8) -- (-3.3,1)  -- (-3.2,1)arc (0.5:180:.4) -- (-4,0.6) -- (-4.217,0.6);
\node[align=center] at(-3.03, 0.63) {200\,Hz};
\node[align=center, text=blue!70] at (-3.5,1.8) {non-RT};

\draw[fill=red!70,line width=1pt] (-0.65,0.6) -- (-0.65,1) arc (180:0.5:0.6cm) -- (.65,1) -- (.45,0.8) -- (.25,1)  -- (.35,1)arc (0.5:180:.4) -- (-.45,0.6) -- (-.667,0.6);
\node[align=center] at(.35, 0.63) {1\,kHz};
\node[align=center, text=red!70] at (0,1.8) {RT};

\begin{scope}[shift={(7.0,0)}]
\draw[fill=red!70,line width=1pt] (-4.2,0.6) -- (-4.2,1) arc (180:0.5:0.6cm) -- (-2.9,1) -- (-3.1,0.8) -- (-3.3,1)  -- (-3.2,1)arc (0.5:180:.4) -- (-4,0.6) -- (-4.22,0.6);
\node[align=center] at(-3.2, 0.63) {1\,kHz};
\node[align=center, text=red!70] at (-3.5,1.8) {RT};
\end{scope}

\end{tikzpicture}
\end{maybepreview}
	\caption{\emph{XBotCore} threads and communication architecture.}
	\label{fig:xbot_centauro}
\end{figure}

A Robot Hardware Abstraction Layer (R-HAL), introduced in~\citep{RiganoMuratoreIRC18} that permits to seamlessly program and control any robotic platform powered by \emph{XBotCore} is also provided by the framework. 
Moreover, a simple and easy-to-use middleware API, for both RT and non-RT control frameworks is available. 
The \emph{XBotCore} API is completely flexible with respect to the external control framework the user wants to utilize. 
Inside the Centauro robot, an RT Cartesian control plugin based on the OpenSoT~\citep{OpenSot17}, a hierarchical whole-body control library and the built-in \emph{ROS} support are used. 
As shown in~\cref{fig:xbot_centauro}, \emph{XBotCore} spawns three threads in the Linux Xenomai\footnote{\url{https://xenomai.org}} RTOS:
\begin{itemize}[topsep=-5pt,noitemsep]
	\item The R-HAL RT thread is running at 1\,kHz and is responsible to manage and synchronize the EtherCAT slaves in the robot, \ie the electronic boards responsible for motor control and sensor data acquisition.  
	\item The Plugin Handler RT thread is running at 1\,kHz and is responsible to start all the loaded plugins, execute them 			sequentially and close them before unload. It is possible to dynamically load and unload one or more plugins in the Plugin Handler. 
	As an example, the above-mentioned RT Cartesian control plugin is running inside the Plugin Handler. 
	A shared memory communication mechanism is used to share data between this component and the R-HAL at 1 kHz.
	\item The Communication Handler non-RT thread is running at 200\,Hz and is responsible for the communication with external frameworks.
This component provides the option to send the desired robot state from the non-RT API to the chosen communication framework and to receive the reference, respectively. 
The Communication Handler uses XDDP (Cross Domain Datagram Protocol) for the asynchronous communication between RT and non-RT threads, guaranteeing a lock-free IPC (Inter-Process Communication). 
The run loop of this component is quite simple: it updates the internal robot state using the XDDP pipe with the non-RT robot API, sends the robot state to all the communication frameworks, receives the new reference from the requested "master" (we avoid to have multiple external frameworks commanding the robot) and finally, sends the received reference to the robot using the XDDP non-RT robot API.	
\end{itemize}

\subsection{The Full-body Telepresence Suit}
\label{sec:telepresence_suit}
\begin{figure}
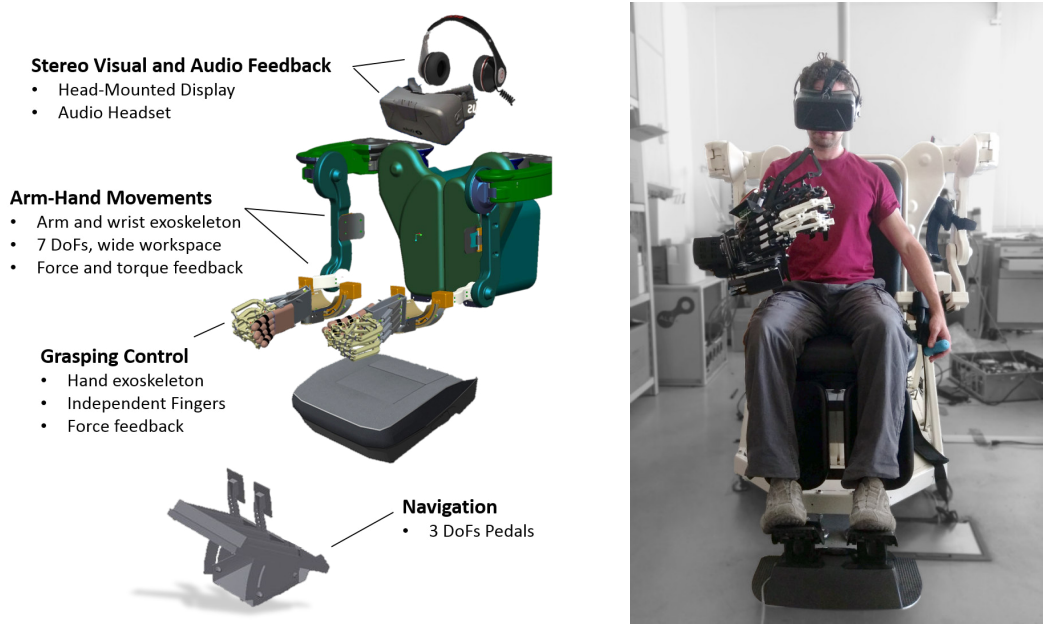

	\centering\begin{maybepreview}
	\includegraphics[height=0.5\columnwidth]{telepresenceStationScheme}~~~~~
	\includegraphics[height=0.5\columnwidth]{TelepresenceStationPercro}\end{maybepreview}
	\caption{The full-body telepresence suit, components and implementation.}
	\label{fig_telepresence_suit}
\end{figure}

An intuitive teleoperation interface is key to control a robot, as complex as Centauro, with all its capabilities. 
A full-body telepresence suit (see~\cref{fig_telepresence_suit}) allows for immersive control of the whole robot, especially for manipulation tasks. 
A high degree of intuition is reached by giving the operator the feeling of being present in the robot by perceiving the scene from the robot head perspective and moving the robot arms and hands as own limbs. 
This is realized through a head-mounted display for visual and acoustic situation awareness, pedals for basic locomotion control, and---most important---an upper-body exoskeleton. 
The latter enables the operator to transfer his upper body movements to the robot. 
Hence, a wide range of movement that covers most of the human workspace is required. 
The presented telepresence suit can reach about 90\% of the human's arm workspace without singularities. 
Moreover, the upper-body exoskeleton provides force feedback by displaying the interaction forces between the robot and its environment. 
Thus, it allows the operator to transfer his or her experience and capability of situation assessment into the scene. 
The upper-body exoskeleton consists of
\begin{itemize}[topsep=-5pt,noitemsep]
\item two upper limb active exoskeletons for the arms with four DoF each (see~\cref{sec:arm_exo}),
\item two active wrist exoskeletons with three DoF each (see~\cref{sec:wrist_exo}),
\item and an underactuated active hand orthosis with one DoF for each finger (see~\cref{sec:hand_exo}).
\end{itemize}

\subsubsection{Arm Exoskeleton}
\label{sec:arm_exo}
The dual arm exoskeleton is a mechanically compliant robotic manipulator that provides full support for shoulder and elbow rotation~\citep{pirondini2014evaluation}. 
It features an innovative and lightweight backdrivable transmission with electric actuators that are located behind the operator's back. 
The torque transmission from the actuators to the joints is achieved through idle pulleys and in-tension metallic tendons. 
This design allows for light moving parts which achieve a high dynamic performance due to low inertia. 
Furthermore, since neither actuators nor force-torque sensors are positioned on the joints or end-effectors, a slim design is possible that results in a large motion range and a low risk of either self-collision or collision with the human body. 
The total weight of the moving parts is only 3\,kg, of which about 2\,kg belong to the first two proximal links.

\begin{figure}
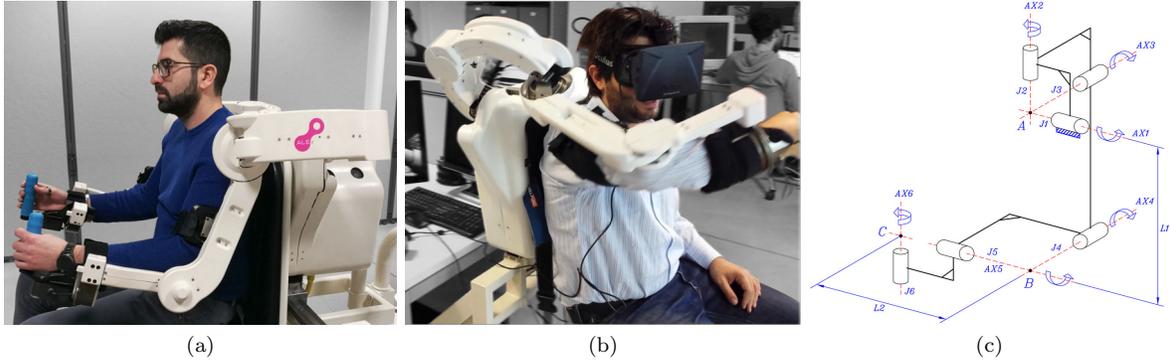

	\centering\begin{maybepreview}
	\subfloat[]{\includegraphics[height=0.26\columnwidth]{Alex}}\label{deviceCad}
	\subfloat[]{\includegraphics[height=0.26\columnwidth]{AlexSbraccio2}}\label{deviceReal}
	\subfloat[]{\includegraphics[height=0.26\columnwidth]{ArmKinematics}}\label{ArmKinematics}\end{maybepreview}
	\caption{(a),(b) The arm exoskeleton features four actuated DoFs (shoulder and elbow) in a wide operative workspace, (c) kinematic scheme.}
	\label{fig_arm_exoskeleton}
\end{figure}

However, the utilized metallic tendons show some compliant behavior which would result in inaccurate motion tracking or force transmission. 
Thus, two position sensors are used: one at the motor shaft, the other at the joint shaft, after the tendon transmission. 
The difference of the measured relative position is an indirect measurement of the tension of the tendon. 
Such method allows a higher degree of robustness regarding the force control stability to the variability of the human limb mechanical impedance. 
To assure a high degree of reliability, the sensory equipment of the arm exoskeleton consists of only two kinds of position sensors with a proven high robustness: optical incremental encoders are embedded in the electric actuators, and Hall effect-based absolute angular sensors are directly integrated in the robotic joints. 
\cref{fig_arm_exoskeleton} shows the manufactured arm exoskeleton and its kinematic design.

The arm exoskeleton kinematic is isomorphic to that of the human arm, \ie the device's axes of rotation are aligned with those of the operator's physiological articulations. 
It has the important property of presenting no singularities in the natural workspace of the upper limb. 
In order to leave enough space for the user’s shoulder without having any interference with the mechanical structure of the device, the implementation of the 2nd DoF utilizes a remote center of rotation mechanism (RCRM) based on a four-link pantograph. 
The ranges of motion for the first, second and third DoF are 120\textdegree, 95\textdegree~and 165\textdegree, respectively.

The major part of the control electronics has been custom built to be embedded into the mechanics of the device, in order to minimize the electrical wiring effort at the motors and sensors. 
The non-moving base link of the device (named the backpack) hosts the electric actuators (brushless DC motors) and the embedded control electronics and is connected to a rigid support structure that holds the power supply unit and the main control computing unit. 
A chair, having a manual hydraulic height regulation of the seat, allows users of different size to operate the device from a comfortable seating position.

\subsubsection{Wrist Exoskeleton}
\label{sec:wrist_exo}
For successful manipulation, the wrist movement is important, as it determines the orientation of the hand. 
Consequently, wrist motions are realized by an active wrist exoskeleton~\citep{buongiorno2018wres}. 
It covers the three rotational DoF of the wrist and provides respective torque feedback. 
The covered articulations are: forearm pronation/supination (PS), wrist flexion/extension (FE), and radial/ulnar deviation (RU).

\begin{figure}
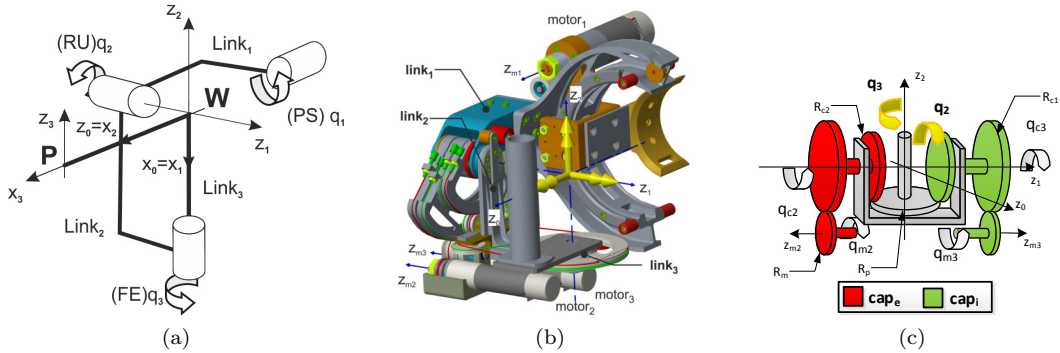

	\centering\begin{maybepreview}
	\subfloat[]{\includegraphics[height=0.25\textwidth]{Kinematics3}}\label{fig:kin_scheme} \hspace{0.5cm}
	\subfloat[]{\includegraphics[height=0.25\textwidth]{CAD_wrist}}\label{fig:cad_scheme} \hspace{0.5cm}
	\subfloat[]{\includegraphics[width=0.25\textwidth]{diff_kin_scheme2}}\label{fig:basic_diff_scheme}\end{maybepreview}
	\caption{Wrist exoskeleton design: (a) kinematic scheme, (b) CAD model, (c) schematic representation of the FE/RU differential transmission.}
	\label{fig:wr_schemes}
\end{figure}

\begin{figure}
	\centering\begin{maybepreview}
	\includegraphics[width=1\textwidth]{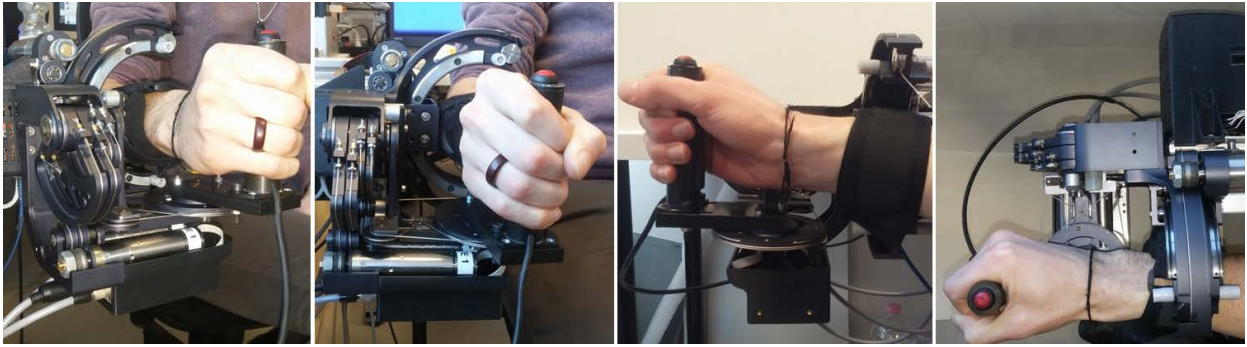}\end{maybepreview}
	\caption{The wrist exoskeleton worn by an operator. Left: Isolated test. Right: Mounted on the arm exoskeleton.}
	\label{fig:WRES_photos}
	\vspace{-0.2cm}
\end{figure}

The design is based on serial kinematics as shown in~\cref{fig:wr_schemes}. 
Again, actuators are located remotely at the non-moving parts and forces are transmitted via metallic tendons to obtain good dynamic behaviors. 
Main design requirements concern the lightness of the device, its easiness to be worn, and the need to have an inwards open structure to avoid collision with other parts of the telepresence suit during bimanual teleoperation tasks. 
The PS joint has been designed to improve the wearability of the wrist device by using an open curvilinear rail and rolling slider solution. 
FE and RU motions are jointly transmitted by two parallel actuators using a differential transmission as shown in~\cref{fig:basic_diff_scheme}. 
A passive regulation of the handle position along the PS axis allows to adapt the last link length to the user’s hand size. 
The total weight of the wrist exoskeleton is 2.9\,kg. \cref{fig:WRES_photos} shows the wrist exoskeleton worn by an operator.

\subsubsection{Hand Exoskeleton}
\label{sec:hand_exo}
\begin{figure}
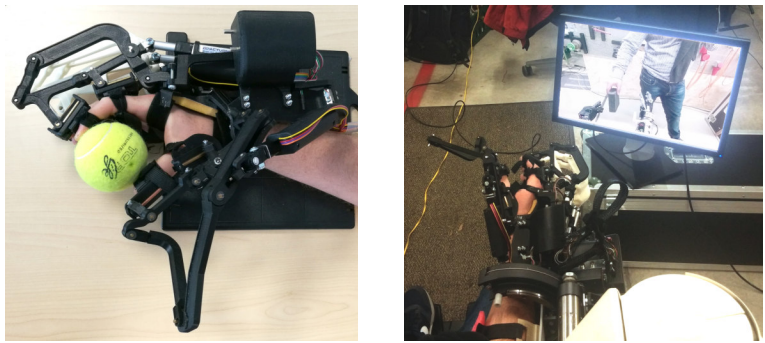

	\centering\begin{maybepreview}
	\subfloat{\includegraphics[height=0.27\textwidth]{hand-exos.jpg}} \hspace{0.5cm}
	\subfloat{\includegraphics[height=0.27\textwidth]{hand_application}}\end{maybepreview}
	\caption{The hand exoskeleton worn by an operator.}
	\label{fig:handexos}
\end{figure}

Finally, a lightweight, underactuated hand exoskeleton was designed for the teleoperation of grasping motions, providing independent finger motion tracking with force feedback. 
Each finger is connected to an individual underactuated parallel kinematic in two points which can adapt to the individual finger length as shown in~\cref{fig:handexos}. 
The hand exoskeleton uses two parameters to describe the pose of each of the four long fingers---index, middle finger, ring finger, and pinkie~\citep{sarac2017design}.
Those parameters do not directly map the fingers' joint positions. 
In addition, one parameter describes the spread between those four fingers. 
The thumb configuration is described with five parameters~\citep{gabardi2018design}. 

Force feedback is applied to each of the five fingers by an individual actuator. 
A total of five actuators is low in comparison to the DoF a human hand provides, but since a low device weight is an important design goal, a trade-off had to be found. 
Grasping teleoperation with force feedback from one actuator per finger showed promising results in previous experiments. 

The five parallel kinematic chains are mounted on a ground link which is rigidly connected with the back of the user's hand, as shown in~\cref{fig:handexos}. 
The ground link shape is modeled in order to follow the shape of the back of a human hand while a layer of foam between the hand surface and the ground link improves the adaptability of the system to different hands and increases comfort. 
The hand exoskeleton has been realized with 3D printed parts, resulting in a low-cost and lightweight device which only weighs about 350\,g. 
For the actuation, \emph{Firgelli L16} linear motors with 50\,mm stroke have been used. 
Moreover, the lack of magnetic hardware, the small case and the low weight of the actuators allow the motors to be placed close to each other to fit all the actuators on top of the hand. 
For sensing, potentiometers of the type \emph{Bourns 3382G-1-103G} are utilized in addition to the potentiometers that are integrated in the actuators.  
	
The utilized linear motors include a mechanical gearbox which prohibits backdriveability. 
However, backdriveability is fundamental to allow the operator to move its fingers. 
Consequently, force sensors are positioned between the ground link and the actuators. 
Those sensors measure the intentional forces applied by the user's fingers and a control algorithm is utilized to let the actuators follow the intended finger motion.

\pagebreak
\section{Advanced Environment Perception}
\label{sec:advanced_environment_perception}
The chosen sensor setup generates data of several types which differs in the direct usability of the data. 
While some sensor data requires processing, other percepts, such as foot camera images, can be directly shown to the operators.
The resulting representations are input to autonomous control functions or are visualized to the operators.

\subsection{Ground Contact Estimation}
\label{sec:ground_contact_estimation}
When navigating in challenging terrain, it is helpful to detect if a foot has ground contact. 
This helps the operator to assess the robot stability since established ground contact might be hard to recognize from camera images in some situations. 
In addition, such ground contact estimation enables reliable semi-autonomous stepping with few additional required environment knowledge.

Inputs are joint torques of the legs which are provided by the actuators.
Via forward dynamics, a 6D force vector is computed for each foot.
If, after gravity compensation, the foot exerts a force onto the ground, contact is detected.
This feature is \eg used within the semi-autonomous stepping controller.
Ground contact estimates are also visualized to the operators to facilitate robot stability assessment.

\subsection{Laser-based 3D Mapping and Localization}
\label{sec:3d_laser_scan_assembly}

\begin{figure}
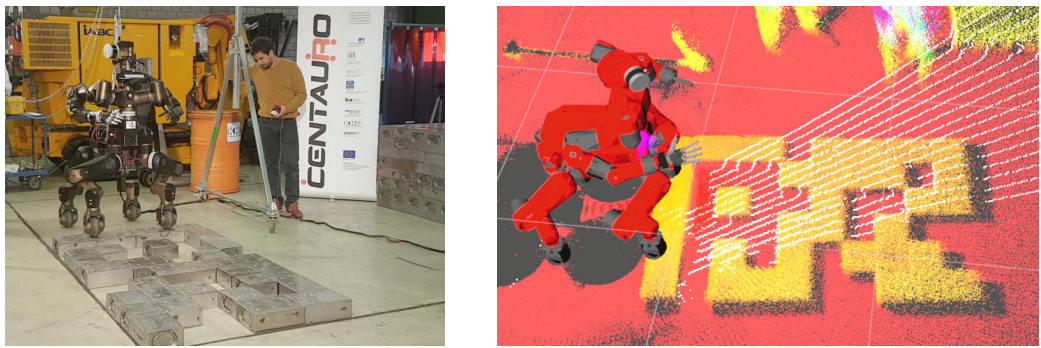

	\centering\begin{maybepreview}
	\includegraphics[height=4.5cm, trim=500 50 200 100, clip]{figures/figures_section4/centauro_on_step_field.png}~~~~~
	\includegraphics[height=4.5cm]{figures/figures_section4/laser_map_stepping_field.png}\end{maybepreview}
	\caption{Centauro overcoming a step field: Scenario (left), localized robot and registered point cloud color coded by height (right).}
	\label{fig:point_cloud}
\end{figure}

The 3D rotating \emph{Velodyne PUCK} laser scanner with spherical field-of-view provides about 300,000 range measurements per second with a maximum range of 100\,m. 
It is rotated at 0.1 rotations per second, resulting in a dense omnidirectional 3D scan per halve rotation every 5 seconds.
Slower rotation is possible if a higher angular resolution is desired. 
Laser range measurements are aggregated to a dense 3D map of the environment.
We use our local multiresolution surfel grid approach for this~\citep{Droeschel2017104}.
We compensate for sensor motion during acquisition by incorporating measurements of the IMU.

3D scans are aggregated in a robot-centric local multiresolution map by registering consecutive scans to a denser egocentric map. 
To allow for allocentric mapping of the environment, the resulting egocentric maps from different view poses form nodes in a pose graph. 
Graph edges connect these view poses and represent spatial constraints, which result from aligning these maps with each other. 
The global registration error is minimized using graph optimization. 
The resulting 3D map allows for localizing the robot in an allocentric frame which is used autonomous locomotion and manipulation panning but also to generate informative visualizations for the operators.
\cref{fig:point_cloud} shows an example of a generated point cloud and a localized robot.

\subsection{Object Segmentation}
\label{sec:perception:object_segmentation}
\label{sec:perception:semantic_segmentation}

To enable autonomous or semi-autonomous manipulation of the workspace, the Centauro robot has to detect and
determine the pose of useful objects in its environment, for example tools.
For the purpose of object segmentation, we train
a CNN-based model on a dataset of tools.
The main sensor for perceiving the workspace is the chest-mounted \emph{Kinect V2} sensor.
This sensor provides registered RGB color and depth estimates, which is especially useful for
pose estimation. A multi-frequency phase unwrapping algorithm improves the depth estimates~\citep{lawin2016efficient}.

\begin{figure}
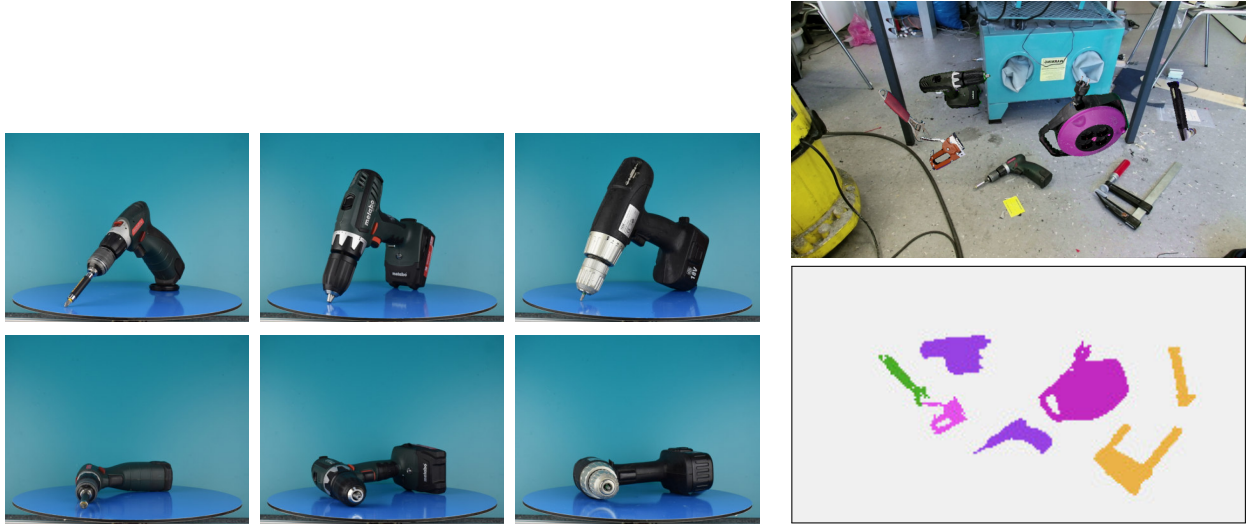

	\centering\begin{maybepreview}
	\begin{minipage}[b]{0.6\textwidth}
 		\setlength{\tabcolsep}{0pt}
 		\includegraphics[height=2.5cm,trim=70 40 60 30,clip]{figures/manipulation/object_perception/turntable/1_1.jpg}~
 		\includegraphics[height=2.5cm,trim=70 40 60 30,clip]{figures/manipulation/object_perception/turntable/2_1.jpg}~
 		\includegraphics[height=2.5cm,trim=70 40 60 30,clip]{figures/manipulation/object_perception/turntable/3_1.jpg}\vspace*{1ex}\\
 		\includegraphics[height=2.5cm,trim=70 40 60 30,clip]{figures/manipulation/object_perception/turntable/1_2.jpg}~
 		\includegraphics[height=2.5cm,trim=70 40 60 30,clip]{figures/manipulation/object_perception/turntable/2_2.jpg}~
 		\includegraphics[height=2.5cm,trim=70 40 60 30,clip]{figures/manipulation/object_perception/turntable/3_2.jpg}
	\end{minipage}~~~~~~
	\begin{minipage}[b]{0.39\textwidth}
		\includegraphics[height=3.4cm]{figures/manipulation/object_perception/turntable/torch_refinenet_vis_rgb.png}\vspace*{0.5ex}\\
 		\includegraphics[height=3.4cm, frame]{figures/manipulation/object_perception/turntable/torch_refinenet_vis_target.png}
	\end{minipage}\end{maybepreview}
 	\caption{Turntable capture and scene synthesis. Left: Different drills on the
 	turntable as captured by a DSLR camera.
 	Right: Synthetic training scene generated by inserting new objects.
 	The top image shows the resulting color image, the bottom shows
 	generated ground truth for training the segmentation.}
 	\label{fig:perception:synthesis}
\end{figure}

Our semantic segmentation pipeline directly produces pixel- (or point-)wise class labels.
Utilizing the recent RefineNet~\citep{lin2016refinenet} architecture, we create a representation of the input
image with both highly semantic information and high spatial resolution.
Based on this representation, a final convolutional layer with softmax transfer function outputs class probabilities.

Deep learning methods require large amounts of training data. We address this
problem by generating new training scenes using data captured from a turntable
setup. Automatically extracted object segments are inserted into pre-captured
scenes, as shown in~\cref{fig:perception:synthesis}.
For details on the capturing and scene synthesis pipeline, we refer
to~\citep{schwarz2018fast}.

\subsection{Pose Estimation}
\label{sec:perception:pose_estimation}
\label{sec:perception:posenet}

To facilitate robust autonomous grasping of objects, we need their 6D pose.
To this end, we developed a 5D pose predictor network for efficient pose estimation on
a rough scale.
Again, RGB-D data from the \emph{Kinect V2} serves as input.

\begin{figure}
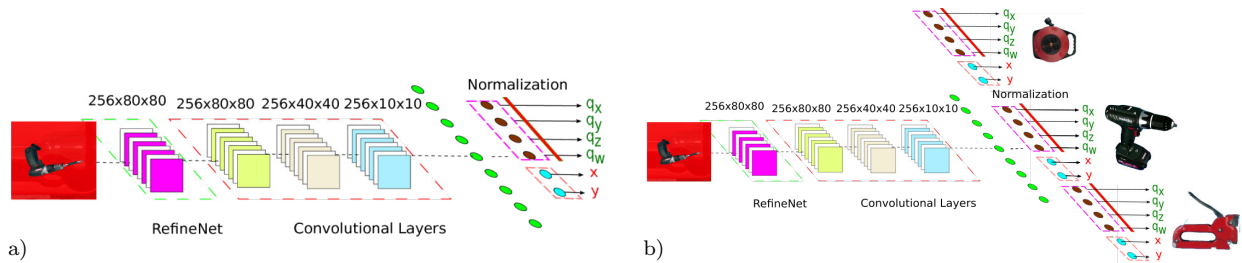

	\footnotesize\begin{maybepreview}  a)\hspace*{-2ex}\raisebox{-0\height}{\includegraphics[width=.48\textwidth]{figures/manipulation/object_perception/pose_single.png}}\hfill
  ~~b)\hspace*{-2ex}\raisebox{-2mm}{\includegraphics[width=.48\textwidth]{figures/manipulation/object_perception/pose_multi.png}		\vspace*{-5mm}}\end{maybepreview}
  	\caption{Pose estimation network architecture. a) Single-block output variant;
  	b) Multi-block output.}
  \label{fig:perception:posenet_architecture}
\end{figure}

For efficient pose prediction, we augmented the semantic segmentation pipeline (\cref{sec:perception:semantic_segmentation}) with an additional CNN to estimate the 5D object pose (rotational, and $x$ and $y$ of translation) from the RGB crops from the scene.
Based on the bounding boxes of detected object contours, those crops are extracted.
Pixels classified as non-object are pushed towards red to encode the segmentation results (see \cref{fig:perception:posenet_architecture}).
By providing the network this representation, it is able to focus on the specific object of which the pose should be determined.
The pretrained RefineNet network from \cref{sec:perception:semantic_segmentation} is used to extract features.
We extend the data acquisition pipeline described in~\citep{schwarz2018fast} to record turntable poses automatically and fuse captures with different object poses or different objects (see \cref{fig:perception:synthesis}).
This is used to generate the ground truth poses for network training.

We developed two different types of CNN architectures shown in~\cref{fig:perception:posenet_architecture}. 
The single-block output variant predicts six values (rotation represented as a unit quaternion and $x$ and $y$ of translation), 
whereas the multi-block output variant predicts these for each object category.
Our evaluation showed that the single-block variant performed slightly better in the presence of occlusion, and was thus chosen for use in the Centauro system.
The predicted 5D pose can be projected into 6D using the mean depth measurement $d$ in a window around the object center.
The 6D object pose is \eg input to the autonomous manipulation pipeline.

\section{Operator Interfaces}
\label{sec:operator_interfaces}
Although the considered environments are too dangerous for humans to work in, the human capabilities of situation assessment, mission planning, and task experience are key to a successful mission, \eg when maintaining a moon or planet base. 
The operator interfaces enable the operators to transfer these desired capabilities into the scene by giving them an awareness of the situation and enabling them to control the robot. 
Both require a communication infrastructure, since visual contact is not available.

Regarding locomotion and manipulation control, we aim at enabling the operators to solve as many previously known and unknown typical construction, maintenance, and exploration tasks as possible. 
Hence, we propose a set of control functions with advantages in different task classes to perform successful teleoperation. 
A key requirement is to utilize the whole range of the robot's kinematic capabilities while keeping the control itself intuitive and flexible to adapt to unforeseen situations. 
Different degrees of autonomy are used to fulfill this requirement.
Parts of the operator interfaces have already been described in~\citep{klamt2018supervised}.

\subsection{Communication}
We use a wired Ethernet connection or a standard IEEE 802.11ac 5\,GHz WiFi link for data transmission between the operator station and the robot, 
Depending on the type of data, different protocols are used.
The low-latency control and feedback messages for the exoskeleton use raw UDP packets.
All other communication is realized via \emph{ROS} topics.
These are either directly accessed using \emph{ROS} network transparency,
or encoded with FEC using the \texttt{nimbro\_network} developed for Momaro~\citep{schwarz2017nimbro} to increase robustness.
This is especially developed to handle communication with low-bandwidth, high latency, and package loss, which is often the case for space missions.
For extending the coverage, Centauro can carry a WiFi repeater and drop it at an appropriate location.

\subsection{Situation Awareness}
\label{sec:oi:situationawareness}

\begin{figure}
	\vspace{-0.1cm}
	\centering\begin{maybepreview}
\makeatletter
\newcommand\current{\the\tikz@lastxsaved,\the\tikz@lastysaved}
\makeatother

\begin{tikzpicture}[
    font=\sffamily\footnotesize,
    every node/.append style={text depth=.2ex},
	sensor/.style={rectangle,rounded corners,draw=black,fill=red!20,align=center,minimum width=2cm,minimum height=1cm},
    module/.style={rectangle,rounded corners,draw=black,fill=yellow!20,align=center,minimum width=2cm,minimum height=1cm},
    group/.style={rectangle,rounded corners,draw=black},
    extmodule/.style={module,dashed,fill=black!10},
    l/.style={font=\sffamily\scriptsize},
   simmodule/.style={module,fill=blue!20},
]

\node[sensor] (kinematic) at (1.2,-1) {Kinematic\\State};
\node[sensor,below=.2cm of kinematic] (kinect) {Kinect v2};
\node[sensor,below=.2cm of kinect] (laser) {3D Laser\\Scanner};
\node[sensor,below=.4cm of laser] (camera) {RGB\\Cameras};

\node[module,right=.5cm of kinect] (segmentation) {Segmentation};
\node[module,right=.5cm of segmentation] (pose) {6D Pose\\Estimation};

\node[module,right=.5cm of laser] (slam) {6D\\SLAM};
\node[module,right=.5cm of slam] (planning) {Locomotion\\Planning};

\coordinate[fit=(kinematic)(camera)] (vcenter);

\coordinate[below=2cm of camera] (below);
\node[module,anchor=north,minimum width=5cm, minimum height=3.2cm,rotate=90] (sim) at (vcenter-|9.5,-2.5) {};
\node[anchor=north,align=center] at (sim.east) {VEROSIM\\3D Simulation};

\node[simmodule] (rendering) at ($(sim)+(0.2,-0.2)$) {Rendering\\Framework};
\node[simmodule,above=.2cm of rendering] {Sensor\\Framework};
\node[simmodule,below=.2cm of rendering] {Rigid Body\\Framework};

\begin{scope}[-latex,label/.style={at end,anchor=south east,align=center,shift={(0,-0.1)},font=\sffamily\scriptsize}]

\draw (kinematic.east) -- (kinematic.east-|sim.north) node [label] {TF Tree};

\draw (kinect) -- (segmentation);
\draw (segmentation) -- (pose);
\draw (pose) -- (pose.east-|sim.north) node [label] {6D Pose\\+ Label};

\draw (laser) -- (slam);
\draw (slam) -- (planning);

\coordinate (c1) at ($(planning.east)+(0,0.2)$);
\draw (c1) -- (c1-|sim.north) node [label] {Path};

\coordinate (c2) at ($(planning.east)+(0,-0.2)$);
\draw (c2) -- (c2-|sim.north) node [label] {Height/Cost Map};

\coordinate (laserjoint) at ($(slam.east)!0.5!(planning.west)$);
\fill (laserjoint) circle[radius=1.5pt];
\draw (laserjoint) -- ++(0,-0.6) -- (\current -| sim.north) node [label] {Global Map};

\coordinate (laser2joint) at ($(laser.east)!0.5!(slam.west)$);
\fill (laser2joint) circle[radius=1.5pt];
\draw (laser2joint) -- ++(0,-1.0) -- (\current -| sim.north) node [label] {Point Cloud};

\draw (camera) -- (\current -| sim.north) node [label] {RGB Images};

\end{scope}

\node [rectangle,anchor=north,rotate=90,draw=black,fill=blue!20,minimum width=4.5cm] at (sim.north) {ROS Interface};

\node [module,fill=orange!20] (operator) at (vcenter-|15,5) {Operator};
\begin{scope}[-latex,label/.style={at end,anchor=south west,align=left,shift={(0,-0.1)},font=\sffamily\scriptsize}]

\draw (sim.south) ++(0,1.5) -| (operator) node [label,at start] {Immersive\\3D Visualization};
\draw[latex-] (sim.south) ++(0,-1.5) -| (operator) node [label,at start] {Control Input\\via Input Devices};

\end{scope}

\coordinate (labelheight) at (0,0);

\node at (labelheight-|kinematic) {Sensors};

\coordinate[fit=(segmentation)(pose)] (processing);
\node at (labelheight-|processing) {Perception \& Planning};

\node at (labelheight-|sim) {Simulation};

\end{tikzpicture}
\end{maybepreview}
	\caption{Pipeline for 3D simulation. Sensors are colored red and pipeline components yellow.}
	\label{fig:ctr:operator:overview}
	\vspace{-0cm}
\end{figure}
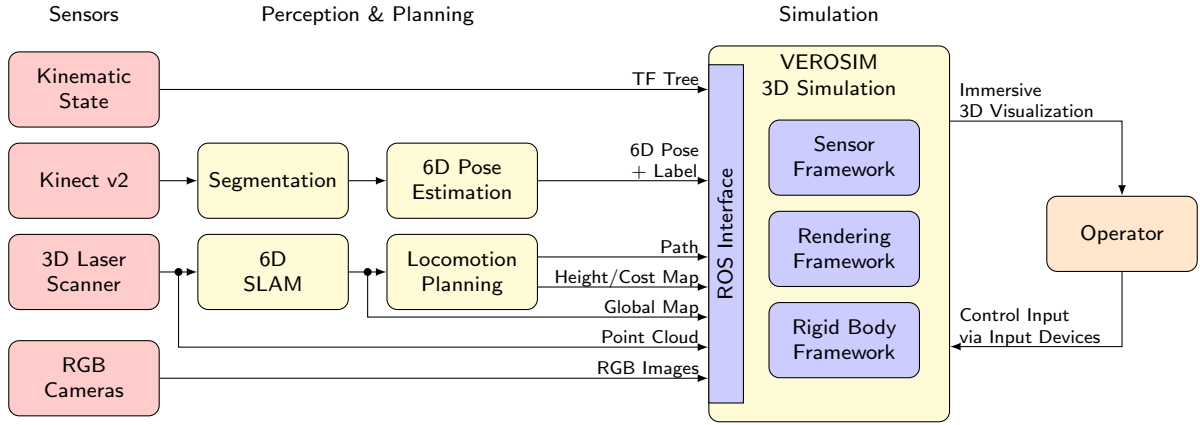

A simulation-based approach is used to generate user interfaces which can be displayed in an \mbox{\emph{HTC Vive}} head-mounted display or on arbitrary regular monitors.
Thus, we aim at providing an intuitive, semantically enriched visualization of data for the teleoperation of real robots in dynamic scenarios.
Based on a digital twin of the real system, we use 3D simulation in-the-loop to visualize the current state of the robot in its environment and add semantic information for the operators, to intuitively interact with the real robotic system.
This involves 
\begin{itemize}[topsep=0pt,noitemsep]
	\item the (planar) visualization of incoming data, raw and preprocessed, 
	\item the projection of data onto the place of occurrence, 
	\item the stereoscopic rendering and thus immersion of the operator, and 
	\item a modular combination and selection of all modalities suited for the given application or situation.
\end{itemize}
For a holistic incorporation of central modules, as well as interfaces to all system soft- and hardware components, we use \textit{VEROSIM} (Virtual Environments and Robotics Simulation System) with its integrated rendering, sensor, and rigid body frameworks as well as its \emph{ROS} interface (see~\cref{fig:ctr:operator:overview}).

\begin{figure}[b!]
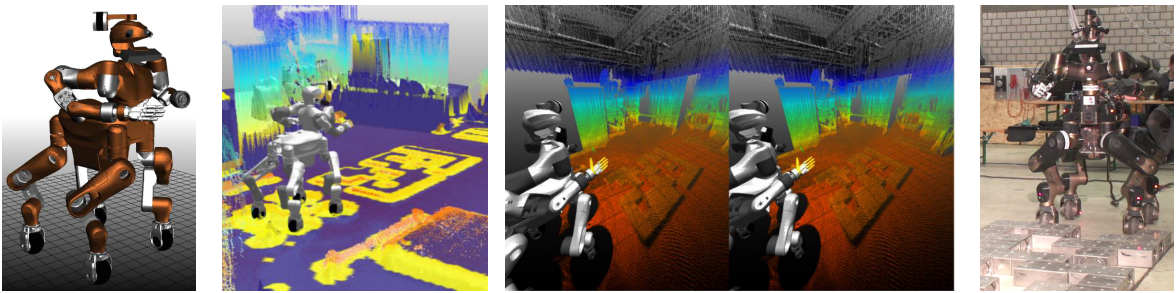

	\vspace{-0cm}
	\centering\begin{maybepreview}
	\includegraphics[height=3.8cm]{figures/operator_interface/RWTH_Centauro_Sim.jpg}~~
	\includegraphics[height=3.8cm]{figures/operator_interface/RWTH_Centauro_Mono.jpg}~
	\includegraphics[height=3.8cm]{figures/operator_interface/RWTH_Centauro_Stereo.jpg}~~
	\includegraphics[height=3.8cm]{figures/operator_interface/RWTH_Centauro_Robot.jpg}\end{maybepreview}
	\caption{l.: \textit{VEROSIM} provides a digital twin of the Centauro robot encompassing all links, joints, sensors etc., c.: 		third person view (standard and stereoscopic) on the current scene with a rigid body height map overlaid by a cost map 				accompanied by color-coded point cloud data, r.: image of the visualized scene, Centauro standing in front of a stepping 			field.}
	\label{fig:ctr:operator:02}
\end{figure}

\begin{figure}
	\centering\begin{maybepreview}
 	\includegraphics[width = 0.8\textwidth,page = 1, trim= 0 200 140 0,clip]{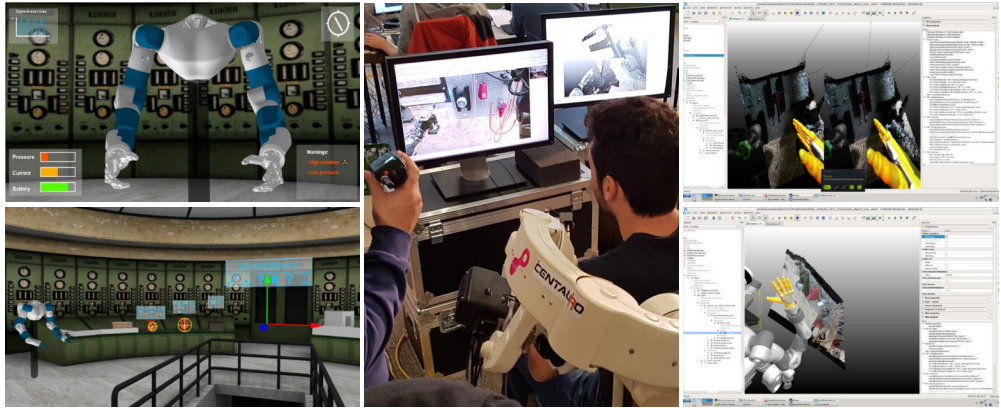}\end{maybepreview}
	\caption{Overview of visualization possibilities: t.l.: head-up display visualization of internal parameters, b.l.: projection 	of semantic object information via billboard visualizations, c.: first person operator using the exoskeleton with direct 			camera view and point cloud visualizations, t.r.: stereoscopic and b.r.: third person view on the scene during a manipulation 		task.}
	\label{fig:ctr:operator:01}
\end{figure}

Based on the raw sensor data (robot pose, point clouds and images), the digital twin of the robot, as well as perceived environment information can be visualized in simulation (see~\cref{fig:ctr:operator:02} l.,c.).
Besides the possibility to visualize raw data streams, the 3D simulation can also use preprocessed data from all software modules to generate dynamic environments based on the robot's perception.
Based on generated point clouds, rigid body height maps can be created and cost map visualizations can be overlaid (see~\cref{fig:ctr:operator:02} c.).
Relying on the semantic detection of objects with a given name and pose, the simulation can add the dimension of space to the data by projecting it onto the place of occurrence.
Consequently, a template-based model insertion can be used to insert a 3D model of the recognized object into the 3D scene which overcomes point cloud gaps  (see~\cref{fig:ctr:operator:01} b.l.).
Head-up display visualization of data can help each operator to create a unique, personalized and individually optimized user interface~\citep[see \cref{fig:ctr:operator:01},][]{cichon2017esm}.
We note that in our system, the environment is rendered on the operator side
and thus head movement of the operator can be tracked with very low latency
and immediately results in a changed viewpoint. This way, uncomfortable lags
are avoided even with high-latency data links to the robot.

Stereoscopic rendering encompasses all of the aforementioned aspects.
Incorporating the \textit{SteamVR} API into \textit{VEROSIM} enables the utilization of software modalities such as using stereoscopic headsets (see~\cref{fig:ctr:operator:02} t.r.). 
This enables an operator wearing a head-mounted display to see through the robot's eyes, increasing the immersion and the embodiment in simulation.
In addition, it is also possible to optimize the operator's view by either physically walking around in the scene or by moving the camera pose through the 3D scene via computer mouse or to predefined poses in the simulator (see~\cref{fig:ctr:operator:01} r.).
This leads to a wide set of opportunities to individually change and reposition the view on the scene for each task and to generally mediate and support robotic teleoperation via 3D simulation (see~\cref{fig:ctr:operator:01} c.).
Further details can be found in~\citet{cichon2017ssrr}.

\begin{figure}
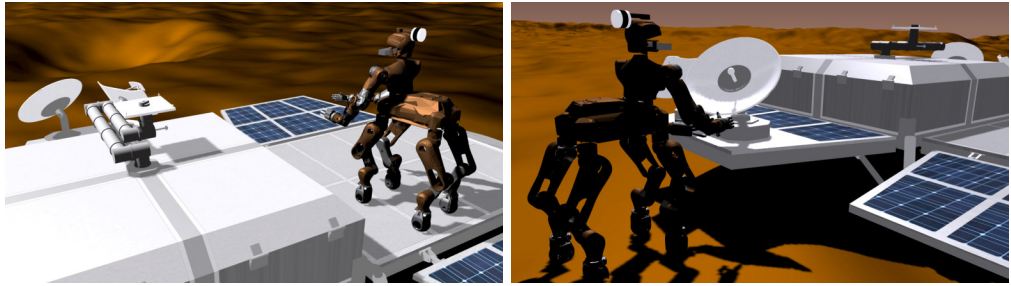

	\centering\begin{maybepreview}
	\includegraphics[width=0.4\textwidth]{figures/figures_section4/CentauroOnMars03_e}
	\includegraphics[width=0.4\textwidth]{figures/figures_section4/CentauroOnMars05_e}\end{maybepreview}
	\caption{Centauro in a simulated Mars mission.}
	\label{fig:centauro_on_mars}
	\vspace{-0cm}
\end{figure}

Although this optimized user interface in terms of visualization is the main focus in this contribution, the use of a 3D simulation backend comes along with additional opportunities.
During the development process of such complex systems, the simulator is used for most of the system components.
The implemented \emph{ROS} interface allows for the combination of internal and external frameworks and the use of the simulation framework as the central integration tool for all system components.
We \eg employed VEROSIM to test the teleoperation with the telepresence suit. 
The operator was able to grasp simulated objects and receive force feedback~\citep{cichon2016combining}.
Moreover, the digital twin can be used to test actions before they are executed by the real system and to predict action results.
\Cref{fig:centauro_on_mars} shows how VEROSIM is used to simulate a maintenance mission on the Mars with Centauro.
Additional optimization, prediction and logging can be accessed during or after operation to analyze the overall system performance~\citep{Atorf2014iSAIRAS,atorf2015flexible}.

\begin{figure}
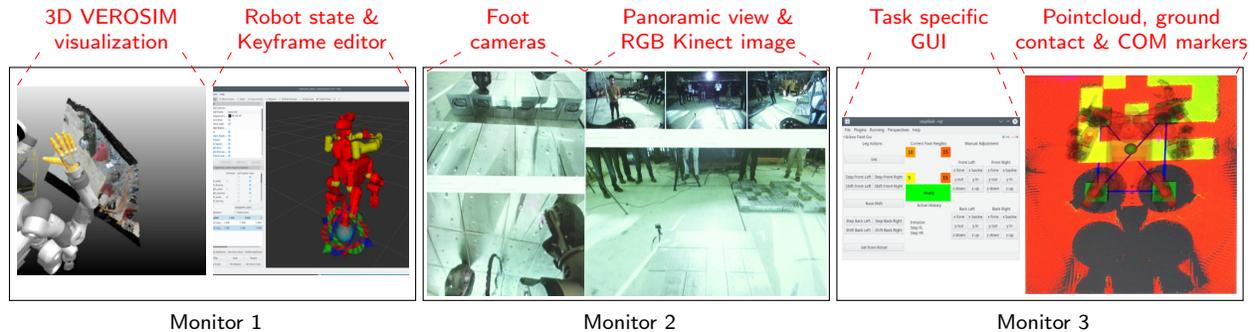

	\centering\begin{maybepreview}
\begin{tikzpicture}[
 	font=\sffamily\footnotesize,
    every node/.append style={text depth=.2ex},
	box/.style={rectangle, inner sep=0.5, anchor=west},
	line/.style={red, thick},
 	l/.style={font=\sffamily\scriptsize},
]

\draw (0.0,0.0) rectangle ++(5.4, 3.1);
\draw (5.5,0.0) rectangle ++(5.4, 3.1);
\draw (11.0,0.0) rectangle ++(5.4, 3.1);

\node[anchor=south west,inner sep=0] (image) at (0.1,0.3) {\includegraphics[width=2.5cm]{figures/figures_section4/verosim_manipulation.png}};

\node[anchor=south west,inner sep=0] (image) at (2.7,0.3) {\includegraphics[width=2.6cm,height=2.5cm,clip, trim=20 0 60 0]{figures/figures_section4/keyframe_editor.jpg}};

\node[anchor=south west,inner sep=0] (image) at (5.55,0.02) {\includegraphics[width=5.3cm,height=3cm]{figures/figures_section4/all_cameras.jpg}};

\node[anchor=south west,inner sep=0] (image) at (13.5,0.05) {\includegraphics[width=2.8cm,height=2.9cm]{figures/figures_section4/pointcloud.jpg}};

\node[anchor=south west,inner sep=0] (image) at (11.07,0.5) {\includegraphics[width=2.35cm,height=1.9cm]{figures/figures_section4/step_field_gui.png}};

\node[box, align=center, color=red](laser_scanner) at(0.45,3.6){3D VEROSIM\\visualization};
\draw[red,dashed](0.4,3.3) -- (0.1,2.85);
\draw[red,dashed](2.3,3.3) -- (2.6,2.85);
\node[box, align=center, color=red](laser_scanner) at(3.0,3.6){Robot state \&\\Keyframe editor};
\draw[red,dashed](2.95,3.3) -- (2.7,2.85);
\draw[red,dashed](5.05,3.3) -- (5.3,2.85);

\node[box, align=center, color=red](laser_scanner) at(8.1,3.6){Panoramic view \&\\RGB Kinect image};
\draw[red,dashed](8.05,3.3) -- (7.65,3.05);
\draw[red,dashed](10.5,3.3) -- (10.9,3.05);

\node[box, align=center, color=red](laser_scanner) at(6.1,3.6){Foot\\cameras};
\draw[red,dashed](6,3.3) -- (5.55,3.05);
\draw[red,dashed](7.2,3.3) -- (7.65,3.05);

\node[box, align=center, color=red](laser_scanner) at(13.35,3.6){Pointcloud, ground\\contact \& COM markers};
\draw[red,dashed](13.35,3.3) -- (13.5,3.0);
\draw[red,dashed](16.45,3.3) -- (16.3,3.0);

\node[box, align=center, color=red](laser_scanner) at(11.4,3.6){Task specific\\GUI };
\draw[red,dashed](11.45,3.3) -- (11.1,2.45);
\draw[red,dashed](13.05,3.3) -- (13.4,2.45);

\node[box, align=center](laser_scanner) at(2.1,-0.3){Monitor 1};
\node[box, align=center](laser_scanner) at(7.6,-0.3){Monitor 2};
\node[box, align=center](laser_scanner) at(13.1,-0.3){Monitor 3};

\end{tikzpicture}
\end{maybepreview}
	\vspace{-1em}
	\caption{Environment and robot state visualization for the support operators.}
	\label{fig:support_operator_monitors}
	\vspace{0cm}
\end{figure}

Besides the main operator sitting in the telepresence suit, there are additional operators who
\begin{itemize}[topsep=-5pt,noitemsep]
	\item can control the robot via control interfaces other than the telepresence suit,
	\item provide the required visualizations for the operator in the telepresence suit, and
	\item are responsible for keeping an overview over the whole situation while the operator in the telepresence suit might focus on task details.
\end{itemize}
Hence, those operators need a wide range of visualizations. 
We provide these by displaying processed and unprocessed data from several sensors on multiple monitors, as shown in~\cref{fig:support_operator_monitors}.
A panoramic view from the robot head perspective is helpful for general scene understanding.
Two RGB cameras under the robot base provide views which are arranged to enable a detailed assessment for the terrain under the robot base facilitating a save stepping locomotion operation (\cref{fig:support_operator_monitors}).
Moreover, results of the ground contact estimation are visualized by a colored marker for each foot. 
Together with the visualized robot center of mass (CoM), this helped assessing robot stability.
3D \emph{VEROSIM} visualizations further increased the scene understanding, especially in manipulation tasks where occluded areas are compensated by the simulation-based approach.
Several control GUIs were developed for task-specific robot control.

\subsection{Locomotion Control}
\label{sec:locomotion}
The Centauro lower body provides a wide range of locomotion capabilities which require suitable interfaces to be controlled efficiently. 
Interfaces for omnidirectional driving control are a joystick and a pedal controller (\cref{sec:joystick_pedal}). 
Leg movements can be controlled by a keyframe editor (\cref{sec:keyframe_editor}). 
A higher degree of autonomy is reached by utilizing a semi-autonomous stepping controller (\cref{sec:stepping_controller}). 
Finally, we developed a hybrid driving-stepping locomotion planner which autonomously plans and executes locomotion to a goal
pose specified by an operator using \emph{VEROSIM}, as described in \cref{sec:locomotion_planner}.

\subsubsection{4D Joystick \& 3D Pedal Controller}
\label{sec:joystick_pedal}
A \emph{Logitech Extreme 3D Pro} joystick with four axes is employed to control omnidirectional driving. 
The velocity components of the robot base $v_x$, $v_y$ and $v_\theta$ are mapped to the three corresponding joystick axes. 
Foot specific velocities and orientations are derived from this robot base velocity as described in~\cref{sec:path_execution}. 
All three velocity components are jointly scaled by the joystick throttle controller.

\begin{figure}
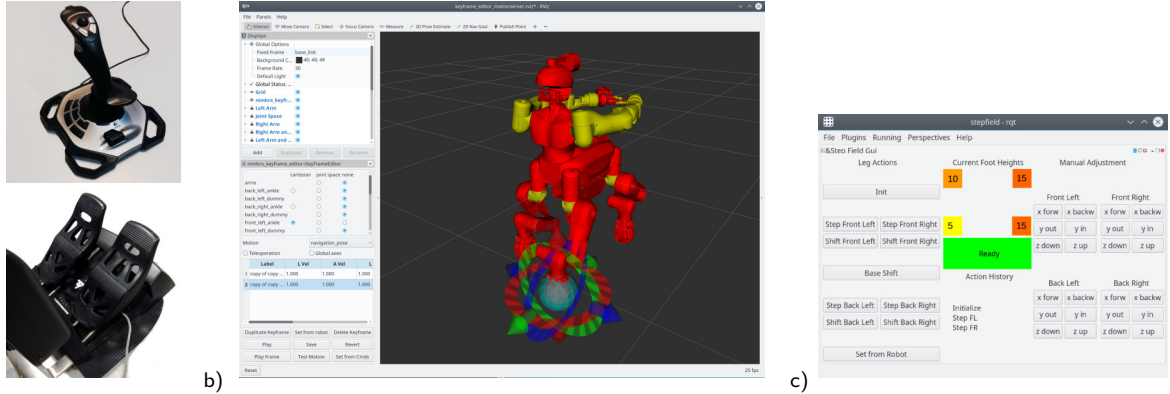

	\centering\begin{maybepreview}
\begin{tikzpicture}[
 	font=\sffamily\footnotesize,
    every node/.append style={text depth=.2ex},
	box/.style={rectangle, inner sep=0.5, anchor=west},
	line/.style={red, thick},
 	l/.style={font=\sffamily\scriptsize},
]

\node[anchor=south west,inner sep=0] (image) at (1.2,2.6) {\includegraphics[height=2.4cm]{figures/figures_section4/joystick1_e.jpg}};
\node[anchor=south west,inner sep=0] (image) at (1.2,0) {\includegraphics[height=2.5cm, trim=0 0 0 0, clip]{figures/figures_section4/pedals_2_e.jpg}};
\node[anchor=south west,inner sep=0] (image) at (4.3,0) {\includegraphics[height=5cm,clip,trim=0 1 0 0]{figures/figures_section4/keyframe_editor.jpg}};
\node[anchor=south west,inner sep=0] (image) at (12,0) {\includegraphics[height=3.5cm]{figures/figures_section4/step_field_gui.png}};

\node[box, align=left] at(0.8,0){a)};
\node[box, align=left] at(3.8,0){b)};
\node[box, align=left] at(11.6,0){c)};

\end{tikzpicture}
\end{maybepreview}
	\caption{Locomotion control interfaces: a) 4D joystick and 3D pedal controller, b) keyframe editor, c) semi-autonomous 				stepping controller GUI.}
	\label{fig:locomotion_control}
\end{figure}

The same three robot base velocity components are also mapped to a 3-axis \emph{Thrustmaster TFRP Rudder} pedal controller.
In addition to individual throttles for each of the two pedals, it also allows for changing the pedal's longitudinal position relative to each other which results in a third DoF.
The pedal controller is less intuitive than the joystick but can be used by the teleoperator in the telepresence suit.
Both devices are shown in~\cref{fig:locomotion_control}a.

\subsubsection{Keyframe Editor}
\label{sec:keyframe_editor}
Leg motions can be controlled by a keyframe editor~\citep{schwarz2016supervised}.
It allows for the control of joint groups (\eg the front left leg) in joint space or Cartesian end-effector space live during the mission or through predefined keyframes.
Keyframes can also be sequenced to motions. 
The GUI provides a graphical interface in which joints and end-effectors can be configured using interactive markers and the computer mouse (\cref{fig:locomotion_control}b). 
A numerical configuration is also possible.
Finally, robot motions are generated by interpolating between given keyframes. 

\subsubsection{Semi-autonomous Stepping Controller}
\label{sec:stepping_controller}
To efficiently control stepping locomotion in irregular terrain, we developed a semi-autonomous controller.
It provides a set of stepping and stepping-related motions that can be triggered by the operator.
The available motions are: step with a chosen foot, drive a chosen foot forward, and shift the robot base forward. 
If a stepping motion is triggered, the robot shifts its base longitudinally and laterally and rolls around its longitudinal axis to establish a stable stepping pose.
The stepping foot is then lifted, extended by a given length and lowered. 
The latter stops as soon as ground contact is detected (see~\cref{sec:ground_contact_estimation}).
Hence, the robot automatically adapts to the ground structure.
Motions are represented as sequences of keyframes, as described for~\cref{sec:keyframe_editor}, and can be executed by the motion player.

The controller is operated through a GUI which provides buttons to trigger the described motions for an individual foot (\cref{fig:locomotion_control} c).
Additional buttons allow the operator manual foot movement in Cartesian space, which is helpful to apply minor corrections.
Furthermore, the GUI visualizes detected terrain heights under the individual feet and a history of the last triggered motions which helps to follow repetitive motion sequences.

\subsubsection{Motion Execution}
\label{sec:motion_player}
Motions from any source need to be transformed to joint space trajectories to be executable by the robot.
We use a keyframe interpolation method which was originally developed for Momaro~\citep{schwarz2017nimbro}.
The interpolation system generates smooth joint trajectories obeying velocity and acceleration constraints set per keyframe.
Input are keyframes consisting of joint space or 6D Euclidean space poses for each of the robot limbs.

\subsection{Manipulation Control} 
\label{sec:manipulation_control}
The Centauro system possesses several degrees of autonomy to control manipulation. 
The upper body can be controlled in joint or Cartesian space or by executing keyframe motions with the keyframe editor which is described in~\cref{sec:keyframe_editor}. 
Furthermore, manipulation can be controlled via the upper-body exoskeleton which mimics the operator's behavior and provides force feedback (\cref{sec:telemanipulation_exo}). 
Moreover, precise control of the wrist pose is provided by a 6D input device (\cref{sec:6d_mouse}). 
Finally, we propose an autonomous grasping functionality which is described in~\cref{sec:autonomous_grasping}.

\subsubsection{Telemanipulation by the Upper-body Exoskeleton}
\label{sec:telemanipulation_exo}
The full-body telepresence station can be used to control the robot through an advanced teleoperation system with force feedback. 
By means of the bilateral and full upper limb exoskeleton, the operator directly controls the pose and the applied forces of the upper limbs of the robot. 
Different teleoperation control architectures have been implemented for the arm-wrist segments of the exoskeleton and for the hand grasping. 

\begin{figure}
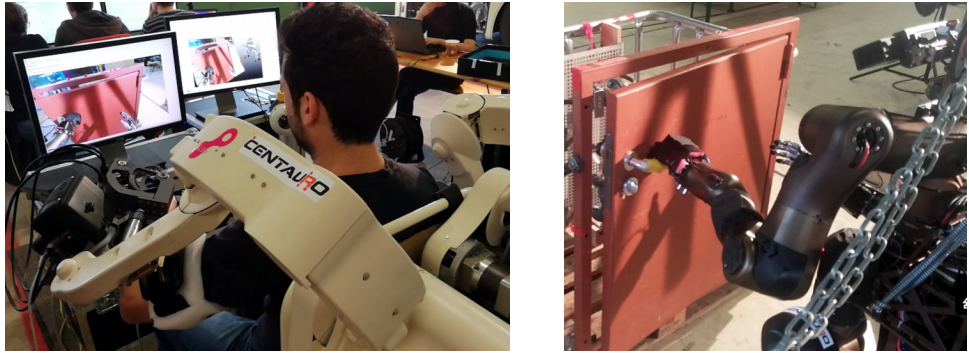

	\centering\begin{maybepreview}
	\includegraphics[height=0.28\columnwidth]{doorteleoperation_master} \hspace{0.5cm}
	\includegraphics[height=0.28\columnwidth]{doorteleoperation_slave}\end{maybepreview}
	\caption{An operator opening a door teleoperating Centauro with the upper-body exoskeleton.}
	\label{fig:handexos_teleoperation}
\end{figure}

Regarding teleoperation of the proximal segments of the upper limb, the operator's arm and wrist pose are measured by the correspondent parts of the exoskeleton and a 6D end-effector pose is computed. 
This pose is then transferred to the robot and applied to the arm using inverse kinematics. 
A direct mapping of joint configurations from the teleoperator to the robot is not helpful since their kinematic structures are not identical. 
Force feedback is generated by transferring measured 6D force-torque vectors from the force-torque sensor between right arm and Schunk hand to the exoskeleton and mapping them to joint torques using inverse dynamics.
\cref{fig:handexos_teleoperation} shows an operator using the upper-body exoskeleton to open a door.

For the teleoperation of grasping motions, the right-hand exoskeleton is connected with the anthropomorphic Schunk hand mounted at the right robot arm. 
Again, a direct joint mapping of control commands and force feedback is not possible due to differences in the kinematic concept, variable operator hand sizes and underactuation for both the hand exoskeleton and the Schunk hand.
To transfer finger and hand motions and applied forces from the operator's hand to the robot hand, we utilize the local admittance control of the hand exoskeleton and the local impedance control embedded in the Schunk hand.
The hand exoskeleton estimates angular position references of the operator's hand. Those are sent to the proportional-derivative (PD) position control system of the Schunk hand. 
The embedded PD control converts position references into actuator torques proportional to the position error.
Hence, once in contact with an object, the operator can increase grasping forces by further closing his or her fingers.

The mapping of positions and forces differs between fingers. 
Thumb, ring finger and pinkie are operated on the basis of a single DoF each on the operator and robot side, estimating the overall closing/opening of the finger. 
Index and middle finger are position controlled by the operator using two DoF, matching the rotation of the first and second phalanxes of each finger.
The Schunk hand provides this additional DoF for those two fingers which results in a higher grasping precision.
In addition, the spread between the four long fingers and the rotation of the thumb's opening/closing plane (abduction/adduction) is controlled. 

\begin{figure}
	\centering\begin{maybepreview}
	\includegraphics[width=0.94\columnwidth]{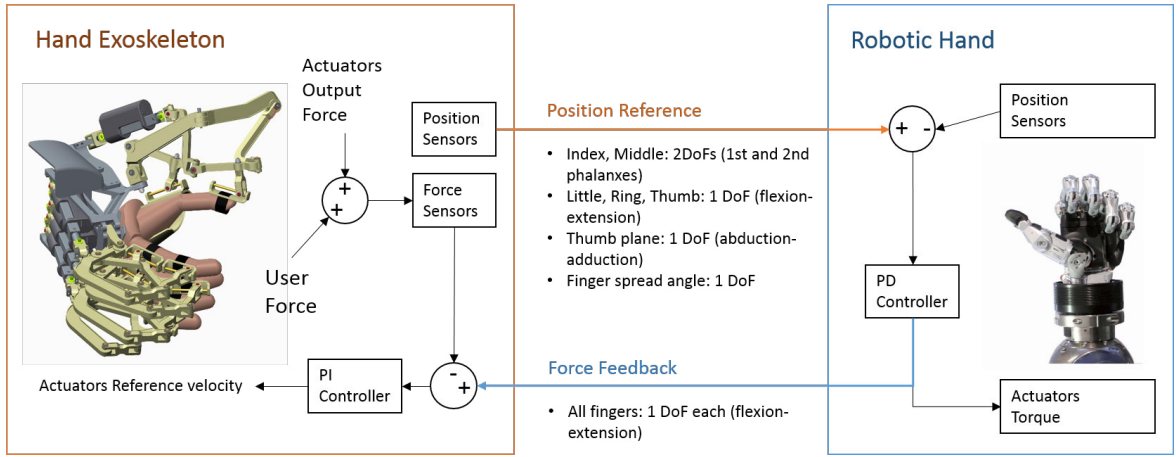}\end{maybepreview}
	\caption{Teleoperation scheme between the Hand Exoskeleton and the Schunk robotic hand.}
	\label{fig:handexos_teleoperation_scheme}
\end{figure}

To provide force feedback, torques applied by the Schunk hand are sent to the local admittance control of the hand exoskeleton and applied to the operator's hand as force feedback. 
The mapping in this direction is different since the exoskeleton provides less actuators than sensors.
Each finger receives a single DoF feedback force which is applied through the adaptive and underactuated mechanism of the hand exoskeleton. 
For the finger spread, force feedback is not applicable since no correspondent actuator exists in the hand exoskeleton. 
\cref{fig:handexos_teleoperation_scheme} visualizes the control architecture between hand exoskeleton and Schunk hand.

To control the 1-DoF SoftHand at the left robot arm, the left side of the exoskeleton is equipped with a lever-shaped 1-DoF grasping controller instead of a hand exoskeleton. Its position is mapped to the SoftHand actuator and the measured force at the robot hand is transferred to the controller as force feedback.

\subsubsection{6D Input Device for Wrist Control}
\label{sec:6d_mouse}

\begin{figure}
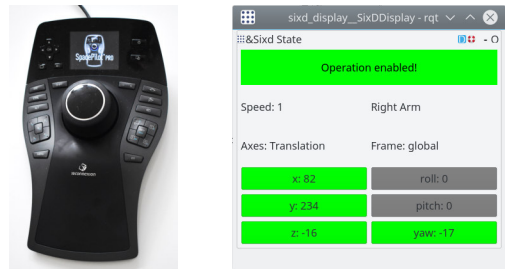

 	\centering\begin{maybepreview}
 	\includegraphics[height=3.5cm]{6D_mouse.png} \hspace{0.5cm}
 	\includegraphics[height=3.5cm]{sixd_gui.png}\end{maybepreview}
 	\caption{6D input device (l.) and corresponding GUI (r.) for dexterous wrist control.}
 	\label{fig:6D_mouse}
\end{figure}

A \emph{3DConnexion SpacePilot Pro} 6D input device and a corresponding GUI provide a teleoperation interface for dexterous wrist control (see~\cref{fig:6D_mouse}).
The developed interface establishes the connection between the device and the motion player mentioned in~\cref{sec:motion_player}.
The 6D \emph{SpacePilot} joystick movement from the operator is streamed as a desired 6D end-effector pose to the motion player which interpolates from the current to the desired pose and executes the motion.
The GUI can be used to easily adjust the following control parameters: 
\emph{End-effector} (a wrist for arm control or an ankle for leg control), 
\emph{Reference frame} (end-effector frame, robot base frame, or a custom frame), 
\emph{Enabled axes} (each translational and rotational axis can be enabled/disabled so the user input on this axis is considered/ignored), and
\emph{End-effector speed}. 
All control parameter can also be changed using buttons on the input device.
This interface is well suited whenever precise motions along certain axes are required.
This is the case for manipulation tasks where \eg an arm needs to be moved along a plane surface or an object needs to be turned around a specified axis.

Note that switching control authority from the exoskeleton to the 6D input device
is easily possible. The other direction induces much harder problems, since
end-effector location is always directly (and absolutely) coupled to hand position
in the exoskeleton.

\section{Hybrid Driving-Stepping Locomotion Planning}
\label{sec:locomotion_planner}
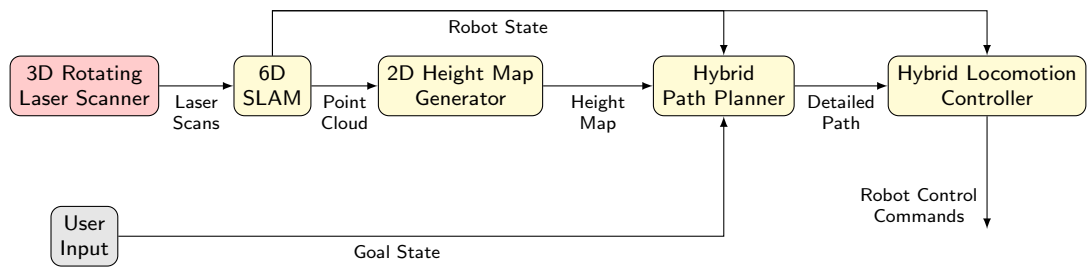
\begin{figure}[bth]
	\centering\begin{maybepreview}
\begin{tikzpicture}[
    font=\sffamily\footnotesize,
    every node/.append style={text depth=.2ex},
	sensor/.style={rectangle,rounded corners,draw=black,fill=red!20,align=center},
    module/.style={rectangle,rounded corners,draw=black,fill=yellow!20,align=center},
    group/.style={rectangle,rounded corners,draw=black},
    extinput/.style={sensor,fill=black!10},
    l/.style={font=\sffamily\scriptsize},
]

\node[sensor] (laser) {3D Rotating\\Laser Scanner};
\node[extinput](user) at($(laser)+(0,-2cm)$) {User\\Input};

\node[module](slam) at($(laser)+(2.5,0cm)$) {6D\\SLAM};
\node[module](height_map_generator) at($(slam)+(2.5,0cm)$) {2D Height Map\\Generator};
\node[module](path_planner) at($(height_map_generator)+(3.5,0cm)$) {Hybrid\\Path Planner};
\node[module](controller) at($(path_planner)+(3.5,0cm)$) {Hybrid Locomotion\\Controller};

\coordinate(path_planner_in1) at($(path_planner.south)+(-0.3,0cm)$);
\coordinate(path_planner_in2) at($(path_planner.south)+(+0.3,0cm)$);
\coordinate(final_out) at($(controller.south)+(0,-1.5cm)$);

\draw[-latex]   (laser) -- (slam) node [l,midway,below,align=center] {Laser\\Scans};
\draw[-latex]   (user) -|(path_planner) node [l,pos=0.23, below] {Goal State};
\draw[-latex]   (slam) -- (height_map_generator) node [l,pos=0.7, below,midway,align=center] {Point\\Cloud};
\draw[-latex]   (height_map_generator) -- (path_planner) node [l,midway, below,align=center] {Height\\Map};
\draw[-latex]   (path_planner) -- (controller) node [l,midway, below,align=center] {Detailed\\Path};
\draw[-latex]   (controller) -- (final_out) node [l,pos=0.8, left,align=center] {Robot Control\\Commands};
\draw[-latex]   (slam) -- ++(0,1) -| (path_planner) node [l,pos=0.25,below] {Robot State};
\draw[-latex]   (slam) -- ++(0,1) -| (controller) node [l,pos=0.84, left] {};

\end{tikzpicture} %
\end{maybepreview}
	\vspace*{-1ex}
	\caption{Overview of the pipeline for locomotion planning. Sensors are colored red, pipeline components yellow, and other 			inputs are colored gray.}
	\label{fig:locomotion_planning_overview}
\end{figure}

Autonomous locomotion planning and execution is a promising approach to increase locomotion speed and safety, to lower the operator's cognitive load, to execute locomotion under bad data transmission between the operators and the robot, and to keep the flexibility to control the robot in unknown tasks and unforeseen situations. 
The only required operator input is a desired robot goal state. 
The locomotion planning pipeline is visualized in~\cref{fig:locomotion_planning_overview}. 
Laser scanner point clouds (\cref{sec:3d_laser_scan_assembly}) are processed to cost maps that represent the environment (\cref{sec:planning_environment_representation}). 
A search-based planning approach uses this environment representation to generate plans (\cref{sec:search_based_planning}). 
Due to the unique hardware design and the many DoF of the robot platform, this planning approach is novel since it merges planning approaches for both driving-based and stepping-based locomotion and is capable of handling the respective complexity.
To handle planning queries for large environments, the approach is extended to plan on multiple levels of abstraction (\cref{sec:planning_abstraction}).
Finally, a controller executes these paths and outputs robot control commands (\cref{sec:path_execution}). 

\subsection{Environment Representation}
\label{sec:planning_environment_representation}
\begin{figure}
	\centering\begin{maybepreview}
\begin{tikzpicture}[
 	font=\sffamily\footnotesize,
    every node/.append style={text depth=.2ex}
]

\node[anchor=south west,inner sep=0] (image) at (0,0) {\includegraphics[width=\textwidth]{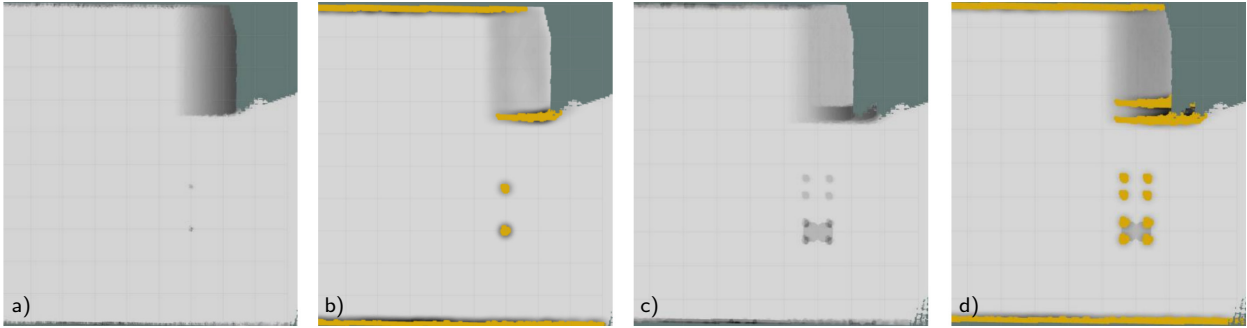}};
\node[] at (0.25,0.3) {a)};
\node[] at (4.41,0.3) {b)};
\node[] at (8.6,0.3) {c)};
\node[] at (12.85,0.3) {d)};

\end{tikzpicture}
\end{maybepreview}
	\vspace*{-2ex}
	\caption{Environment representation: The map shows two walls, a ramp and two poles of different height. a) Heights, b) Foot 		costs, c) Body costs, d) Pose costs.}
	\label{fig:environment_representation}
\end{figure}

A 2D height map is generated from the registered point clouds and serves as the environment representation. 
Height values are processed to foot and body costs. 
Foot costs describe the cost to place an individual foot at a given position in the map. 
They include information about the terrain surface and obstacles in the vicinity. 
Body costs describe the costs to place the robot base in a given configuration in the map. 
They include information about obstacles and the terrain slope under the robot. 
Foot costs of all four feet and body costs are combined to pose costs which describe the costs to place the whole robot in a given configuration on the map (see~\cref{fig:environment_representation}).

\subsection{Search-based Planning}
\label{sec:search_based_planning}
\begin{figure}
	\centering\begin{maybepreview}
\begin{tikzpicture}[
 	font=\sffamily\footnotesize,
    every node/.append style={text depth=.2ex},
	l/.style={font=\sffamily\scriptsize},
]

\definecolor {robot_base}{RGB}{80, 130, 255};

\draw (2.75,0.2) -- ++(0, 3.5);
\draw[thick] (5.5,0.2) -- ++(0, 3.5);
\draw (8.25,0.2) -- ++(0, 3.5);
\draw (11.0,0.2) -- ++(0, 3.5);
\draw (13.75,0.2) -- ++(0, 3.5);

\coordinate(robot_straight) at(0.9,1.8);
\fill[robot_base] (robot_straight) rectangle ++(1,0.6);
\fill[red] ($(robot_straight) + (-0.1,-0.25)$) rectangle ++(0.2,0.2);
\fill[red] ($(robot_straight) + (0.95,-0.25)$) rectangle ++(0.2,0.2);
\fill[red] ($(robot_straight) + (-0.1,0.65)$) rectangle ++(0.2,0.2);
\fill[red] ($(robot_straight) + (0.95,0.65)$) rectangle ++(0.2,0.2);
\fill[red] ($(robot_straight) + (0.7,0.3)$) circle (0.07);

\draw[gray] ($(robot_straight) + (-0.25,-0.75)$) -- ++(0,2);
\draw[gray] ($(robot_straight) + (0.05,-0.75)$) -- ++(0,2);
\draw[gray] ($(robot_straight) + (0.35,-0.75)$) -- ++(0,2);
\draw[gray] ($(robot_straight) + (0.65,-0.75)$) -- ++(0,2);
\draw[gray] ($(robot_straight) + (0.95,-0.75)$) -- ++(0,2);
\draw[gray] ($(robot_straight) + (1.25,-0.75)$) -- ++(0,2);

\draw[gray] ($(robot_straight) + (-0.5,-0.45)$) -- ++(2,0);
\draw[gray] ($(robot_straight) + (-0.5,-0.15)$) -- ++(2,0);
\draw[gray] ($(robot_straight) + (-0.5,0.15)$) -- ++(2,0);
\draw[gray] ($(robot_straight) + (-0.5,0.45)$) -- ++(2,0);
\draw[gray] ($(robot_straight) + (-0.5,0.75)$) -- ++(2,0);
\draw[gray] ($(robot_straight) + (-0.5,1.05)$) -- ++(2,0);

\draw[-latex] ($(robot_straight) + (0.5,0.3)$) -- ++(0,0.3);
\draw[-latex] ($(robot_straight) + (0.5,0.3)$) -- ++(0.3,0);
\draw[-latex] ($(robot_straight) + (0.5,0.3)$) -- ++(-0.3,0);
\draw[-latex] ($(robot_straight) + (0.5,0.3)$) -- ++(0,-0.3);
\draw[-latex] ($(robot_straight) + (0.5,0.3)$) -- ++(0.3,0.3);
\draw[-latex] ($(robot_straight) + (0.5,0.3)$) -- ++(-0.3,0.3);
\draw[-latex] ($(robot_straight) + (0.5,0.3)$) -- ++(0.3,-0.3);
\draw[-latex] ($(robot_straight) + (0.5,0.3)$) -- ++(-0.3,-0.3);
\draw[-latex] ($(robot_straight) + (0.5,0.3)$) -- ++(0.6,0.3);
\draw[-latex] ($(robot_straight) + (0.5,0.3)$) -- ++(0.3,0.6);
\draw[-latex] ($(robot_straight) + (0.5,0.3)$) -- ++(-0.3,0.6);
\draw[-latex] ($(robot_straight) + (0.5,0.3)$) -- ++(-0.6,0.3);
\draw[-latex] ($(robot_straight) + (0.5,0.3)$) -- ++(-0.6,-0.3);
\draw[-latex] ($(robot_straight) + (0.5,0.3)$) -- ++(-0.3,-0.6);
\draw[-latex] ($(robot_straight) + (0.5,0.3)$) -- ++(0.3,-0.6);
\draw[-latex] ($(robot_straight) + (0.5,0.3)$) -- ++(0.6,-0.3);

\coordinate(robot_straight) at(3.5,1.8);
\fill[robot_base] (robot_straight) rectangle ++(1,0.6);
\fill[red] ($(robot_straight) + (-0.1,-0.25)$) rectangle ++(0.2,0.2);
\fill[red] ($(robot_straight) + (0.95,-0.25)$) rectangle ++(0.2,0.2);
\fill[red] ($(robot_straight) + (-0.1,0.65)$) rectangle ++(0.2,0.2);
\fill[red] ($(robot_straight) + (0.95,0.65)$) rectangle ++(0.2,0.2);
\fill[red] ($(robot_straight) + (0.7,0.3)$) circle (0.07);

\draw[dashed]($(robot_straight) + (0.4,0.3)$) -- ++(1.2,0);
\draw[thick] ($(robot_straight) + (0.45,0.25)$) -- ++(0.1,0.1);
\draw[thick] ($(robot_straight) + (0.45,0.35)$) -- ++(0.1,-0.1);
\draw[-latex]($(robot_straight) + (1.3,0.35)$) arc (5:60:0.35);
\draw[-latex]($(robot_straight) + (1.3,0.25)$) arc (-5:-60:0.35);

\coordinate(robot_straight) at(6.,2.6);
\fill[gray!40] ($(robot_straight) + (1.25,-0.4)$) rectangle ++(0.7,1.45);
\fill[robot_base] (robot_straight) rectangle ++(1,0.6);
\fill[red] ($(robot_straight) + (-0.1,-0.25)$) rectangle ++(0.2,0.2);
\fill[red] ($(robot_straight) + (0.95,-0.25)$) rectangle ++(0.2,0.2);
\fill[red] ($(robot_straight) + (-0.1,0.65)$) rectangle ++(0.2,0.2);
\fill[red] ($(robot_straight) + (0.95,0.65)$) rectangle ++(0.2,0.2);
\fill[red] ($(robot_straight) + (0.7,0.3)$) circle (0.07);
\draw[-latex]($(robot_straight) + (1.05,0.75)$) arc (135:35:0.35);

\draw[-latex,very thick]  ($(robot_straight) + (0.5,-0.3)$) -- ++(0.0,-0.4);

\coordinate(robot_straight) at(6.,1);
\fill[gray!40] ($(robot_straight) + (1.25,-0.4)$) rectangle ++(0.7,1.45);
\fill[robot_base] (robot_straight) rectangle ++(1,0.6);
\fill[red] ($(robot_straight) + (-0.1,-0.25)$) rectangle ++(0.2,0.2);
\fill[red] ($(robot_straight) + (0.95,-0.25)$) rectangle ++(0.2,0.2);
\fill[red] ($(robot_straight) + (-0.1,0.65)$) rectangle ++(0.2,0.2);
\fill[red] ($(robot_straight) + (1.4,0.65)$) rectangle ++(0.2,0.2);
\fill[red] ($(robot_straight) + (0.7,0.3)$) circle (0.07);

\coordinate(robot_straight) at(8.7,2.6);
\fill[gray!40] ($(robot_straight) + (1.25,-0.4)$) rectangle ++(0.7,1.45);
\fill[robot_base] (robot_straight) rectangle ++(1,0.6);
\fill[red] ($(robot_straight) + (-0.1,-0.25)$) rectangle ++(0.2,0.2);
\fill[red] ($(robot_straight) + (1.55,-0.25)$) rectangle ++(0.2,0.2);
\fill[red] ($(robot_straight) + (-0.1,0.65)$) rectangle ++(0.2,0.2);
\fill[red] ($(robot_straight) + (1.35,0.65)$) rectangle ++(0.2,0.2);
\fill[red] ($(robot_straight) + (0.7,0.3)$) circle (0.07);
\draw[-latex]($(robot_straight) + (0.9,0.3)$) -- ++(0.45,0);

\draw[-latex,very thick]  ($(robot_straight) + (0.5,-0.3)$) -- ++(0.0,-0.4);

\coordinate(robot_straight) at(8.7,1);
\fill[gray!40] ($(robot_straight) + (1.25,-0.4)$) rectangle ++(0.7,1.45);
\fill[robot_base] ($(robot_straight) + (0.4,0)$) rectangle ++(1,0.6);
\fill[red] ($(robot_straight) + (-0.1,-0.25)$) rectangle ++(0.2,0.2);
\fill[red] ($(robot_straight) + (1.55,-0.25)$) rectangle ++(0.2,0.2);
\fill[red] ($(robot_straight) + (-0.1,0.65)$) rectangle ++(0.2,0.2);
\fill[red] ($(robot_straight) + (1.35,0.65)$) rectangle ++(0.2,0.2);
\fill[red] ($(robot_straight) + (1.1,0.3)$) circle (0.07);

\coordinate(robot_straight) at(11.4,2.6);
\fill[gray!40] ($(robot_straight) + (0.7,-0.4)$) rectangle ++(1.25,1.45);
\fill[robot_base] (robot_straight) rectangle ++(1,0.6);
\fill[red] ($(robot_straight) + (-0.1,-0.25)$) rectangle ++(0.2,0.2);
\fill[red] ($(robot_straight) + (1.15,-0.25)$) rectangle ++(0.2,0.2);
\fill[red] ($(robot_straight) + (-0.1,0.65)$) rectangle ++(0.2,0.2);
\fill[red] ($(robot_straight) + (0.95,0.65)$) rectangle ++(0.2,0.2);
\fill[red] ($(robot_straight) + (0.7,0.3)$) circle (0.07);
\draw[-latex]($(robot_straight) + (1.05,0.75)$) -- ++(0.45,0);

\draw[-latex,very thick]  ($(robot_straight) + (0.5,-0.3)$) -- ++(0.0,-0.4);

\coordinate(robot_straight) at(11.4,1);
\fill[gray!40] ($(robot_straight) + (0.7,-0.4)$) rectangle ++(1.25,1.45);
\fill[robot_base] (robot_straight) rectangle ++(1,0.6);
\fill[red] ($(robot_straight) + (-0.1,-0.25)$) rectangle ++(0.2,0.2);
\fill[red] ($(robot_straight) + (1.15,-0.25)$) rectangle ++(0.2,0.2);
\fill[red] ($(robot_straight) + (-0.1,0.65)$) rectangle ++(0.2,0.2);
\fill[red] ($(robot_straight) + (1.35,0.65)$) rectangle ++(0.2,0.2);
\fill[red] ($(robot_straight) + (0.7,0.3)$) circle (0.07);

\coordinate(robot_straight) at(14.7,2.6);
\fill[robot_base] (robot_straight) rectangle ++(1,0.6);
\fill[red] ($(robot_straight) + (-0.4,-0.25)$) rectangle ++(0.2,0.2);
\fill[red] ($(robot_straight) + (1.15,-0.25)$) rectangle ++(0.2,0.2);
\fill[red] ($(robot_straight) + (-0.1,0.65)$) rectangle ++(0.2,0.2);
\fill[red] ($(robot_straight) + (0.95,0.65)$) rectangle ++(0.2,0.2);
\fill[red] ($(robot_straight) + (0.7,0.3)$) circle (0.07);
\draw[-latex]($(robot_straight) + (-0.3,-0.15)$) -- ++(0.45,0);

\draw[-latex,very thick]  ($(robot_straight) + (0.5,-0.3)$) -- ++(0.0,-0.4);

\coordinate(robot_straight) at(14.7,1);
\fill[robot_base] (robot_straight) rectangle ++(1,0.6);
\fill[red] ($(robot_straight) + (-0.1,-0.25)$) rectangle ++(0.2,0.2);
\fill[red] ($(robot_straight) + (1.15,-0.25)$) rectangle ++(0.2,0.2);
\fill[red] ($(robot_straight) + (-0.1,0.65)$) rectangle ++(0.2,0.2);
\fill[red] ($(robot_straight) + (0.95,0.65)$) rectangle ++(0.2,0.2);
\fill[red] ($(robot_straight) + (0.7,0.3)$) circle (0.07);

\node at (0.3,0.4) {a)};
\node at (3.05,0.4) {b)};
\node at (5.75,0.4) {c)};
\node at (8.5,0.4) {d)};
\node at (11.25,0.4) {e)};
\node at (14,0.4) {f)};

\end{tikzpicture}
\end{maybepreview}
	\vspace*{-2ex}
	\caption{Neighbor states can be reached by driving or stepping related motions: a) Omnidirectional driving with fixed 				orientation, b) Turning on the spot to the next discrete orientation, c) Abstract step, d) Longitudinal base shift, e) Shift a 	single foot forward, f) Shift a foot back to its neutral position.}
	\label{fig:neighbor_states}
\end{figure}
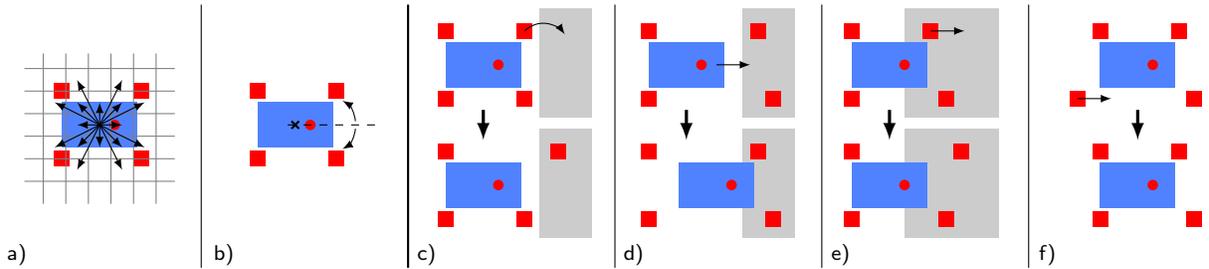

Path planning is done by a search-based approach on these pose costs. 
The used algorithm is Anytime Repairing A*~\citep{likhachev2003ara}. 
Neighbor states are generated online during the planning. 
They include omnidirectional driving (see~\cref{fig:neighbor_states}~l.) and stepping motions (see~\cref{fig:neighbor_states}~r.). 
Since driving shall be the preferred locomotion mode, stepping motions are only considered if several criteria are fulfilled. 
During path search, steps are represented as abstract maneuvers---the direct transition from a pre-stepping pose to an post-stepping pose. 
Robot stability and detailed motion sequences are explicitly not considered in this planning layer.

The resulting path is expanded to a motion sequence which can be executed by the robot. 
During this path expansion, abstract steps are transformed into stable motion sequences. 
Stable stepping poses are established through roll motions, foot shifts and longitudinal base shifts. 
Stability computation is limited to static stability since motion execution is sufficiently slow and thus, dynamic effects can be neglected. 
In addition, leg length information is generated for each pose in the resulting path.
Further details about the approach can be found in~\citep{klamt2017anytime}.

\subsection{Planning on Multiple Levels of Abstraction}
\label{sec:planning_abstraction}
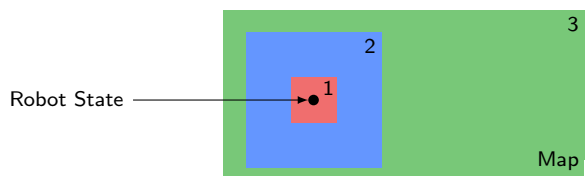
\begin{figure}
	\centering\begin{maybepreview}
\begin{tikzpicture}[
 	font=\sffamily\footnotesize,
    every node/.append style={text depth=.2ex}
]

\definecolor {mygreen}{RGB}{120, 200, 120};
\definecolor {myblue}{RGB}{100, 150, 255};
\definecolor {myred}{RGB}{240, 110, 110};

\fill [mygreen] (0.2,0.0) rectangle ++(4.8,2.2);
\fill[myblue] (0.5,0.1) rectangle ++(1.8,1.8);
\fill[myred] (1.1,0.7) rectangle ++(0.6,0.6);
\fill[black] (1.4,1) circle (2pt);

\fill[white] (-2.7,0.00) rectangle ++(2.7,0.2);
\fill[white] (5,0.00) rectangle ++(2.7,0.2);

\node[] at (1.6,1.15) {1};
\node[] at (2.15,1.7) {2};
\node[] at (4.85,2.) {3};
\node[] at (4.65,0.2) {Map};

\draw[-latex]   (-1.0,1) -- (1.33,1) node [pos=0, left] {Robot State};

\end{tikzpicture}
\end{maybepreview}
	\caption{Representation level positioning. A fine planning representation is only provided in the vicinity of the robot. With increasing distance from the robot, the representation becomes more abstract.}
	\label{fig:representation_level_positioning}
\end{figure}

The search-based planning approach is extended to plan on multiple levels of abstraction, as illustrated in~\cref{fig:representation_level_concept} and~\cref{fig:representation_level_positioning}, which allows for planning for significantly longer path queries while the result quality stays comparable. 
A detailed planning representation is only utilized in the vicinity of the robot. 
This representation is called \emph{Level~1} representation and is described in~\cref{sec:search_based_planning}. 
With increasing distance from the robot, the planning representation becomes more abstract. 
This is achieved by representing the environment and possible actions in a coarser resolution and by using a robot representation with less DoF. 
The loss of information that comes along with such abstraction is compensated by enriching those representations with additional semantics. 
The environment is represented with additional features and robot action generation accesses these features. 
All three representation levels are unified in a single planner which is able to handle transitions between these levels.
Moreover, during path execution, path segments can be refined from a coarse representation to a finer representation, which decreases the amount of required re-planning steps. 
Finally, we utilized a heuristic for the planner which is based on precomputed costs in the \emph{Level~3} representation.
Further detail can be found in~\citep{klamt2018planning}.

\begin{figure}
	\centering\begin{maybepreview}
\begin{tikzpicture}[
 	font=\sffamily\footnotesize,
    every node/.append style={text depth=.2ex},
	l/.style={font=\sffamily\scriptsize},
]

\definecolor {mygreen}{RGB}{120, 200, 120};
\definecolor {myblue}{RGB}{100, 150, 255};
\definecolor {myred}{RGB}{240, 110, 110};
\definecolor {robot_base}{RGB}{80, 130, 255};

\fill[myred] (0.0,2.8) rectangle ++(1,1.4);
\fill [myblue] (0.0,1.4) rectangle ++(1,1.4);
\fill[mygreen] (0.0,0.0) rectangle ++(1,1.4);

\node[] at (0.52,4.4) {Level};
\node[] at (0.52,3.45) {1};
\node[] at (0.52,2.08) {2};
\node[] at (0.52,0.73) {3};

\node[] at (2.75,4.4) {Map Resolution};
\node[] at (2.5,3.8) [l, anchor=west] {$\bullet$ 2.5\,cm};
\node[] at (2.5,3.5) [l, anchor=west] {$\bullet$ 64 orient.};
\node[] at (2.5,2.4) [l, anchor=west] {$\bullet$ 5.0\,cm};
\node[] at (2.5,2.1) [l, anchor=west] {$\bullet$ 32 orient.};
\node[] at (2.5,1.0) [l, anchor=west] {$\bullet$ 10\,cm};
\node[] at (2.5,0.7) [l, anchor=west] {$\bullet$ 16 orient.};

\node[] at (6.52,4.4) {Map Features};
\node[] at (6,3.8) [l, anchor=west] {$\bullet$ Height};
\node[] at (6,2.4) [l, anchor=west] {$\bullet$ Height};
\node[] at (6,2.1) [l, anchor=west] {$\bullet$ Height Difference};
\node[] at (6,1.0) [l, anchor=west] {$\bullet$ Height};
\node[] at (6,0.7) [l, anchor=west] {$\bullet$ Height Difference};
\node[] at (6,0.4) [l, anchor=west] {$\bullet$ Terrain Class};

\node[] at (10.75,4.4) {Robot Representation};
\fill[robot_base] (11,3.2) rectangle ++(1,0.6);
\fill[red] (10.9,2.95) rectangle ++(0.2,0.2);
\fill[red] (12.15,2.95) rectangle ++(0.2,0.2);
\fill[red] (11.1,3.85) rectangle ++(0.2,0.2);
\fill[red] (11.9,3.85) rectangle ++(0.2,0.2);
\fill[red] (11.7,3.5) circle (0.07);
\draw[-latex](11.1,3.05) -- ++(0.25, 0);
\draw[-latex](10.9,3.05) -- ++(-0.25, 0);
\draw[-latex](12.35,3.05) -- ++(0.25, 0);
\draw[-latex](12.15,3.05) -- ++(-0.25, 0);
\draw[-latex](11.3,3.95) -- ++(0.25, 0);
\draw[-latex](11.1,3.95) -- ++(-0.25, 0);
\draw[-latex](12.1,3.95) -- ++(0.25, 0);
\draw[-latex](11.9,3.95) -- ++(-0.25, 0);
\draw[-latex](11.77,3.5) -- ++(0.25, 0);
\draw[-latex](11.7,3.57) -- ++(0,0.25);
\draw[-latex](11.63,3.5) -- ++(-0.25,0);
\draw[-latex](11.7,3.43) -- ++(0,-0.25);
\draw[-latex](12.2,3.525) arc (5:60:0.35);
\draw[-latex](12.2,3.475) arc (-5:-60:0.35);

\fill[red!60] (10.9,1.55) rectangle ++(0.45,0.25);
\fill[red!60] (12.1,1.55) rectangle ++(0.45,0.25);
\fill[red!60] (10.9,2.4) rectangle ++(0.45,0.25);
\fill[red!60] (12.1,2.4) rectangle ++(0.45,0.25);
\fill[red!60] (11.15,1.55) rectangle ++(0.15,1.1);
\fill[red!60] (12.25,1.55) rectangle ++(0.15,1.1);
\fill[robot_base] (11,1.8) rectangle ++(1,0.6);
\fill[red] (11.7,2.1) circle (0.07);
\draw[-latex](11.35,1.65) -- ++(0.25, 0);
\draw[-latex](10.9,1.65) -- ++(-0.25, 0);
\draw[-latex](12.55,1.65) -- ++(0.25, 0);
\draw[-latex](12.1,1.65) -- ++(-0.25, 0);
\draw[-latex](11.77,2.1) -- ++(0.25, 0);
\draw[-latex](11.7,2.17) -- ++(0,0.25);
\draw[-latex](11.63,2.1) -- ++(-0.25,0);
\draw[-latex](11.7,2.03) -- ++(0,-0.25);
\draw[-latex](12.2,2.125) arc (5:60:0.35);
\draw[-latex](12.2,2.075) arc (-5:-60:0.35);

\fill[red!60] (10.7,0.15) rectangle ++(1.6,1.1);
\fill[robot_base] (11,0.4) rectangle ++(1,0.6);
\fill[red] (11.7,0.7) circle (0.07);
\draw[-latex](11.77,0.7) -- ++(0.25, 0);
\draw[-latex](11.7,0.77) -- ++(0,0.25);
\draw[-latex](11.63,0.7) -- ++(-0.25,0);
\draw[-latex](11.7,0.63) -- ++(0,-0.25);
\draw[-latex](12.2,0.725) arc (5:60:0.35);
\draw[-latex](12.2,0.65) arc (-5:-60:0.35);

\node[] at (14.92,4.4) {Action Semantics};
\node[] at (14.5,3.8) [l, anchor=west] {$\bullet$ Individual};
\node[] at (14.73,3.5) [l, anchor=west] {Foot Actions};
\node[] at (14.5,2.4) [l, anchor=west] {$\bullet$ Foot Pair};
\node[] at (14.73,2.1) [l, anchor=west] {Actions};
\node[] at (14.5,1.0) [l, anchor=west] {$\bullet$ Whole Robot};
\node[] at (14.73,0.7) [l, anchor=west] {Actions};

\draw[](0,0) rectangle (16.5,4.6);
\draw[](1,0) -- ++(0,4.6);
\draw[](4.5,0) -- ++(0,4.6);
\draw[](8.5,0) -- ++(0,4.6);
\draw[](13.0,0) -- ++(0,4.6);
\draw[](0,4.2) -- ++(16.5,0);
\draw[](0,2.8) -- ++(16.5,0);
\draw[](0,1.4) -- ++(16.5,0);

\draw[thick](1.8,0.2) -- ++(0.6,3.8) -- ++(-1.2,0) -- ++(0.6,-3.8);
\draw[thick](9.3,0.2) -- ++(0.6,3.8) -- ++(-1.2,0) -- ++(0.6,-3.8);
\draw[thick](4.65,0.2) -- ++(1.2,0) -- ++(-0.6,3.8) -- ++(-0.6,-3.8);
\draw[thick](13.2,0.2) -- ++(1.2,0) -- ++(-0.6,3.8) -- ++(-0.6,-3.8);

\end{tikzpicture}
\end{maybepreview}
	\vspace{-1em}
	\caption{The planning representation is split into three levels of abstraction. Coarser representations are enriched by additional semantics to compensate the loss of information, caused by abstraction.}
	\label{fig:representation_level_concept}
\end{figure}

\subsection{Path Execution}
\label{sec:path_execution}
Path segments that are represented in \emph{Level~1} and which are expanded to detailed motion sequences are executed by a driving-stepping controller. 
This is split into the control of omnidirectional driving and the control of leg movements.

To control path segments that require omnidirectional driving, a three-dimensional $(x,y,\theta)$ B-Spline~\citep{deboor1978splines} is laid through the next five robot poses and a pose on this B-Spline is chosen as the controller set value $\w{r_\text{sv}} = (r_{x\text{,sv}}, r_{y\text{,sv}}, r_{\theta\text{,sv}})$. 
Extracting this set value from the B-Spline in some distance from the current robot pose $\w{r} = (r_x, r_y, r_\theta)$ leads to a smoother controller output. 
A distance of half the B-Spline length yields the desired behavior. The velocity command $\w{v} = (v_x, v_y, v_\theta)$ is computed with
\begin{equation}
	\w{v} = \left( \w{r_\text{sv}} - \w{r} \right) \times k,
\end{equation}
where $k$ is chosen in a way that the linear velocity component norm $\left\lVert (v_x, v_y)^\text{T} \right\rVert$ is equal to $v_\text{des}$, the desired velocity. 
$v_\text{des}$ is 0.1\,m/s close to obstacles and 0.25\,m/s otherwise. 
With the given relative foot position $\w{f}^{(i)}$, the individual velocity command 
\begin{equation}
	\begin{pmatrix} v_x^{(i)} \\ v_y^{(i)} \\ v_z^{(i)} 	\end{pmatrix} = \begin{pmatrix} v_x \\ v_y \\ 0 	\end{pmatrix} + \begin{pmatrix} 0 \\ 0 \\ v_\theta 	\end{pmatrix} \times \w{f}^{(i)}
\end{equation} 
can be computed for each of the four wheels. 
Before moving with the linear velocity $||(v_x^{(i)}, v_y^{(i)})^\text{T}||$ in the desired direction, each wheel needs to rotate to the respective yaw angle $\alpha^{(i)} = \text{atan2}(v_y^{(i)}, v_x^{(i)})$. 
While driving, the robot continuously adjusts the ankle orientations.

For path segments that require leg movement, actions are expanded to sequences of linear end-effector movements. 
It is distinguished between movements with wheel rotation (\eg shift an individual foot on the ground relative to the robot base) and actions without wheel rotation (\eg longitudinal robot base shift). 
The transformation from the Cartesian space to joint space is described in~\cref{sec:motion_player}.

\section{Autonomous Manipulation}
\label{sec:autonomous_grasping}

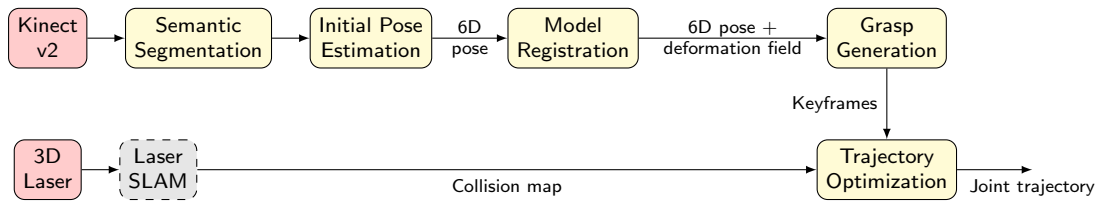
\begin{figure}[b]
 	\centering\begin{maybepreview}
\begin{tikzpicture}[
    font=\sffamily\footnotesize,
    every node/.append style={text depth=.2ex},
	sensor/.style={rectangle,rounded corners,draw=black,fill=red!20,align=center},
    module/.style={rectangle,rounded corners,draw=black,fill=yellow!20,align=center},
    group/.style={rectangle,rounded corners,draw=black},
    extmodule/.style={module,dashed,fill=black!10},
    l/.style={font=\sffamily\scriptsize},
]

\node[sensor] (kinect) {Kinect\\v2};

\node[module,right=.5cm of kinect] (semseg) {Semantic\\Segmentation};

\node[module,right=.5cm of semseg] (initial_pose) {Initial Pose\\Estimation};
\node[module,right=1cm of initial_pose] (registration) {Model\\Registration};

\node[sensor] (laser) at ($(kinect)+(0,-1.75cm)$) {3D\\Laser};
\node[extmodule,right=.5cm of laser] (slam) {Laser\\SLAM};

\node[module,right=2.5cm of registration] (grasp_generation) {Grasp\\Generation};
\node[module] (trajectory_optimization) at (grasp_generation|-laser) {Trajectory\\Optimization};

\coordinate[right=of trajectory_optimization] (out);

\draw[-latex]   (kinect) -- (semseg);
\draw[-latex] (semseg) -- (initial_pose);

\draw [-latex] (initial_pose.east) -- (registration) node [l,midway,align=center] {6D\\pose};

\draw[-latex] (registration) -- (grasp_generation) node [l,midway,align=center] {6D pose + \\deformation field};

\draw [-latex] (grasp_generation) -- (trajectory_optimization) node [l,midway,left]{Keyframes};

\draw [-latex] (laser) -- (slam);
\draw [-latex] (slam) -- (trajectory_optimization) node [l,midway,below] {Collision map};

\draw [-latex] (trajectory_optimization) -- (out) node [l,right,below] {Joint trajectory};

\end{tikzpicture} %
\end{maybepreview}
 	\vspace*{-1ex}
 	\caption{Pipeline for autonomous manipulation.
 	Sensors are colored red, pipeline components yellow, and external modules from other workpackages are colored gray.}	
 	\label{fig:manipulation:overview}
\end{figure}

Autonomous manipulation capabilities can help reducing the load on the main operator and the dependency on a good data connection.
For example, repetitive motions such as grasping a tool, opening a door, and placing an object
may be automated, allowing the operator to focus on the current higher-level task
rather than direct teleoperation of the manipulation actions.
To realize autonomous manipulation, several components were integrated
(see \cref{fig:manipulation:overview}).
Input to the system are RGB-D measurements from the \emph{Kinect V2} sensor and range measurements from the rotating laser scanner at the robot head.
In addition to the perception components described in \cref{sec:perception:object_segmentation,sec:perception:pose_estimation},
we will discuss the grasp generation and trajectory optimization modules here.
The pipeline outputs a joint trajectory describing the grasping motion which is executed by low-level control software.
A bi-manual extension of this pipeline has been proposed in~\citep{pavlichenko2018autonomous}.

\subsection{Grasp Planning}
\label{sec:grasp_gen}
Objects belonging to a category often exhibit several similarities in their extrinsic geometry. 
Based on this observation, we transfer grasping skills from known instances to novel instances belonging to the same category such as drills, hammers, or screwdrivers.
Our approach has two stages: a learning stage and an inference stage. 

\begin{figure}
	\centering\begin{maybepreview}
	\includegraphics[width=.95\linewidth]{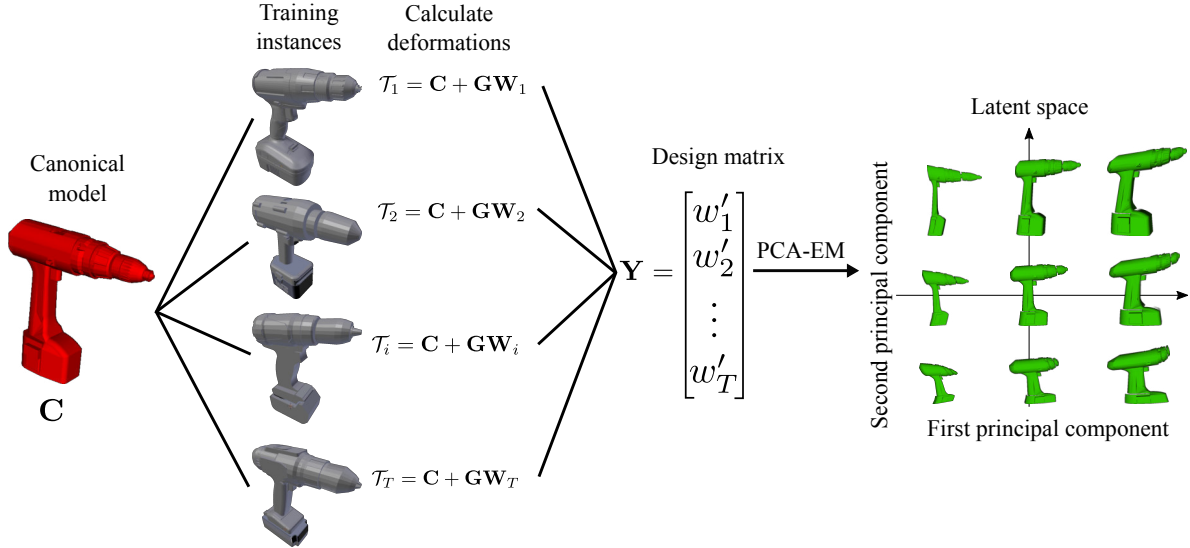}\end{maybepreview}
	\vspace*{-1ex} 
	\caption{Training phase. Deformations between the canonical model and each instance are calculated by using CPD. The deformations are then assembled into the design matrix $\mathbf{Y}$. Finally, the principal components (latent space) are found by using PCA-EM.}
	\label{fig:CLS_training}
\end{figure}

During learning, we build a category-specific linear model of the deformations that a category of objects can undergo. 
For that, we firstly define a single canonical instance of the category, and then we calculate the deformation fields relating the canonical instance to all other instances of the category using Coherent Point Drift (CPD)~\citep{myronenko2010point}. 
Next, we find a linear subspace of these deformation fields, which defines the deformation model for the category (\cref{fig:CLS_training}). 

A category is defined as a group of objects with similar extrinsic shape.
From the set of objects, we select one to be the canonical model of the class.
For each training sample, we find the deformation field that the canonical model has to undergo to transform itself into the training sample.
As explained in~\citep{Rodriguez2018}, this transformation can be expressed as:
\begin{equation}
\label{eq:defomation_field}
\mathcal{T}_i = \mathbf{C} + \mathbf{G}\mathbf{W}_i,
\end{equation}
where $\mathbf{C}$ refers to the canonical model, $\mathbf{G}$ is a Gaussian kernel matrix, and $\mathbf{W}_i$ is a matrix of kernel weights.

Because $\mathbf{C}$ and $\mathbf{G}$ remain constant across all training samples, the uniqueness of the deformation field is captured only by $\mathbf{W}_i$.
Each of the $\mathbf{W}_i$ matrices contains the same number of elements.
This allows us to assemble a $\mathbf{Y}$ design matrix containing all deformation fields as column vectors.
Finally, we apply Principle Component Analysis Expectation Maximization (PCA-EM) on the design matrix $\mathbf{Y}$ to find a lower-dimensional manifold of deformation fields for this category. 

In the inference stage, we formulate the problem as: given a newly observed instance, search this subspace of deformation fields to find the deformation field which best relates the canonical instance to the novel one. 
Associated control poses used for grasping defined for the canonical model are also transformed to the observed instance and used for the final grasping motion.

The transformation between a novel instance and the canonical model is defined by its latent vector plus an additional rigid transformation.
The function of the rigid transformation is to reduce the impact of minor misalignments in the pose between the canonical shape and the target shape. 
We use gradient descent to simultaneously optimize for pose and shape. 
In general, we aim for an aligned dense deformation field that when exerted to the canonical shape $\mathbf{C}$ minimizes the following energy function:

\begin{equation}
\label{eq:Energy}
E(\mathbf{x},\bm{\theta}) = \sum^{M}_{m=1}{\sum^{N}_{n=1}{P\norm{\mathbf{O}_n-\Theta(\mathcal{T}_m(\mathbf{C}_m,\mathcal{W}(\mathbf{x})_m),\bm{\theta})}^2}} ,
\end{equation}
where $M$ is the number of points of the canonical model, $N$ is the number of points of the observed instance $\mathbf{O}$, $\mathbf{x}$ is the latent vector, $\Theta$ is a function that applies a rigid transformation with parameters $\theta$,
and $P$ represents the probability or importance weights between points expressed as:
\begin{equation}
P = \frac{e^{\frac{1}{2\sigma^2}||\mathbf{O}_n-\Theta(\mathcal{T}(\mathbf{C}_m,\mathcal{W}(\mathbf{x})_m),\bm{\theta})||^2}}{\sum^{M}_{k=1}{{{e^{\frac{1}{2\sigma^2}||\mathbf{O}_n-\Theta(\mathcal{T}_k(\mathbf{C}_k,\mathcal{W}(\mathbf{x})_k),\bm{\theta})||^2}}}}} .
\end{equation}

\begin{figure}
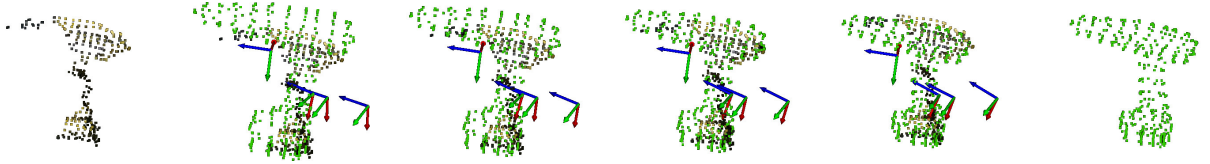

	\centering\begin{maybepreview}
	\includegraphics[height=2.1cm]{figures/manipulation/input_CLS.png}\hfill
	\includegraphics[height=2.1cm]{figures/manipulation/CLS_frame_0.png}\hfill
	\includegraphics[height=2.1cm]{figures/manipulation/CLS_frame_1.png}\hfill
	\includegraphics[height=2.1cm]{figures/manipulation/CLS_frame_2.png}\hfill
	\includegraphics[height=2.1cm]{figures/manipulation/CLS_frame_3.png}\hfill
	\includegraphics[height=2.1cm]{figures/manipulation/CLS_result.png}\end{maybepreview}
	\vspace{2ex}
	\caption{Transferring grasping knowledge to the presented novel instance. The input point cloud is at the leftmost while the inferred shape is at the rightmost.}
	\label{fig:eval:CLS}
\end{figure}

A grasping action is composed of a set of parametrized motion primitives.
Poses expressed in the same coordinate system of the shape of the object serve as the parameters of these motion primitives.
These poses are defined only for the canonical model.
For novel instances, the poses are calculated by warping the poses of the canonical model to the novel instance.
We orthonormalize the warped poses because the warping process can violate the orthogonality of the orientation. 
\cref{fig:eval:CLS} shows how the canonical model of a \textit{Drill} category is warped to fit to the observed point cloud (leftmost), the warped grasping poses are also shown. For a complete analysis and discussion about this method, please refer to~\citep{Rodriguez2018} and~\citep{Rodriguez2018b}.

The output of the grasp generation pipeline is end-effector poses on the object. These are converted into joint-space trajectories
using inverse kinematics and our keyframe-based interpolation system (see \cref{sec:motion_player}).

\subsection{Trajectory Optimization}
\label{sec:traj_opt}

In an unstructured environment, there may be obstacles which obstruct a direct way to the final pose, which would render
the output of the keyframe interpolation step unusable. An additional optimization step can repair the trajectory.
Moreover, by performing optimization of trajectory duration and actuator loads, it is possible to decrease the power consumption, which is of high importance during extended missions.
In order to reduce the time of the task completion, it is necessary to perform the planning fast.
In this subsection, we describe our approach to trajectory optimization which satisfies these criteria.

Our approach~\citep{Pavlichenko2017} extends Stochastic Trajectory Optimization for Motion Planning (STOMP)~\citep{Kalakrishnan2011}.
An initial trajectory $\Theta$ is the input to this method.
This trajectory may be very na{\"i}ve, for example a straight interpolation between the start and the goal.
The method outputs a trajectory, optimized with respect to a cost function.
Both trajectories are represented as a set of $N$ keyframes $\bm{\theta}_i \in \mathbb{R}^J$ in joint space.
The initial and goal configurations, as well as the number of keyframes $N$ are fixed during the optimization.
In order to gradually minimize the cost, the optimization is performed in an iterative manner.
The method reliably finds a feasible solution despite the fact that the initial trajectory is far from a valid one.

For collision avoidance, we consider the robot and the unstructured environment.
We assume that the robot base does not move during the trajectory execution and that the environment is static.
Two signed Euclidean Distance Transforms (EDT) are used for static objects.
One of them is used to represent the environment and is precomputed before each optimization task.
The other EDT represents the static part of the robot and is computed only if the robot base moves.
The moving parts of the robot are approximated by a set of spheres, which allows for performing fast collision checking against EDTs.

\begin{figure}
	\centering\begin{maybepreview}
	\includegraphics[width=0.26\linewidth]{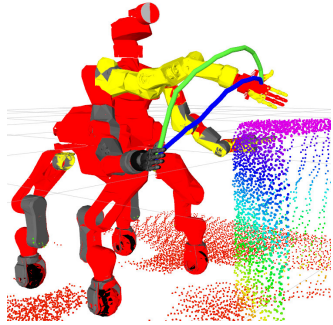}\end{maybepreview}
	\vspace{-2em}
	\caption{Two qualitatively different trajectories generated by our trajectory optimization: priority on obstacle avoidance (green) and priority on minimizing trajectory duration (blue).}
	\label{fig:manipulation:arm_trajectories}
	\vspace{-0.3em}
\end{figure}

In contrast to the original STOMP, our cost function is defined as a sum of costs of \textit{transitions} between the consequent keyframes instead of the keyframes themselves.
The cost estimated by each component is normalized to be within a $[0,1]$ interval.
This allows to attach a weight $\lambda_j \in [0,1]$ to each cost component $q_j(.,.)$ in order to introduce a prioritized optimization. An example of two qualitatively different trajectories obtained with two different values of weight $\lambda_{obst}$ for obstacle costs is shown in~\cref{fig:manipulation:arm_trajectories}.

In order to accelerate motion planning, the optimization is split into two phases.
During the first phase, we use a simplified cost function, which only includes obstacle, joint limit, and constraint costs.
Optimization with the simplified cost function continues until a valid solution is found.
In the second phase, we use the original cost function. 
This approach allows reducing the overall optimization time by eliminating duration and torque costs as long as no valid solution is found due to obstacles.
From our observations, the two-phased approach in most cases did not yield qualitatively different trajectories compared to direct full optimization.

\section{Evaluation}
\label{sec:eval}

We evaluated the CENTAURO system with dedicated tests at facilities
of the Kerntechnische Hilfsdienst GmbH in Karlsruhe, Germany (KHG).
We note that our evaluation intentionally focuses on holistic, system-level
testing of the CENTAURO system and thus aims to provide insight into the current
state-of-the-art in larger-scale robotic problems.
The chosen evaluation tasks are typical for mobile manipulation in the context of construction, maintenance, and exploration in planetary and man-made environments.
Some of the designed tasks were inspired by the DARPA robotics challenge~\citep{ROB:ROB21683}.
Each task was attempted at least once, with the final count of attempts depending
on task execution time and the assessed task difficulty. When failures were
encountered, more attempts were added to gain insight into the possible failure
modes.

In general, there were three operators with specified responsibilities. 
One operator was sitting in the telepresence suit.
A support operator was responsible for system parameter monitoring, interface management, and supported omnidirectional driving with the 4D joystick.
A second support operator controlled robot limb motions through the keyframe editor, the 6D input device, and task-specific GUIs .
Depending on the individual task, not all operators were required.
Moreover, some tasks required an additional operator who was specialist for the employed interface such as autonomous controllers.
All tasks were performed without visual contact such that the operators had to rely on information provided by our interfaces.
While the acting operators were generally experienced in the operation of robots though the provided interfaces, the tasks themselves were new to them.
The operators were allowed to inspect the task site before execution, but training runs were not allowed for any of the tasks. 
A video with footage from the experiments is available\footnote{\url{https://www.ais.uni-bonn.de/videos/JFR_2018_Centauro}}.
The results are summarized in \cref{tab:eval_tasks}.
Some of the experiments have been previously described in less detail in~\citep{klamt2018supervised}.

\begin{table}[t]
 \centering
 \begin{threeparttable}
 \caption{Tasks evaluated at Kerntechnische Hilfsdienst GmbH.}\label{tab:eval_tasks}
 \begin{tabular}{lrr@{/}lr lrr@{/}lrr@{/}lr}
  \toprule
  \multicolumn{5}{c}{Locomotion}                            &  \multicolumn{8}{c}{Manipulation} \\
  \cmidrule(lr){1-5}                                           \cmidrule(lr){6-13}
  Task              & Diff.\tnote{1} & \multicolumn{2}{c}{Success} & $t$ [s]\tnote{2}  & Task                 & Diff.\tnote{1} & \multicolumn{2}{c}{6D\tnote{3}} & $t_{\text{6D}}$ [s]\tnote{2} & \multicolumn{2}{c}{Exo\tnote{3}} & $t_{\text{Exo}}$ [s]\tnote{2} \\
  \midrule                                                                         
  Ramp              & 2              & \hspace{.6cm}3 & 3          & 83       &  Surface             & 3              & 2              & 2              & 32           & \hspace{.1cm}1 & 1          & 30 \\
  Regular door      & 6              &              3 & 3          & 680      &  Small door          & 3              & \multicolumn{2}{c}{-}           & -            & 3              & 3          & 42 \\
  Gap               & 4              &              3 & 4          & 210      &  Valve (lever)       & 4              & \multicolumn{2}{c}{-}           & -            & 2              & 2          & 44 \\
  Step field        & 7              &              2 & 2          & 885      &  Valve (gate)        & 5              & \multicolumn{2}{c}{-}           & -            & 3              & 3          & 80 \\
  Stairs            & 9              &              0 & 1          & -        &  Snap hook           & 6              & \multicolumn{2}{c}{-}           & -            & 3              & 3          & 20 \\
   \cmidrule(lr){1-5}
  Auto locomotion\tnote{4}& 9        &              3 & 3          &  256        &  Fire hose           & 7              & \multicolumn{2}{c}{-}           & -            & 3              & 3          & 128 \\
                    &                & \multicolumn{2}{c}{}        &          &  230\,V connector    & 7              & 3              & 3              & 151          & 1              & 5          & 300 \\
                    &                & \multicolumn{2}{c}{}        &          &  Cutting tool        & 8              & \multicolumn{2}{c}{-}           & -            & 3              & 9          & 33 \\
                    &                & \multicolumn{2}{c}{}        &          &  Driller             & 8              & \multicolumn{2}{c}{-}           & -            & 2              & 2          & 64 \\
                    &                & \multicolumn{2}{c}{}        &          &  Screw driver        & 10             & 3              & 3              & 613          & 0              & 2          & - \\
\cmidrule(lr){6-13}
                    &                & \multicolumn{2}{c}{}        &          &  Auto grasping & 8              & \multicolumn{6}{c}{7/14\hspace{2ex}220\,s} \\
  \bottomrule
 \end{tabular}
 \begin{tablenotes}\scriptsize
  \item [1] Difficulty score estimated by operators before the attempts (1-10).
  \item [2] Times are averaged over successful attempts.
  \item [3] Success rates are shown separately for different control methods (6D mouse and exoskeleton).
    No inputs were used for the autonomous grasping experiment.
  \item [4] This task was performed in a lab environment after the official evaluation.
 \end{tablenotes}
 \end{threeparttable}
\end{table}

In the following, we discuss the performed evaluation tasks, divided into largely-locomotion and largely-manipulation tasks.
For component level evaluation we refer to the cited works in~\cref{sec:mechanic_design,sec:advanced_environment_perception,sec:operator_interfaces,sec:locomotion_planner,sec:autonomous_grasping}.

\subsection{Locomotion Tasks}

\begin{figure}
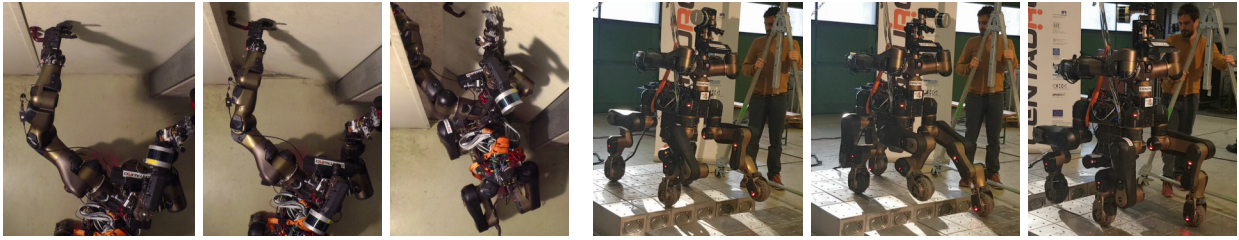

 	\centering\begin{maybepreview}
 	\newlength{\doorheight}\setlength{\doorheight}{3.1cm}
 	\includegraphics[height=\doorheight, clip, trim=0 0 0 0]{figures/evaluation/door/door2.png}
 	\includegraphics[height=\doorheight, clip, trim=0 0 0 0]{figures/evaluation/door/door3.png}
 	\includegraphics[height=\doorheight, clip, trim=0 0 0 0]{figures/evaluation/door/door4.png}\hfill
  	\includegraphics[height=\doorheight,clip,trim=60 0 0 0]{figures/evaluation/gap/gap2.png}
  	\includegraphics[height=\doorheight,clip,trim=60 0 0 0]{figures/evaluation/gap/gap3.png}
  	\includegraphics[height=\doorheight,clip,trim=60 0 0 0]{figures/evaluation/gap/gap7.png}\end{maybepreview}
 	\caption{Left: Opening the door and driving through it. 
 	Right: Overcoming a gap with the Centauro robot.}
 	\label{fig:eval_door}
\end{figure}

The tested locomotion tasks mainly focused on proving that the robot can be
effectively navigated across terrain types of different complexity:

\begin{description}
 \item[Ramp:] The robot was required to drive up and down a ramp of 120 cm length with
$20^\circ$ incline. This was accomplished using joystick teleoperation. The robot faced towards the higher platform for both subtasks.
To compensate for the slope, the base pitch was adjusted by predefined motion primitives to keep the robot base parallel to the ground.
This task was performed by the two support operators who reported that it posed no challenges and was easy to conduct with the provided interfaces.
The mainly used visualization were camera images of which especially the two cameras under the robot base provided valuable perspectives.

 \item[Regular door:] In the regular door experiment (see \cref{fig:eval_door}), the robot had to open a regular sized door (200\,cm high and 90\,cm wide)
 and drive through it. 
The door had to be opened away from the robot using a standard door handle and was not locked.
The manipulation part was successfully tested using the 6D mouse. The door handle was pushed either by moving the left wrist downwards in
just one axis, or by lowering the whole robot base.
The latter was performed through a parametrizable motion primitive.
The joystick was used to let the robot approach the door and drive through it.
Again, the two support operators conducted this task.
They reported that the robust SoftHand was valuable for this task since they had not to worry about damaging the hand, even when applying high forces to push down the door handle which reached its mechanical limit at some point.
Since no force feedback was provided for this interface, it was challenging to assess when this limit was reached.
After opening the door, the operators chose a right arm configuration such that the door was automatically pushed open when driving through it such that challenging simultaneous locomotion and manipulation control was not required.
Overall, the operators assessed the difficulty of this task to be rather low.
\end{description}

\begin{figure}
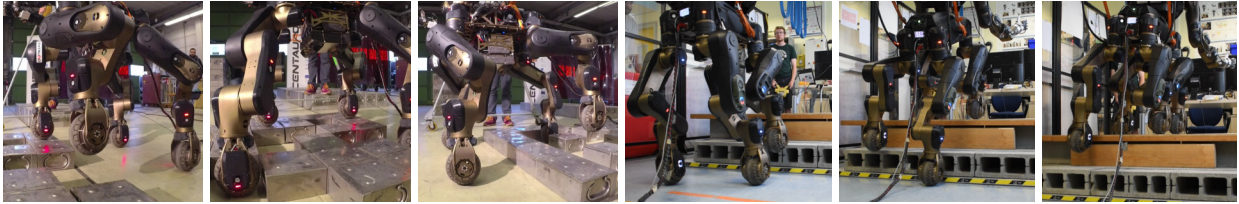

 	\centering\begin{maybepreview}
 	\newlength{\stairsheight}\setlength{\stairsheight}{2.65cm}
	\includegraphics[height=\stairsheight]{figures/evaluation/stepfield/step2.png}
	\includegraphics[height=\stairsheight]{figures/evaluation/stepfield/step6.png}
 	\includegraphics[height=\stairsheight]{figures/evaluation/stepfield/step7.png}
 	\hfill
 	\includegraphics[height=\stairsheight,clip,trim=150 150 800 0]{figures/evaluation/stairs/stairs_01.png}
 	\includegraphics[height=\stairsheight,clip,trim=450 200 600 0]{figures/evaluation/stairs/stairs_02.png}
 	\includegraphics[height=\stairsheight,clip,trim=550 300 600 0]{figures/evaluation/stairs/stairs_03.png}\end{maybepreview}
 	\caption{Complex locomotion tasks. Left: Traversing a step field. Right: Climbing stairs.}
 	\label{fig:eval_stairs_step}
\end{figure}

\begin{description}
\item[Gap:] The gap test required the robot to overcome a 30\,cm gap, which was accomplished
using predesigned motion primitives interleaved with joystick
driving commands (see \cref{fig:eval_door}). 
Motion primitives which incorporated the capability to move individual legs while under load were very helpful to obtain stable stepping configurations and solve this task successfully.
In a first attempt the robot lost balance during a stepping motion.
Moving the arms further backwards shifted the CoM and solved this problem.
Both support operators jointly accomplished this task.
Since a precise coordination of motion primitives and joystick commands was required---which were distributed among the operators---it was helpful that both operators sat next to each other and shared the same visualization.
This allowed for \eg easily pointing at details in images facilitating communication.

\item[Step field:]
Metal blocks of size 20$\times$20$\times$10\,cm were arranged in a random grid scheme and built a 200\,cm long and 120\,cm wide step field (see~\cref{fig:eval_stairs_step}).
The robot started 100\,cm in front of the step field and was controlled via the semi-autonomous stepping controller and the respective GUI (see~\cref{sec:stepping_controller}) to solve this task.
The joystick was used to approach the field and to adjust the robot's position within the task.
The latter required very careful maneuvers for which the capability of the joystick to scale the velocity command was very helpful.
Accomplishing this task put a high cognitive load on the operators which had two reasons:
\begin{enumerate}
\item Since many actions had to be triggered to overcome the whole step field, execution took long resulting in a long time of high concentration. 
A higher degree of autonomy would have been desirable to accelerate execution and relieve cognitive load from the operators.
\item Button positions and captions in the GUI could be improved to increase intuitiveness. As depicted in \cref{fig:locomotion_control}\,c, buttons were located depending on the robot top view foot position and captions were clear.
Nevertheless, especially in the later phase of the experiment, the operator controlling the GUI tended to choose wrong maneuvers or wrong feet for the next maneuver. 
Operators avoided false decisions by verbally announcing their planned next action and waiting for approval of a second operator before triggering the action.
\end{enumerate}
Nevertheless, the step field was traversed reliably two out of two attempts. 

\item[Stairs:]
The objective was to climb a flight of stairs consisting of three steps with 20\,cm height, 30\,cm depth, and 120\,cm width and ending in a platform of the same width. 
In a first test, predefined motion primitives were used to climb the steps. 
In this demanding task, leg actuators failed mainly due to overheating.
After improving the fan design for the respective actuators, this test was repeated in a later lab environment using the autonomous locomotion planner (see~\cref{sec:autonomous_locomotion}).

\end{description}

Overall, the locomotion capabilities were demonstrated successfully during the evaluation.
Especially the more complex tasks would have been impossible to finish in acceptable time without the developed autonomous assistance functions or operator interfaces.

\subsection{Telemanipulation Tasks}

\begin{figure}
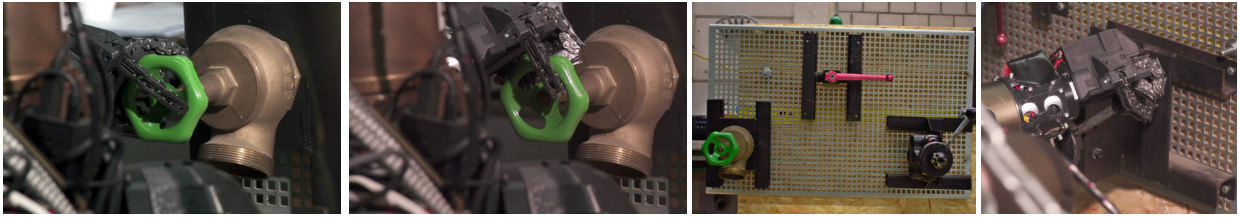

 	\centering\begin{maybepreview}
 	\newlength{\valveheight}\setlength{\valveheight}{2.8cm}
 	\includegraphics[height=\valveheight, clip, trim=170 200 650 200]{figures/evaluation/valve/valve1.png}
 	\includegraphics[height=\valveheight, clip, trim=170 200 650 200]{figures/evaluation/valve/valve2.png}\hfill
 	\includegraphics[height=\valveheight, clip, trim=50 0 50 0]{figures/evaluation/valve/valves.jpg}\hfill
 	\includegraphics[height=\valveheight, clip, trim=400 0 200 0]{figures/evaluation/valve/lever3.png}\end{maybepreview}
 	\caption{Valve experiments. Left to right: Grasping and turning of the green gate type valve,
 overview showing gate type and lever type valves, turning a lever type valve.}
 	\label{fig:eval_valve}
\end{figure}

\begin{figure}
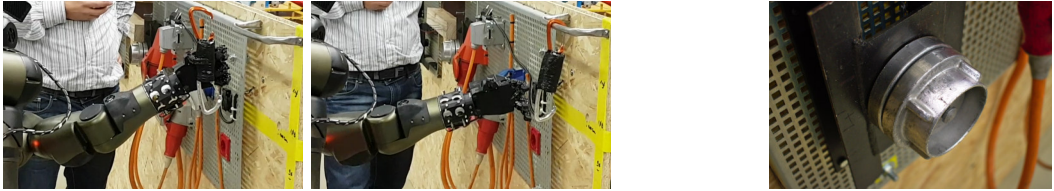

 	\centering\begin{maybepreview}
 	\newlength{\cuttingheight}\setlength{\cuttingheight}{2.5cm}
 	\includegraphics[height=\cuttingheight, clip, trim=600 250 620 400]{figures/evaluation/snap_hook/snap_hook1.png}
 	\includegraphics[height=\cuttingheight, clip, trim=500 200 720 450]{figures/evaluation/snap_hook/snap_hook3.png}\hspace*{2cm}
 	\includegraphics[height=\cuttingheight, clip, trim=0 0 0 0]{figures/evaluation/fire_hose/fire_hose.jpg}\end{maybepreview}
 	\caption{Left: Clipping a snap hook on a fixed metal bar. Right: The fire hose plug to be removed.}
 	\label{fig:eval_snap_fire}
\end{figure}

\begin{figure}
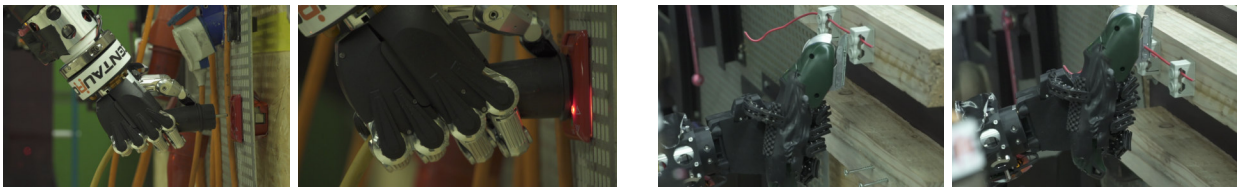

 	\centering\begin{maybepreview}
 	\newlength{\plugheight}\setlength{\plugheight}{2.4cm}
 	\includegraphics[height=\plugheight,clip,trim=200 0 0 0]{figures/evaluation/plug/plug2.png}
 	\includegraphics[height=\plugheight]{figures/evaluation/plug/plug3.png}\hfill
 	\includegraphics[height=\plugheight, clip, trim=100 0 100 0]{figures/evaluation/cutting/cutting1.png}
 	\includegraphics[height=\plugheight, clip, trim=200 0 0 0]{figures/evaluation/cutting/cutting3.png}\end{maybepreview}
 	\caption{Manipulation experiments. Left: Inserting a power plug.
 	Right: Cutting a fixed wire using a powered cutting tool.}
 	\label{fig:eval_plug_cut}
\end{figure}

The telemanipulation tasks were designed to test a large variety of manipulation capabilities. 
A first set of experiments consists of direct object manipulation task using the robot hands. 
The second set of experiments was designed to test extended robot manipulation capabilities through tool usage.
All manipulation tasks were mainly performed using the exoskeleton (see \cref{sec:telemanipulation_exo}),
with the support operators in a monitoring role.
As an alternative manipulation interface, the 6D mouse control (see \cref{sec:6d_mouse}) was employed for experiments which required very precise manipulation.

\begin{figure}
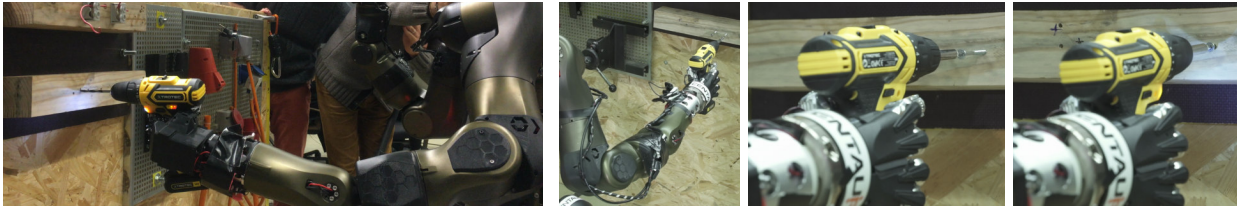

 	\centering\begin{maybepreview}
 	\newlength{\drillheight}\setlength{\drillheight}{2.7cm}
 	\includegraphics[height=\drillheight,clip,trim=400 0 50 900]{figures/evaluation/drilling/drilling.png}\hfill
 	\includegraphics[height=\drillheight,clip,trim=325 0 25 100]{figures/evaluation/screw/screw1.png}
 	\includegraphics[height=\drillheight,clip,trim=0 0 100 0]{figures/evaluation/screw/screw2.png}
 	\includegraphics[height=\drillheight,clip,trim=0 0 100 0]{figures/evaluation/screw/screw3.png}\end{maybepreview}
 	\caption{Power tool usage. Left: Drilling a hole into wood. Right: Driving a screw into wood.}
 	\label{fig:eval_drill_screw}
\end{figure}

\begin{description}
 \item [Surface:] The first task required the robot to sweep a planar surface with a (dummy)
radiation sensor in a distance less than 10\,cm without touching the surface.
This task was successfully performed both using the exoskeleton and the 6D mouse for wrist control and locomotion via joystick.
Especially useful was the ability of the 6D mouse control to constrain hand movement to the horizontal plane, parallel to the surface.
As can be seen in \cref{tab:eval_tasks}, both control modalities provided a similar fast operation and a 100\% success rate.

 \item [Small door:] In this experiment a small door (90\,cm$\times$90\,cm) had to be opened towards the robot.
 This task was successfully performed using the exoskeleton controlling the left arm and hand. 
 The robot started in front of the door
 and the handle was located roughly 1\,m above the ground.
 The exoskeleton provided fast and successful teleoperation.
 
 \item [Valve:] Two different valve types (gate and lever) were successfully manipulated using
the exoskeleton control (see \cref{fig:eval_valve}). Both valves were mounted in a height of 1\,m above the ground.
The exoskeleton operator intuitively used the palm of the hand to exert higher
forces on the lever type valve. In both cases, the mechanical limit of the valve
was detected using force feedback. Since the higher fidelity of the Schunk hand
was not required, this experiment was performed with the more robust SoftHand.

 \item [Snap hook:] In this task, a snap hook was to be clipped onto a fixed metal bar (see \cref{fig:eval_snap_fire}).
 Under the assumption that the robot brought the snap hook with it, we modified
 it slightly to make it more easily graspable.
 Again, the exoskeleton was used
 for sending wrist movement commands to the robot and a 100\,\% success rate was achieved.
 As can be seen in \cref{tab:eval_tasks}, this task proved to be fairly easy
 using this interface and was solved in little time.

 \item [Fire hose:] In a more complex experiment, a fire hose connector was to be disconnected from the wall (\cref{fig:eval_snap_fire}). The connector was secured by a lock which required a 90\textdegree\, rotation to connect and disconnect.
The robot managed to grasp and disconnect the connector. Additionally, we tried
to connect it again, but failed in all three trials that were made
due to insufficient precision under exoskeleton control.

 \item [230V connector:] An electrical plug had to be inserted by the
robot (\cref{fig:eval_plug_cut}). We attempted both \mbox{exo}skeleton and 6D mouse control.
Here, the exoskeleton test suffered from inconvenient wrist poses and insufficient precision
for inserting the plug. The 6D mouse control was more suited for this task,
since very small adjustments could be made easily.
After successful completion of the third attempt, a plastic part in the robot wrist
broke due to excessive force---the operators
had misjudged the situation slightly and did not have sufficient feedback of the
exerted forces.

 \item [Cutting tool:] In the first experiment with tool usage, the robot had to cut a wire
 held at both ends using a powered cutting tool (see~\cref{fig:eval_plug_cut}). The test was performed under
exoskeleton control. The tool was not easy to trigger by the operator, which accounts for the large
number of failed attempts. After a modification of the tool to enlarge the trigger,
operators had much less difficulty with the task and were able to complete it
quickly in the last three attempts (see~\cref{tab:eval_tasks}).
 
 \item [Driller:] The next tool experiment used a power drill to drill a hole
through wood at a specified point (see~\cref{fig:eval_drill_screw}).
The robot started with the tool in the hand. 
This task was performed without problems using exoskeleton control.

 \item [Screw driver:] The most complex telemanipulation task required the robot to fasten
a screw into a wooden block (see~\cref{fig:eval_drill_screw}).
The robot used a cordless screw driver for this task, starting with the tool in hand approx. 1\,m away from the workspace. 
The wooden block was approached using joystick locomotion control, mainly guided by camera images and the 3D laser scanner point cloud.
After approaching the workspace, the tip of the screw driver was aligned with the
screw using 6D mouse control.
This was the most difficult part of the task. Since our depth sensors were not able to measure the thin tip, operators
were forced to use 2D images for estimating the 6D offset.
For gaining an additional perspective, a small webcam was mounted on the other hand,
providing a controllable-viewpoint perspective to the operators.
After alignment was visually confirmed, the cordless screwdriver was activated using
the index finger of the robot. Multiple times, a visually confirmed alignment
turned out to be imprecise (i.e., the tip slipped off) and a further alignment
attempt was required.
During the screwing process, the operators had
to ensure that the tool tip was in constant contact with the screw head, which was facilitated using the single-axis
mode of the 6D mouse interface.
Overall, three out of three task attempts were successful.

\end{description}

In summary, the manipulation tests successfully showcased a wide variety of manipulation capabilities.
The exoskeleton allowed for intuitive and robust manipulation, while the 6D mouse control was superior for tasks which required precise incremental adjustments.
VEROSIM (see \cref{sec:oi:situationawareness}) was a very helpful tool, increasing the operator's situation awareness.
The digital twin of the robot provided views from any required perspectives onto the scene.
Additionally, full 3D models of only partial visible 3D objects or tools could be rendered into the scene to compensate for occlusions.

In some cases, the mentioned failures highlighted the deficiencies of each control mode. The exoskeleton operators
experienced difficulty performing fine-grained manipulation tasks, especially alignment tasks.
While the 6D mouse control could perform small movements and even axis-aligned movements with relative ease,
it offers no tactile feedback, leading to a misjudgment of the involved forces.
This could be seen during the plug task, where the operators damaged the robot.

Many telemanipulation tasks required cooperation between different operators,
e.g. the operator controlling locomotion using the joystick, and exoskeleton
or 6D mouse operators. Because an integrated interface including all controls
was not available, we could only evaluate this configuration.
In our opinion, this split of concerns performed well,
although it places special importance on clear communication between the
operators. One particular inconvenience is that the exoskeleton operator,
while wearing the HMD, can only communicate verbally, while certain manipulation
ideas could be much more clearly and quicker communicated by demonstration.
In the other direction, it was often first unclear what the main operator
could see from his HMD perspective, which then had to be described verbally.

\subsection{Autonomous Manipulation}

\begin{figure}
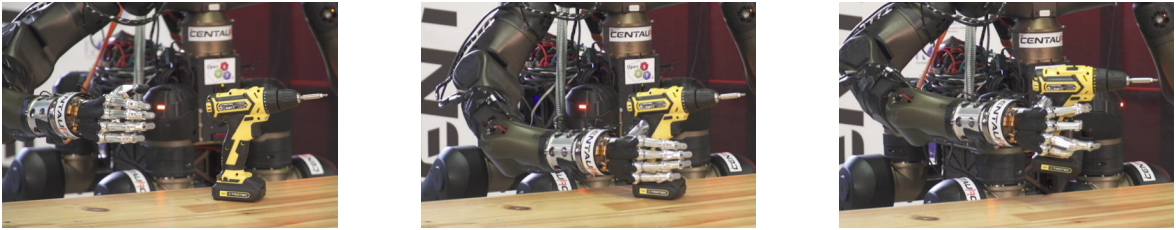

 	\centering\begin{maybepreview}
 	\newlength{\graspheight}\setlength{\graspheight}{3cm}
 	\includegraphics[height=\graspheight]{figures/evaluation/grasping/grasping1.png}\hspace*{1cm}
 	\includegraphics[height=\graspheight]{figures/evaluation/grasping/grasping3.png}\hspace*{1cm}
 	\includegraphics[height=\graspheight]{figures/evaluation/grasping/grasping4.png}\vspace*{-1ex}\end{maybepreview}
 	\caption{Autonomous grasping: approaching (left), grasping (center), and lifting the drill (right).}
 	\label{fig:eval_grasping}
\end{figure}

We tested our autonomous grasping pipeline in a dedicated experiment.
The objective was to detect, segment, and estimate the pose of a cordless driller in front of the robot (\cref{fig:eval_grasping}). 
Since this driller was previously unknown, a grasping pose was to be transferred from a known model to
the new instance after pose estimation.
Subsequently, the driller was to be grasped.
Before any motion was executed, the support operators could assess the planned
arm trajectory in detail and could trigger a re-computation if necessary.

Due to the complexity and the number of involved components this task had a higher failure rate such that we performed it multiple times. 
Failure cases include imprecise segmentation or misregistration,
both resulting in missed grasps.
In addition, we experienced hardware failures such as overheated actuators in
the arm due to longer times of inactivity between attempts.
Overall, the success rate improved during testing.

\subsection{Autonomous Locomotion}
\label{sec:autonomous_locomotion}

After improvement of the fan design for some of the lower body actuators, we tested the hybrid driving-stepping locomotion planner (see~\cref{sec:locomotion_planner}) in a later lab experiment.
The objective was to reach a goal pose on top of a flight of stairs (each step was 20\,cm high and 30\,cm long) while the starting pose was in some distance to these stairs.
After an environment representation was generated based on registered point clouds (see~\cref{sec:3d_laser_scan_assembly}), a support operator defined the desired goal pose.
Since this can be easily done by positioning a marker on visualized cost maps, no extensive experience was required. 
The planner subsequently generated a path.
Environment perception, map generation, and path computation took roughly 30 seconds.
The generated path was inspected by the operator.
For path inspection, the planner result was visualized on the generated cost map which, again, required no extensive experience.
Subsequently, the operator triggered execution.

Omnidirectional driving was used to approach the stairs. 
The robot then used the capability to shift individual feet under load to obtain stable configurations to perform steps (see~\cref{fig:eval_stairs_step}).
In addition, base pitch and roll motions as well as base shift motions were used for stabilization.
To facilitate balancing, the robot changed between different arm configurations for steps with the front and rear feet.

As can be seen in \cref{fig:eval_stairs_step}, we used concrete blocks to build the staircase.
Those possessed holes such that the vertical segments of the staircase were not plane.
Unfortunately, when performing a step with the right rear foot, the wheel got slightly stuck in such a hole which affected the robot balance.
Since this setup does not represent a realistic step designs, a person at location was allowed to give the robot a slight push to regain balance.
However, to increase system robustness, we derived the need for further improvement of the robot localization such that it can position itself on the staircase with a higher precision and avoids such contacts.
Overall, three out of three attempts were successful for this experiment.

\section{Lessons Learned \& Conclusion} 

As always, the development and testing of such a complex system is a learning experience for all involved parties.
We already reported some task-level insights in \cref{sec:eval}.
In this section, we will take a step back and discuss the resulting insights on a higher level.

First of all, system integration is challenging, especially in an international project, where site visits between
the partners are possible only for short durations.
A precise definition of interfaces and interactions between the partner modules is of key importance, but
difficult to achieve at the beginning of the project.
In particular, interdependencies might not be immediately obvious. For example, a slightly bigger battery than
expected resulted in less movement range for the legs.

A second insight is that operators still prefer 2D camera images over 3D visualizations. While the latter are
certainly helpful in addition to 2D images, in general, operators tended to rely on camera images.
Since we still believe that immersive 3D visualization is a fundamentally better solution, this can 
mean that the available 3D sensing, aggregation, and fusion with 2D camera images is not good enough yet.
Also, more operator training with the 3D visualization could help.

The most appropriate interface for teleoperation is highly dependent on the specific task.
Our telepresence suit provides a higher degree of immersion and thus allows more intuitive teleoperation, but for some tasks, other interfaces showed better performance.
\eg operating an electrical screw driver for driving a screw was easier with the 6D mouse interface.
Still, as highlighted by the plug task, the 6D mouse interface has its own drawbacks.
Careful choice of the right teleoperation interface is thus highly important.
Also, further developments of the operator interfaces, e.g. towards hybrid interfaces which
can change input modalities on the fly to match the current situation, should improve
the ease of use considerably.

The performed system-level evaluation was especially useful for identifying issues
and inconveniences that do not turn up during component-level testing.
For example, the exoskeleton operators had trouble due to slightly different
workspaces of the exoskeleton and the real robot (in particular, wrist positions
were sometimes suboptimal for the operator). After the evaluation, a dynamic
offset between the operator hand position and the robot end-effector was introduced,
which can be adjusted by a support operator and allows the exoskeleton operator
to manipulate in a comfortable posture.
Furthermore, as evidenced by several experiments, the 6D input device suffers
from inexistent force feedback. As it nevertheless proved its worth for intricate
manipulation, we will add further feedback mechanisms such as visualizations
of measured contact forces and auditory signals at predefined force thresholds.

During the evaluation we observed that communication between the different
operators is highly important. While the described arrangement worked sufficiently
for solving the tasks, some situations showed potential for improvement,
especially communication between the main operator and the support operators.
We recommend a) some method for the HMD-wearing main operator to see the other
operators (e.g. by rotating the HMD out of the way or by means of an additional
camera) and b) providing an additional view of the HMD perspective to the support
operators, so that they can better refer to what the main operator sees.

A detailed simulation-based and specification-driven hardware design is not sufficient if the simulations or specifications are imprecise.
In our case, the actuators failed in some cases, such as in complex robot configurations on the staircase.
A detailed prediction of occurring forces and moments was not possible since the maneuvers were not known during the hardware development but where developed in parallel as part of the autonomous planning components.
After identifying particularly stressed actuators during the evaluation, we
improved the actuator cooling system with special focus on these points,
which already showed highly positive results during the repeated stair experiments.

More general, in the context of space application, data transmission is usually affected by high latency and potential risk of data loss.
While interfaces with a high degree of autonomy are immune to such problems, a weakened data link strongly affects e.g., the telepresence suit control or operator visualization.
Nevertheless, since, we believe that a set of interfaces with complementary strengths is indispensable for advanced teleoperation, we aim at conducting future research in this direction.
A promising idea is to extend the presented pipeline to an architecture in which the operator solely interacts with digital twins of the robot and its environment which are frequently updated from delayed data of the real robot.
The same applies to the transmission pipeline from the digital twins to the real robot.
While we already demonstrated how 3D visualizations and force feedback are generated from VEROSIM, updates of the digital twins require prediction from the delayed signals to the current time stamp.
This is challenging but recent developments in this direction make such ideas promising.

Overall, we demonstrated a versatile, effective, and robust robotic system capable of solving a wide range of complex mobile manipulation tasks under the guidance of remote human operators, e.g., for construction, maintenance, and exploration on other planets as well as disaster response tasks on earth.
The robot is accompanied by a set of operator interfaces on different levels of autonomy which have complementary strengths and are capable of addressing a wide range of the provided robot flexibility.
All components have been integrated into a holistic remote mobile manipulation system.
During the evaluation, the system met all expectations and provided valuable insights for the developments in the final phase of the CENTAURO project, towards providing robotic assistance and removing the need for direct human work in dangerous and inaccessible environments altogether.

\subsubsection*{Acknowledgment}

This work was supported by the European Union's Horizon 2020 Programme under Grant Agreement 644839 (CENTAURO).

\bibliographystyle{apalike}
\bibliography{paper}

\end{document}